\documentclass[journal]{IEEEtran}

\usepackage{lineno}
\usepackage{graphicx}
\graphicspath{{./figures//}}
\usepackage{amsmath} 
\usepackage{amssymb}
\usepackage{times}
\usepackage{epsfig}
\usepackage{enumerate}
\usepackage{epstopdf}
\usepackage{algorithm}
\usepackage{algorithmic}
\usepackage{color}
\usepackage{xcolor}
\usepackage{verbatim}
\usepackage{flushend}
\usepackage{url}

\usepackage{array}
\usepackage{subfigure}
\usepackage{multirow}
\usepackage{mathrsfs}
\usepackage{xcolor}
\usepackage{comment}
\usepackage{setspace}
\usepackage{ragged2e}
\usepackage{paralist}
\usepackage[font=small]{caption}
\usepackage[square,numbers,sort&compress]{natbib}
\usepackage{hyperref}
\usepackage[english]{babel}
\usepackage[normalem]{ulem}

\newcommand{\reffig}[1]{Fig.\,\ref{#1}}

\newcommand{\mysubsubsection}[1]{{\bf #1:}}

\ifCLASSINFOpdf
\else
\fi

\begin{document}
%
\title{\huge Beyond Counting: Comparisons of Density Maps for Crowd Analysis Tasks - Counting, Detection, and Tracking}
%
%
%
\author{Di Kang,
        Zheng Ma,~\IEEEmembership{Member,~IEEE,}
        Antoni B. Chan~\IEEEmembership{Senior Member,~IEEE,}
}

%
%

\markboth{Journal of \LaTeX\ Class Files,~Vol.~XX, No.~X, MM~YY}%
{Shell \MakeLowercase{\textit{et al.}}: Bare Demo of IEEEtran.cls for IEEE Journals}
%




\maketitle


\begin{abstract}

For crowded scenes, the accuracy of object-based computer vision methods declines when the images are low-resolution  and objects have severe occlusions.
Taking counting methods for example, almost all the recent state-of-the-art counting methods bypass explicit detection and adopt regression-based methods to directly count the objects of interest.
Among regression-based methods, density map estimation, where the number of objects inside a subregion is the integral of the density map over that subregion, is especially promising because it preserves spatial information, which makes it useful for both counting and localization (detection and tracking). With the power of deep convolutional neural networks (CNNs) the counting performance has improved steadily.
The goal of this paper is to evaluate density maps generated by density estimation methods on a variety of crowd analysis tasks, including counting, detection, and tracking.
Most existing CNN methods produce density maps with resolution that is smaller than the original images, due to the downsample strides in the convolution/pooling operations.
To produce an original-resolution density map, we also evaluate a classical CNN that uses a sliding window regressor to predict the density for every pixel in the image.
We also consider a fully convolutional (FCNN) adaptation, with skip connections from lower convolutional layers to compensate for loss in spatial information during upsampling.
In our experiments, we found that the lower-resolution density maps sometimes have better counting performance.
In contrast, the original-resolution density maps improved localization tasks, such as detection and tracking, compared to bilinear upsampling the lower-resolution density maps.
Finally, we also propose several metrics for measuring the quality of a density map, and relate them to experiment results on counting and localization.
\end{abstract}



%

\section{Introduction}

Automatic analysis of crowded scenes in images and videos has applications in crowd management, traffic control, urban planning, and surveillance.
The number of people and how they are spatially arranged are two useful measurements for understanding crowded scenes.
However, counting, detection and tracking are still very challenging in low resolution surveillance videos, where people may only be a few pixels tall and occlusion frequently occurs -- sometimes it can be very difficult even for a human expert.
Detection-based methods can be used to count objects, but both their detection and counting performance will decrease as the scene becomes more crowded and the object size decreases.
In contrast, regression-based counting methods \cite{Chan2008,Ryan2009,NIPS2010_4043,Arteta2014,Idrees2013,Wang2015} are more suitable for very crowded situations.
However, most regression-based methods are designed only to solve the counting task, and cannot be used to localize the individual object in the scene.
Methods that can simultaneously count and predict the spatial arrangement of the individuals will be more useful, since situations where many people are crowded into a small area are very different from those where the same number of people are evenly spread out.

\par
Counting using {\em object density maps} has been shown to be effective at object counting \cite{NIPS2010_4043,Fiaschi2012,Arteta2014,Zhang2015,Pham2015}.
In an object density map, the integral over any region is the number of objects within the corresponding region in the image.
Similar to other regression-based methods, density-based methods bypass the difficulties caused when objects are severely occluded, by avoiding explicit detection, while also maintaining spatial information about the crowd,
making them very effective at solving the object counting problem, especially in situations where objects have heavy inter-occlusion and appear in low resolution surveillance videos.
Several recent state-of-the-art counting approaches \cite{NIPS2010_4043,Fiaschi2012,Arteta2014,Zhang2015,zhang2016single} are based on density estimation.
Furthermore, several works have explored how density maps can be directly used to localize individual objects \cite{Ma_2015_CVPR}, or used as a prior for improving traditional detection and tracking methods \cite{Rodriguez2011}.
These works \cite{Ma_2015_CVPR,Rodriguez2011} show the potential of applying density maps to other crowd analysis tasks.

\par
The performance of density-based methods highly depends on the types of features used, and many methods use hand-crafted features \cite{NIPS2010_4043,Arteta2014,Idrees2013} or separately-trained random-forest features \cite{NIPS2010_4043,Arteta2014,Fiaschi2012,Pham2015}.
For very challenging tasks, e.g., very crowded scenes with perspective distortion, it is hard to choose which features to use.
Indeed \cite{Idrees2013} used multi-source information to improve the counting performance
and improve robustness.

\par
One advantage of using deep neural networks is its ability to learn powerful feature representations.
Recently, \cite{Zhang2015,Xie2015,zhang2016single,onoro2016towards,Arteta2016} introduced deep learning for estimating density maps for object counting.
In \cite{Zhang2015}, a CNN is alternatively trained to predict the crowd density map of an image patch, and predict the count within the image patch.
The density map of the image is obtained by averaging density predictions from overlapping patches.
In \cite{zhang2016single}, a density map is predicted using three CNN columns with different filter sizes, which encourages the network to capture responses from objects at different scales.
Both \cite{Zhang2015,zhang2016single} predict a reduced-resolution density map, due to the convolution/pooling stride in the CNNs. Density maps obtained by \cite{Zhang2015} also have problems of block artifacts and poor spatial compactness due to the way that the patches are merged together to form the density map.
While these characteristics do not affect counting performance much, they do prevent individual objects from being localized well in the density map.


\par
\begin{figure}[tbp]
\centering
\includegraphics[width=0.8\linewidth]{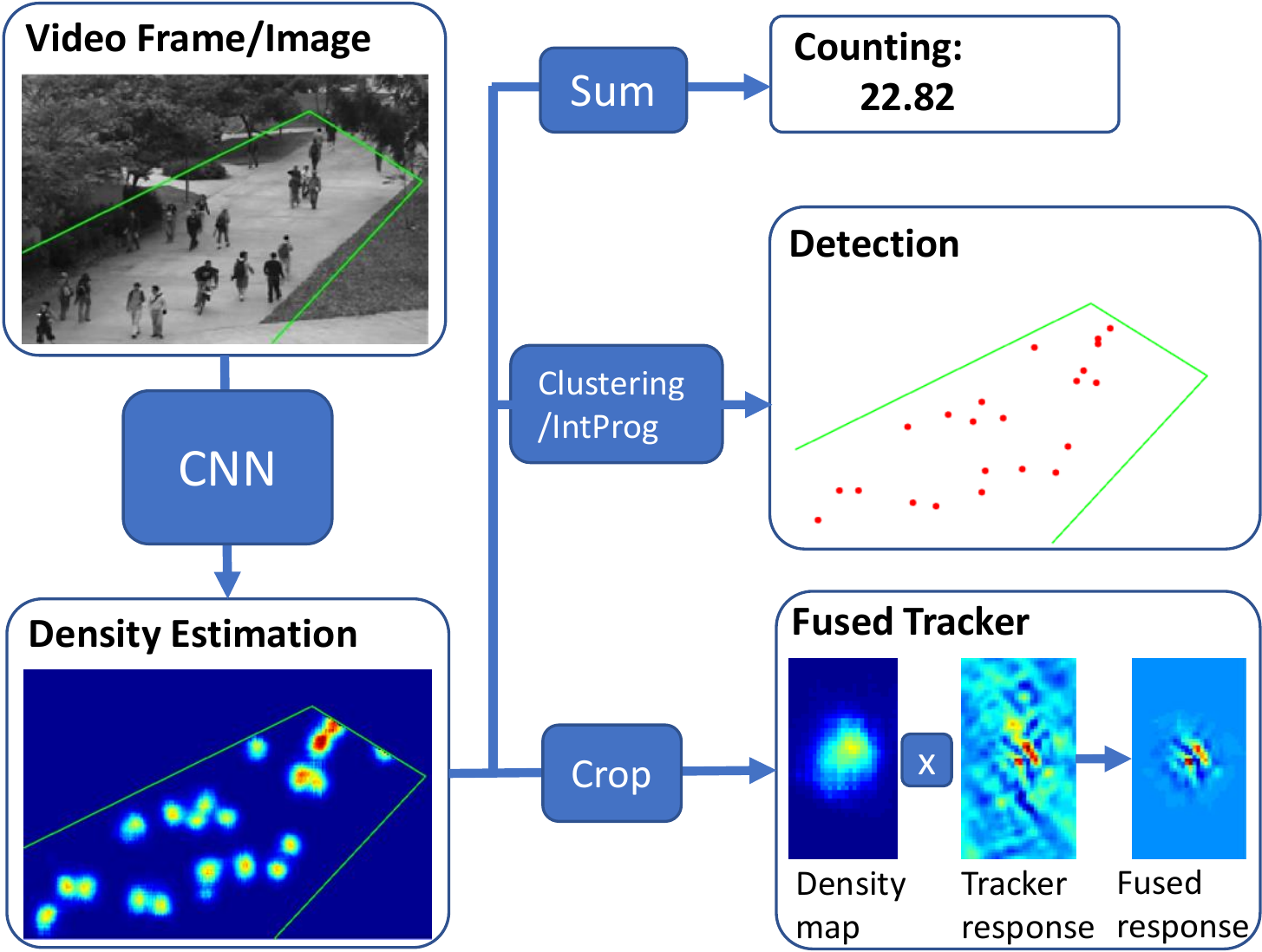}
\caption{
Crowd density maps for counting, detection, and tracking.
A crowd count is obtained by summing the density map predictions within the ROI.
Detections (the red dots in the figure) are obtained by GMM-weighted clustering or an Integer Programming formulation.
Person trackers are improved by fusing (multiplying) the response of a traditional KCF tracker with the crowd density map.
}
\label{fig:pipeline}
\end{figure}

In this paper, we focus on a comparison of density map estimation methods and their performance on counting and two localization tasks, detection and tracking (see Fig.~\ref{fig:pipeline}).
Since existing CNN-based methods normally generate a reduced-resolution density image,
either to reduce parameters in the fully-connected layers or due to the downsample stride in the convolution/pooling layers,
we also evaluate a full-resolution density map produced using a classic CNN and sliding-window to predict the density value for every pixel in the image (denoted as CNN-pixel).
We also test a fully-convolutional adaptation of the CNN-pixel. To ensure its full-resolution output, the reduced resolution density map is upsampled to the original resolution using upsampling and convolution layers. Skip connections from the lower convolutional layers are used  to compensate the spatial information.
Various metrics are used to evaluate the quality of the density maps, in order to reveal why some density maps can perform better on localization tasks.
We find that density maps that are spatially compact, well-localized, and temporally smooth are more suitable for detection and other future applications, such as object tracking.

\par
The contributions of this paper are four-fold:
1) we present a comparison of existing CNN-based density estimation methods and two classic CNN-based methods on three crowd analysis tasks: counting, detection, and tracking;
2) we show that good detection results can be achieved using simple methods on high-resolution high-quality density maps;
3) we show that high-quality density maps can be used to improve the tracking performance of standard visual trackers in crowds;
4) we propose several metrics on density maps as indicators for good performance at localization tasks.

\par
The remainder of the paper is organized as follows.
Section \ref{related_work} reviews related work in crowd counting, and
Section \ref{framework} introduces our object counting method.
Metrics for measuring the quality of various kinds of density maps are discussed in Section \ref{measure}, and experimental results on counting, detecting, and tracking are presented in Section \ref{experiments}.

\section{Related Work} \label{related_work}

In this section we review crowd counting methods, and detection and tracking methods based on density maps.

\subsection{Regression-based Counting}
Perfectly detecting all the people or tracking every person in a video can solve the counting problem, and detection-based \cite{wu2005detection} and tracking-based \cite{brostow2006unsupervised,rabaud2006counting} counting methods have been proposed. However, their performances are often limited by low-resolution and severe occlusion.
In contrast, regression-based counting methods directly map from the image features to the number of people, without explicit object detection. Such methods normally give better performance for crowded scenes by bypassing the hard detection problem.
For regression-based methods, initial works are based on regressing from global image features to the whole image or input patch count \cite{Kong2006,Cho1999,Chen2013,Chen2012,Idrees2013}, discarding all the location information, or mapping local features to crowd blob count based on segmentation results \cite{Chan2008,Ryan2009,Chan2012}.
These methods ignore the distribution information of the objects within the region, and hence cannot be used for object localization.
Extracting suitable features is a crucial part of regression-based methods: \cite{Chan2008} uses features related to segmentation, internal edges, and texture;
\cite{Idrees2013} uses multiple sources (confidence of head detector, SIFT, frequency-domain analysis) because the performance of any single source is not robust enough, especially when the dataset contains perspective distortion and severe occlusion.

\par
However, CNNs can be trained to extract suitable task-specific features automatically, and are inherently multi-source because they use many feature maps.
Another advantage of the CNN approaches is that they do not rely on foreground-background segmentation (e.g., normally based on motion segmentation), and hence can better handle single images or stationary objects in a scene.

\begin{table*}[tbhp]
\centering
\small
  \resizebox{\textwidth}{!}{
  \begin{tabular}{|c|c|c|c|c|}
      \hline
      Methods                           & loss function & prediction & output resolution & feature/prediction method  \\
      \hline
      MESA \cite{NIPS2010_4043}         & MESA (region) & pixel   & full       & random forest features, linear  \\
      Ridge regression (RR) \cite{Arteta2014}       & per-pixel squared-error       & pixel   & full       & random forest features, linear  \\
      Regression forest \cite{Fiaschi2012}  & Frobenius norm & patch   & full    & regression random forests \\
      COUNT forest \cite{Pham2015}      & Frobenius norm or entropy & patch & full & regression random forests \\
      \hline
      CNN-patch \cite{Zhang2015}        & per-pixel squared-error & patch reshaped from FC & reduced & CNN \\
      Hydra CNN \cite{onoro2016towards}     & per-pixel squared-error & patch reshaped from FC & reduced & CNN \\
      CNN-boost \cite{Walach2016}       & per-pixel squared-error & patch reshaped from FC & reduced & CNN \\
      cell-FCNN \cite{Xie2015}        & per-pixel squared-error & image from conv layer & full & FCNN \\
      MCNN \cite{zhang2016single}       & per-pixel squared-error & image from conv layer & reduced & FCNN \\
      CNN-pixel                 & per-pixel squared-error & pixel   & full & CNN \\
      FCNN-skip                 & per-pixel squared-error & image from conv layer & full & FCNN \\
      \hline
    \end{tabular}
    }
\caption{
Comparison of methods for estimating object density maps: (top) using traditional features; (bottom) using deep learning.
}
\label{tab:strategy}
\end{table*}

\begin{figure*}[tbhp]
\centering
\small
\begin{tabular}{@{}c@{\hspace{1.5mm}}c@{\hspace{1.5mm}}c@{\hspace{1.5mm}}c@{}}
  (a) image & (b) Ground Truth & (c) MESA \cite{NIPS2010_4043} & (d) Ridge regression \cite{Arteta2014} \\
  \includegraphics[width=0.24\textwidth]{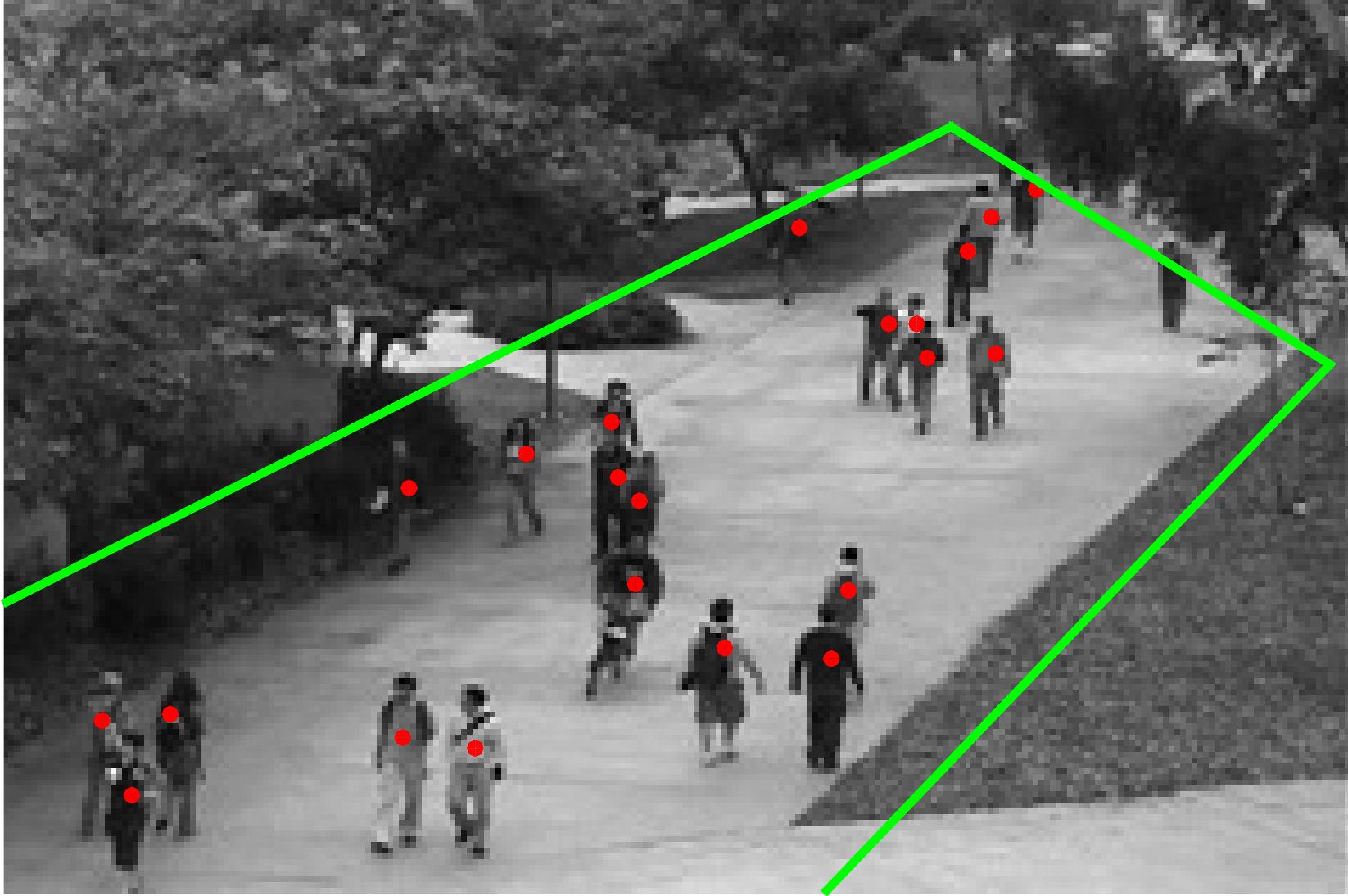} &
  \includegraphics[width=0.24\textwidth]{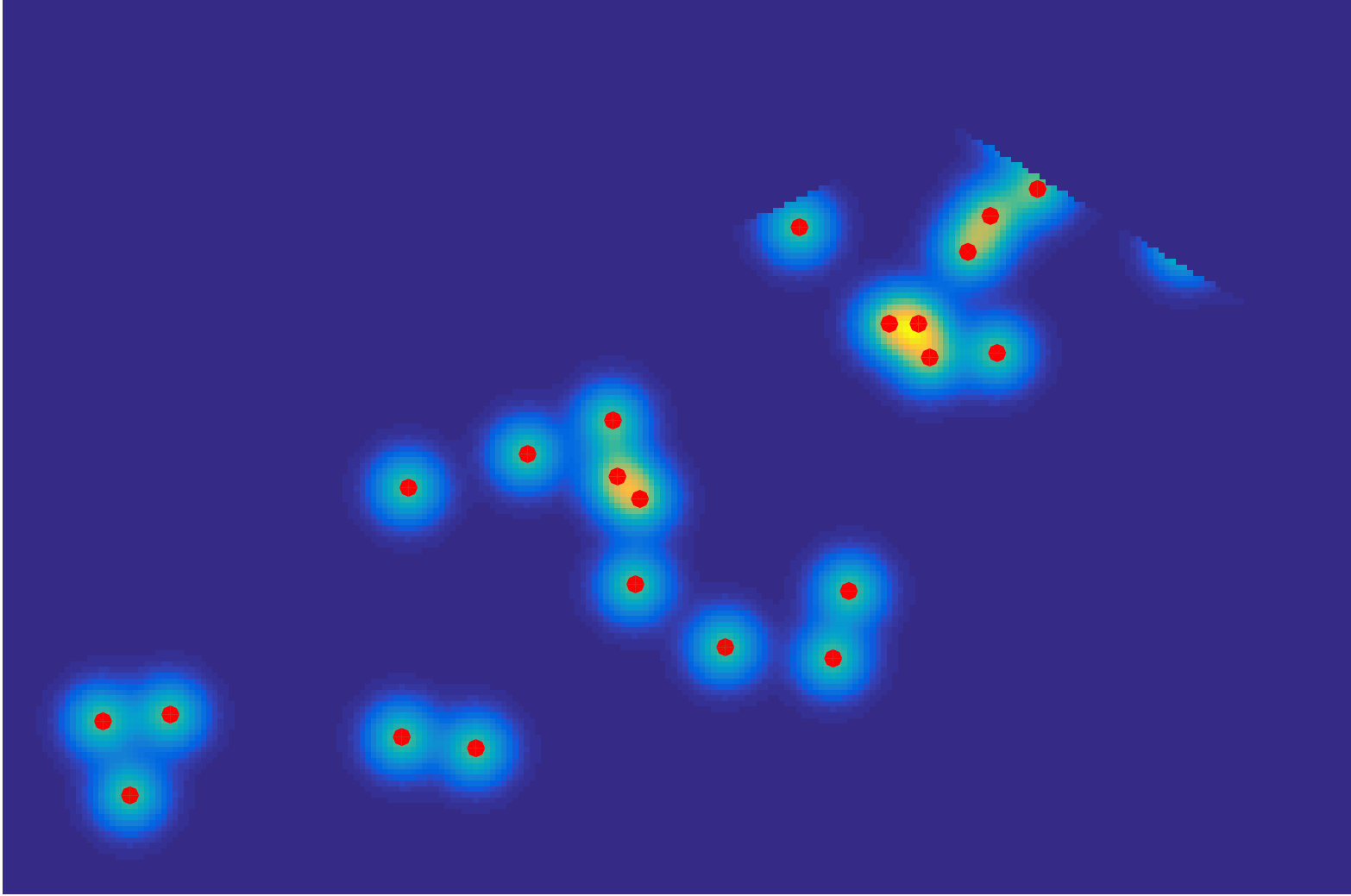} &
  \includegraphics[width=0.24\textwidth]{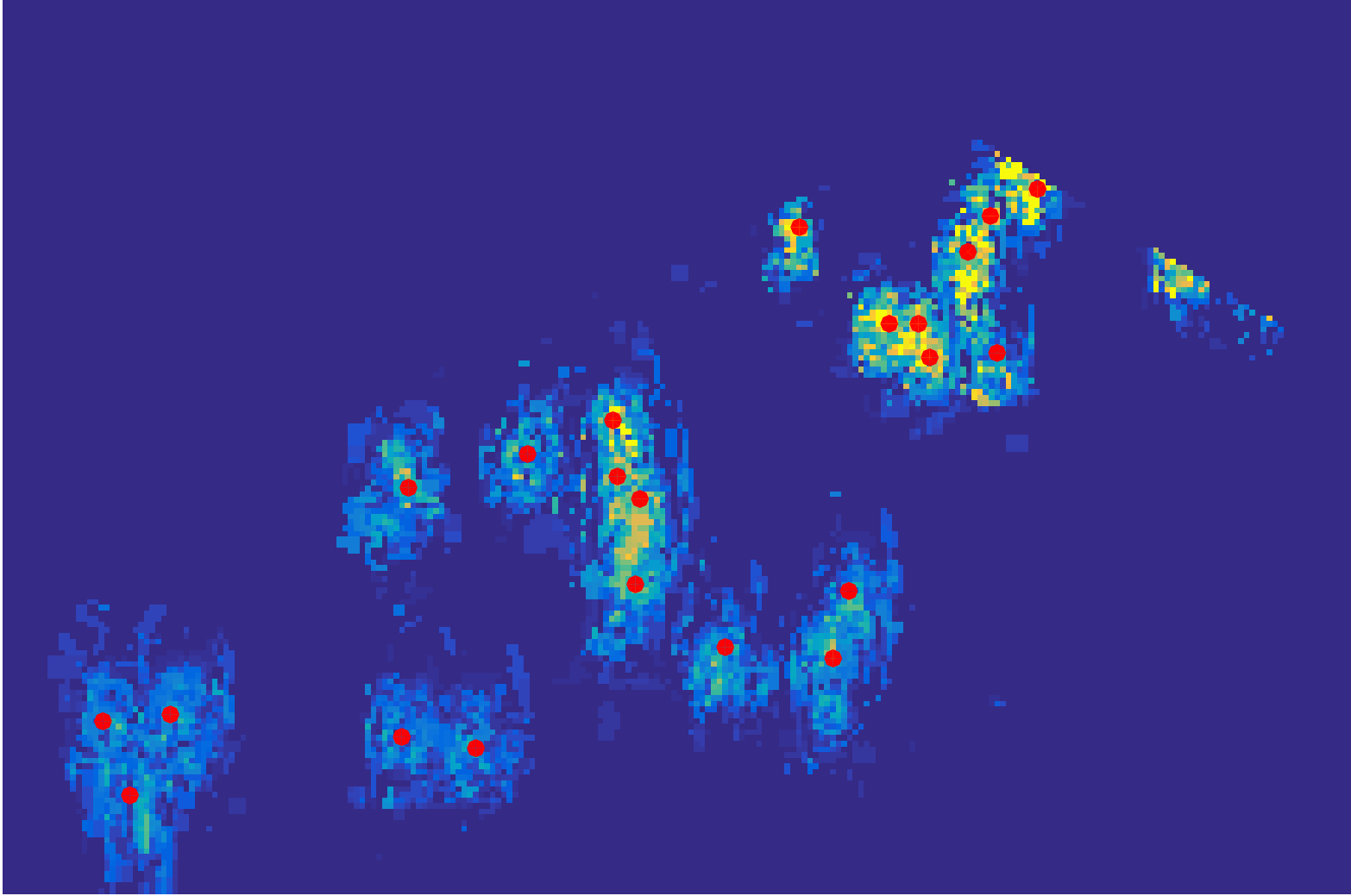} &
  \includegraphics[width=0.24\textwidth]{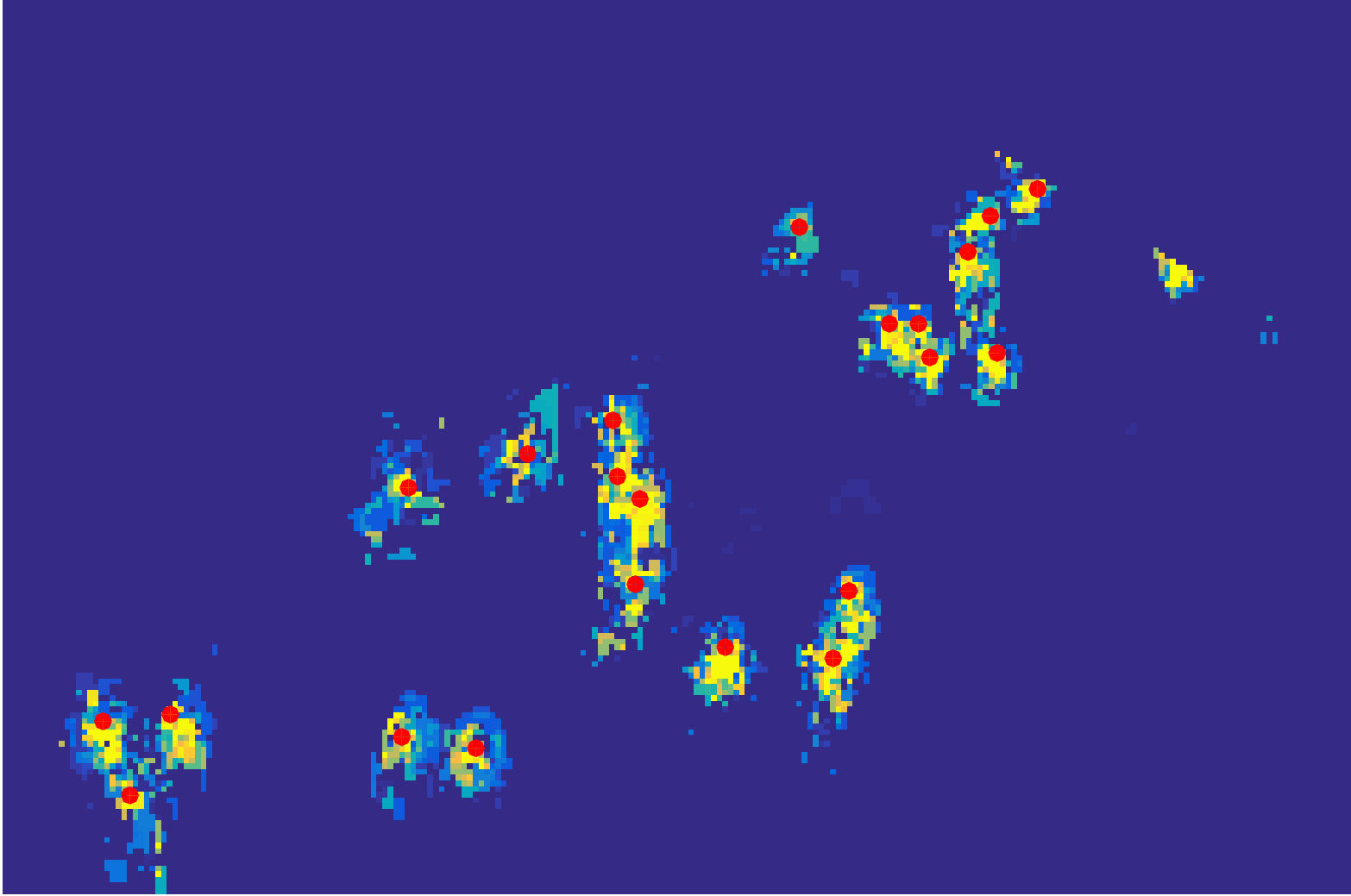} \\
  (e) CNN-patch \cite{Zhang2015} & (f) MCNN \cite{zhang2016single} & (g) FCNN-skip (ours) & (h) CNN-pixel (ours)  \\
  \includegraphics[width=0.24\textwidth]{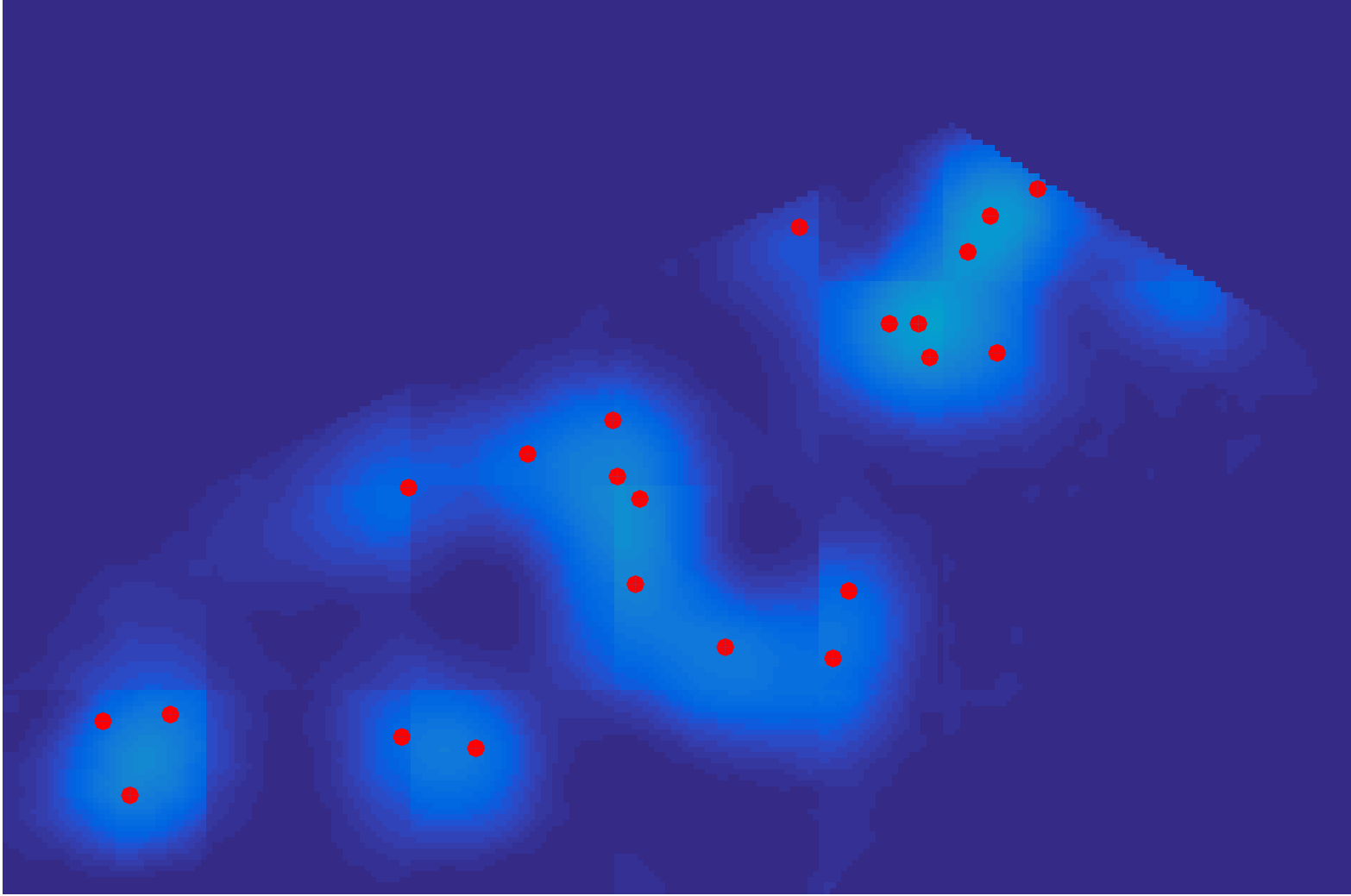} &
  \includegraphics[width=0.24\textwidth]{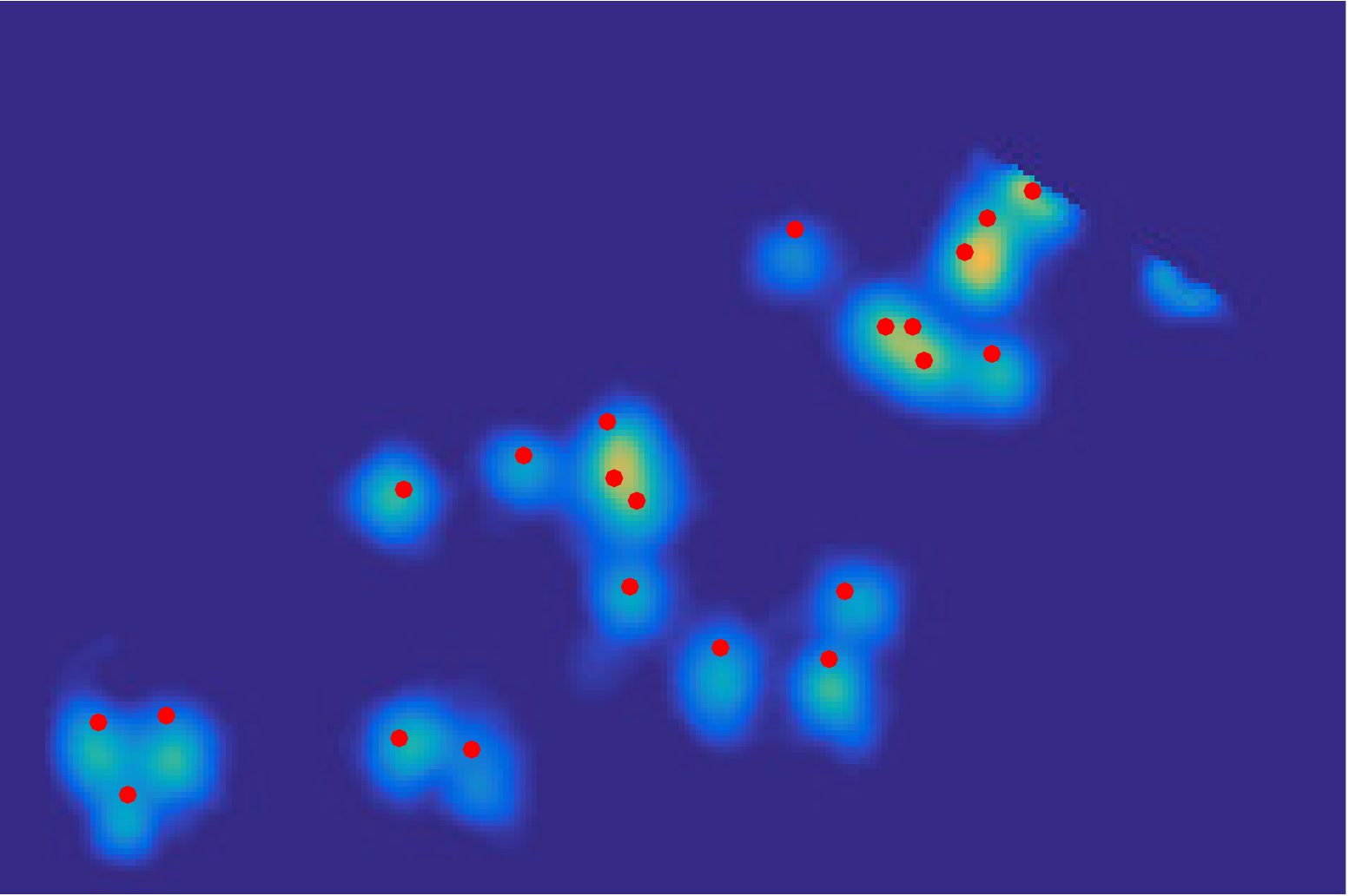} &
  \includegraphics[width=0.24\textwidth]{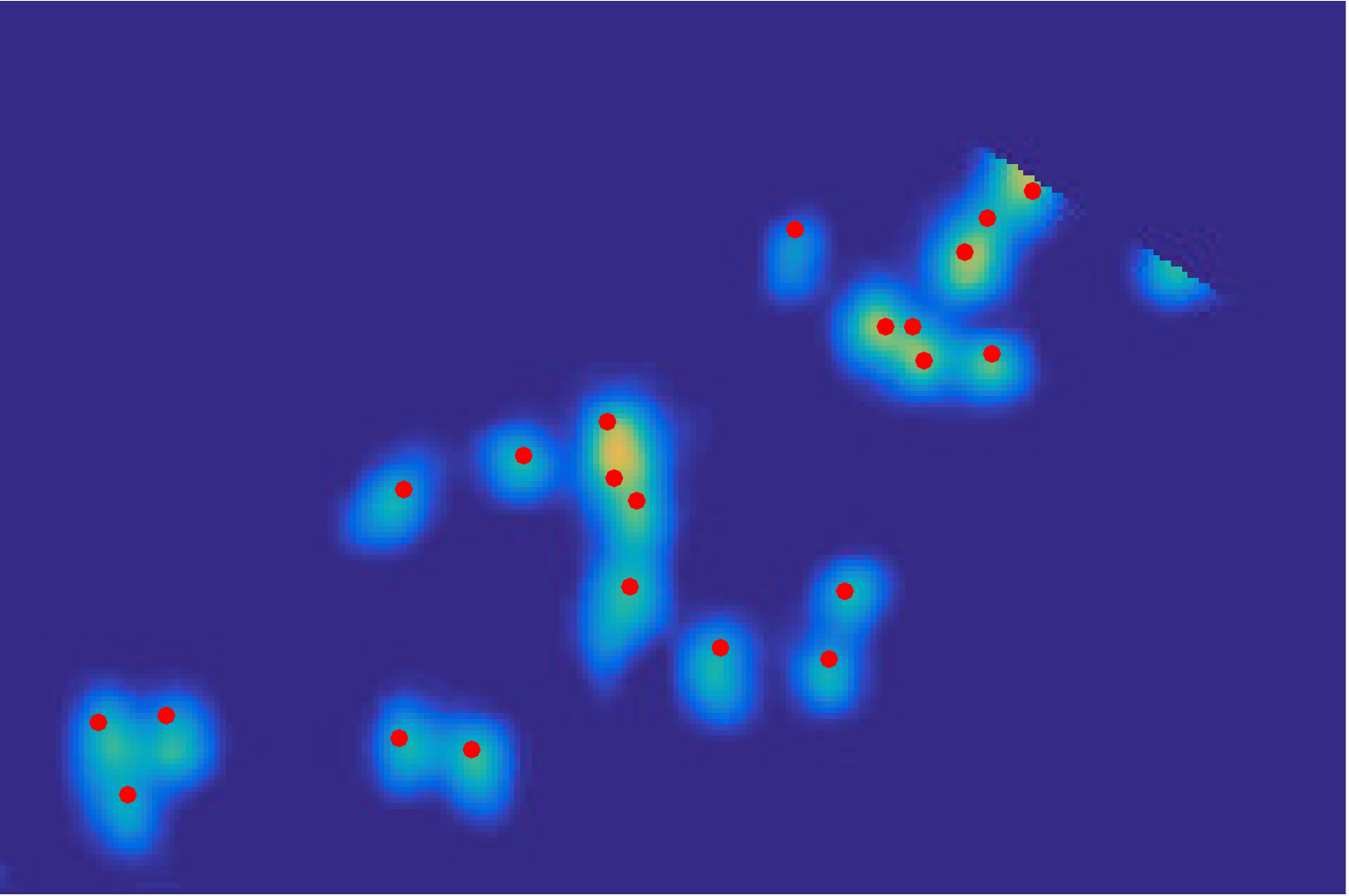} &
  \includegraphics[width=0.24\textwidth]{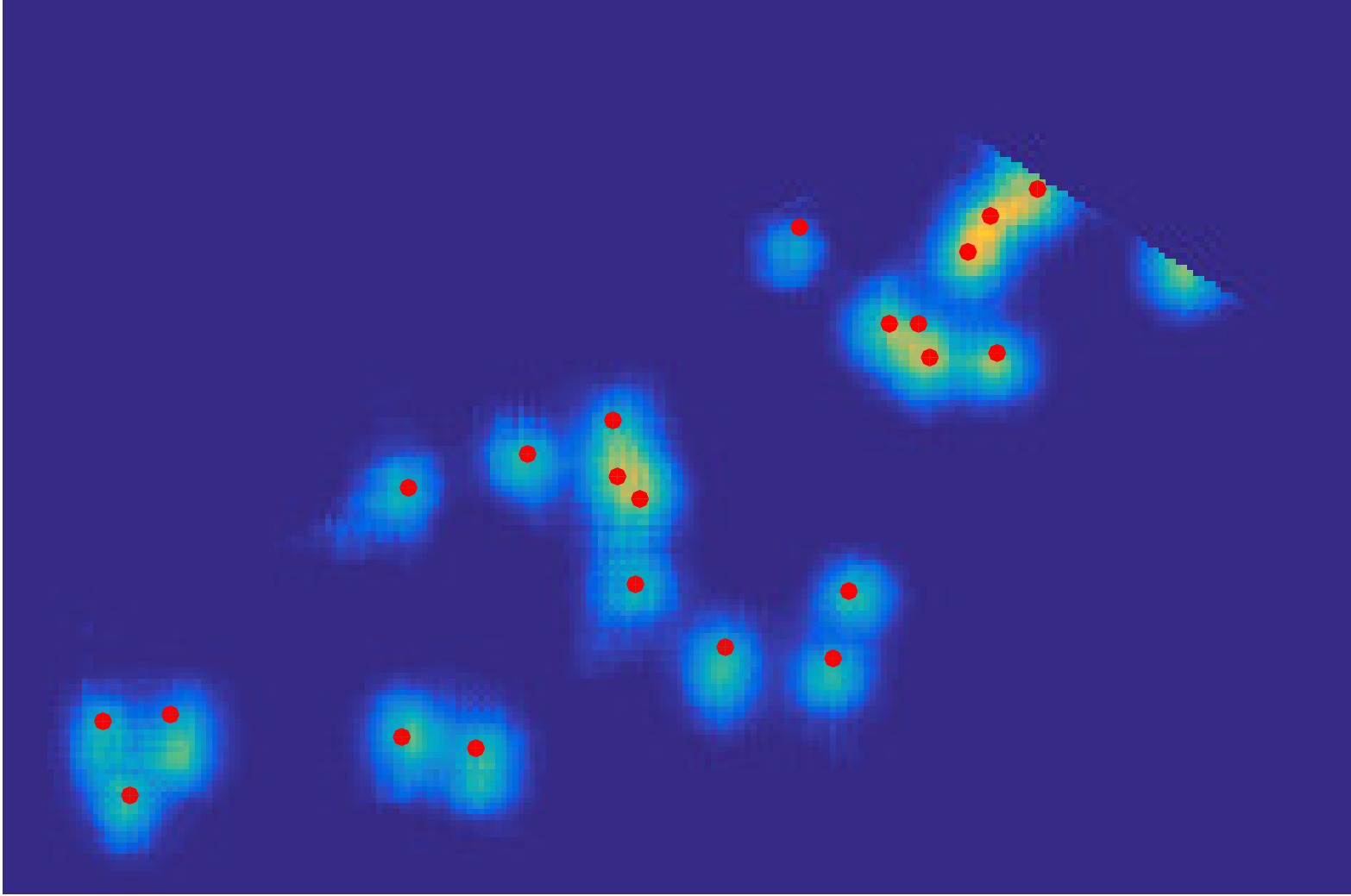} \\
\end{tabular}
\caption{Comparison of different density map methods.
All the density maps are in same color scale (a particular density value corresponds to the same color across images).
The green line in the image shows the region of interest (ROI).
The red dots are the ground-truth person annotations.
The density maps of \cite{Zhang2015,zhang2016single} are resized to the original resolution via bicubic interpolation.
}
\label{fig:dmap-comp}
\end{figure*}

\subsection{Density-based Counting}

The concept of an {\em object density map}, where the integral (sum) over any subregion equals the number of objects in that region, was first proposed in \cite{NIPS2010_4043}.
The density values are estimated from low-level features, thus sharing the advantages of general regression-based methods, while also maintaining location information.
\cite{NIPS2010_4043} uses a linear model to predict a pixel's density value from extracted features, and proposed the Maximum Excess over Sub Array (MESA) distance, which is the maximum counting error over all rectangular subregions, as a learning objective function.
\cite{Arteta2014} uses ridge regression (RR), instead of the computationally costly MESA, in their interactive counting system.
Both \cite{NIPS2010_4043} and \cite{Arteta2014} use random forest to extract features from several modalities, including the raw image, the difference image (with its previous frame), and the background-subtracted image.

\par
\cite{Fiaschi2012,Pham2015} used regression random forests, generated using the Frobenius norm as their criteria to obtain the best splits of their nodes.
\cite{Fiaschi2012} uses a number of standard filter bank responses (Laplacian of Gaussian, Gaussian gradient magnitude and eigenvalues of the structure tensor at different scales), along with the raw image, as their input of the regression random forest.
Other than the filter channels used in \cite{Fiaschi2012}, \cite{Pham2015} also uses the background subtraction result and the temporal derivative as their input of the regression random forest.
Unlike \cite{Fiaschi2012}, which directly regress the density patch, \cite{Pham2015} regresses the vector label pointing at the location of the objects within certain radius from the patch center, saving memory.
The density patch is finally generated from predicted vector labels.

\par
Among deep learning methods, CNN-patch \cite{Zhang2015}, Hydra CNN \cite{onoro2016towards} and CNN-boost \cite{Walach2016} use the input patch to predict a patch of density values, which is reshaped from the output of fully connected layer.
In contrast to \cite{Zhang2015}, Hydra CNN extracts patches of different sizes and scales them to a fixed size before feeding them to each head of the Hydra CNN, while CNN-boost uses a second and/or a third CNN to predict the residual error of earlier predictions.
\cite{Xie2015} uses a classical
FCNN to predict a density map for cell counting. In contrast, MCNN \cite{zhang2016single} is an FCNN with multiple feature-extraction columns, corresponding to different feature scales, whose output feature maps are concatenated together before feeding into later convolution layers. In contrast to Hydra CNN \cite{onoro2016towards}, the three columns in MCNN \cite{zhang2016single} use the same input patch but have different receptive field settings to encourage different columns to better capture objects of different sizes.

\par
Density map estimation methods differ in their choice of training loss function and form of prediction, which result in different characteristics of the predicted density maps. Both the loss and the prediction can be either pixel-wise or region-wise.
Table \ref{tab:strategy} summarizes the differences, and a visual comparison can be seen in \reffig{fig:dmap-comp}.
For the loss function, \cite{NIPS2010_4043} uses a region-based loss consisting of the Maximum Excess over Sub Arrays (MESA) distance, which is specifically designed for the counting problem, and aims to keep good counting performance over all sub-regions. However, for MESA, per-pixel reconstruction of the density map is not the first priority so that \cite{NIPS2010_4043} cannot well maintain the monotonicity around the peaks and spatial continuity in the predicted density maps (see \reffig{fig:dmap-comp}), although it is partially preserved due to neighboring pixels being assigned to the same feature codeword and thus same density value.
Most other methods \cite{Arteta2014,Fiaschi2012,Zhang2015,onoro2016towards,Walach2016,Xie2015,zhang2016single} use a pixel-wise loss function, e.g., the squared error between the predicted density value and the ground-truth.
While per-pixel loss does not optimize the counting error, it typically can yield good estimators of density maps for counting.

\par
For density map prediction, traditional methods in \cite{NIPS2010_4043,Arteta2014} choose a pixel-wise density prediction so as to obtain a full-resolution density map. The density map of the whole image is obtained by running the predictor over a sliding window in the image.
In contrast, most deep learning-based methods choose a patch-wise or image-wise prediction to speed up the prediction \cite{Zhang2015,Xie2015,onoro2016towards,zhang2016single,Walach2016}. Image-wise predictions using FCNNs \cite{Long2015,Xie2015,zhang2016single} are especially fast since they reuse computations.

\par
For patch-wise predictions, as in \cite{Zhang2015}, patches of density maps are predicted for overlapping image patches.
The density map of the whole image is obtained by placing the density patches at their image position, and then averaging pixel density values across overlapping patches.
The averaging process overcomes the double-counting problem of using overlapping patches to some extent. However, due to the lack of context information around the borders of the image patch, the density predictions around the corners of the neighboring patches are not always consistent with (as good as) those of the central patch.
The overlapping prediction and averaging operation can temper these artifacts, but also results in density maps that are overly smooth (e.g., see Fig.~\ref{fig:dmap-comp}e).

\par
The current CNNs using image- or patch-wise prediction normally only produce reduced-resolution density maps, due to either the convolution/pooling stride operation for FCNN-based methods, or to avoid very wide fully-connected layers for the patch-wise methods.
Accurate counting does not necessarily require original-resolution density maps, and using reduced-resolution maps in \cite{Zhang2015,zhang2016single} can make the predictions faster, while still achieving good counting performance.
On the other hand, accurate detection requires original resolution maps -- upsampling the reduced-resolution maps, in conjunction with averaging overlapping patches, sometimes results in an overly spread-out density map that cannot localize individual object well.
Considering these factors, our study will also consider full-resolution density maps produced with CNNs, in order to obtain a complete comparison of counting and localization tasks.

\subsection{Detection and Tracking with Object Density Maps}

Besides counting, \cite{Ma_2015_CVPR,Rodriguez2011} have also explored using density maps for detection and tracking problems in crowded scenes.
\cite{Ma_2015_CVPR} performs detection on density maps by first obtaining local counts from sliding windows over the density map from \cite{NIPS2010_4043}, and then uses integer programming to recover the location of individual objects.
In contrast to \cite{Ma_2015_CVPR}, our predicted density maps have clearer peaks, thus allowing for simpler methods for detection, such as weighted GMM clustering.

\cite{Rodriguez2011} uses density maps in a regularization term to improve  standard detectors and tracking.
In particular, a term is added to their objective function that encourages the density map generated from the detected locations to be similar to the predicted density map, so as to reduce the number of false positives and increase the recall.
The density maps estimated in \cite{Rodriguez2011} are predicted from the detector score map, rather than image features, resulting in spread-out density maps.
In contrast to \cite{Rodriguez2011}, we show that, when the density maps are compact and focused around the people, a simple fusion strategy can be used to combine the density map and the response map of a visual tracker (e.g., kernel correlation filter).

\section{Methodology} \label{framework}

\begin{figure}[tbp]
\begin{tabular}{c}
  CNN-pixel \\
  \includegraphics[width=0.46\textwidth]{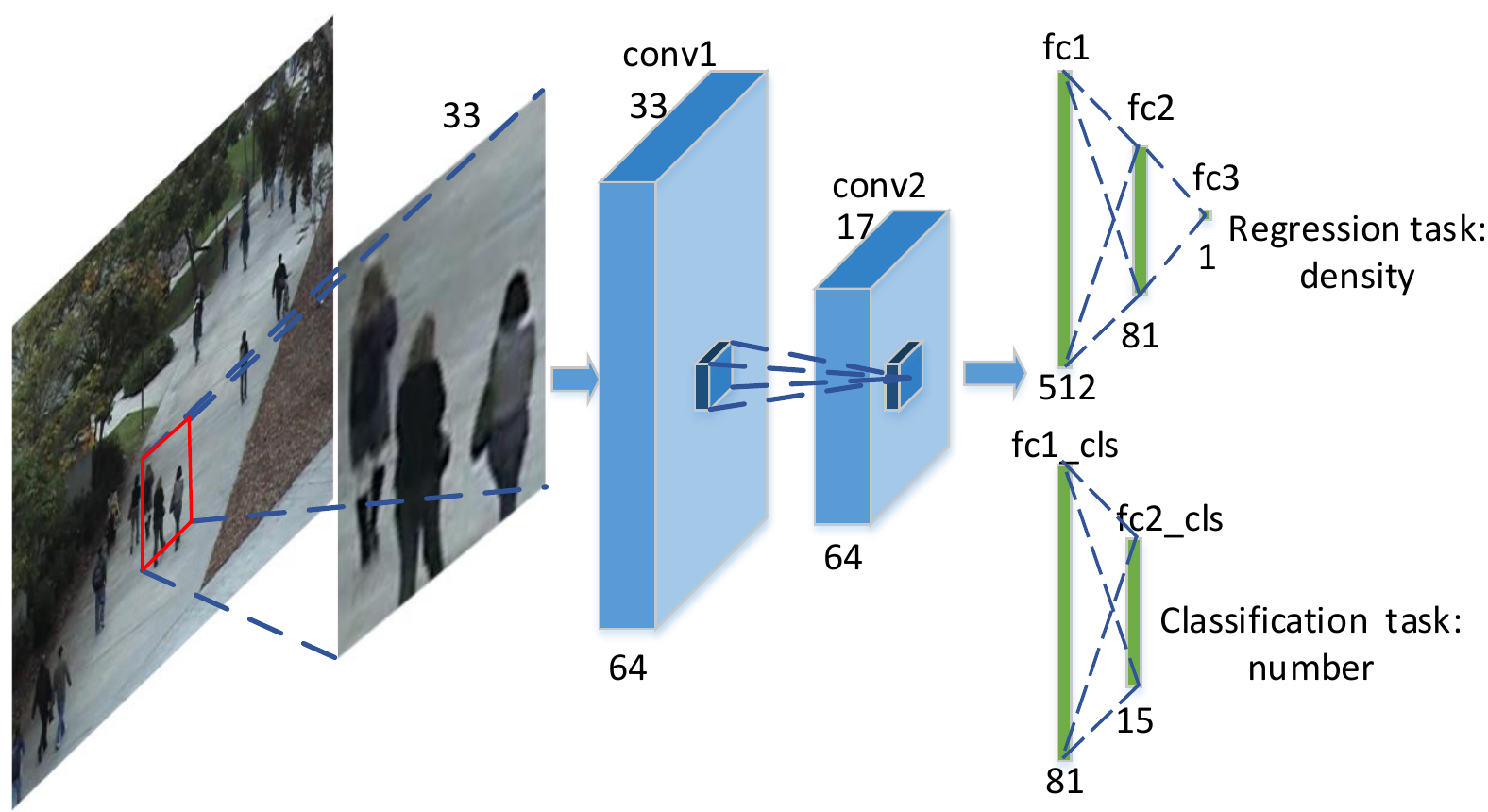} \\
  \hline
  FCNN-skip \\
  \includegraphics[width=0.46\textwidth]{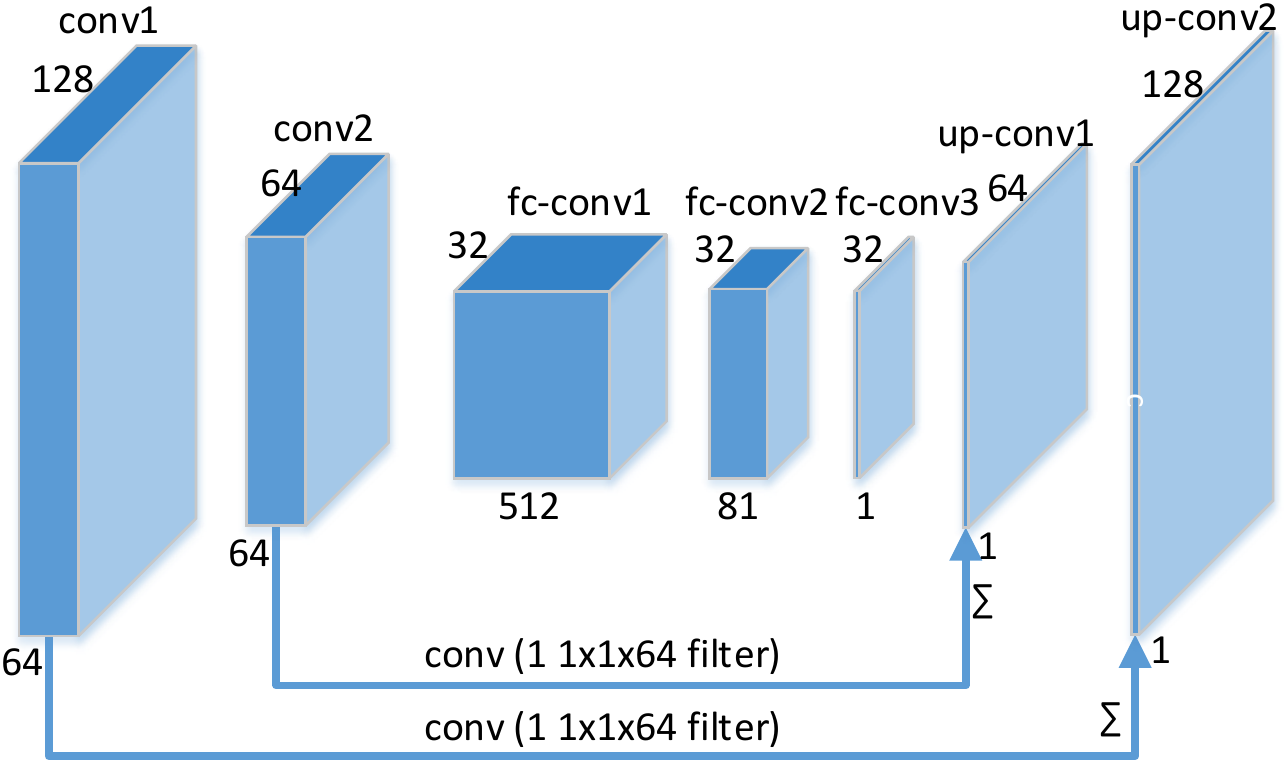}
\end{tabular}
\centering
\caption{CNN architectures for full-resolution density maps: (CNN-pixel) pixel-wise prediction using CNN; (FCNN-skip) fully-convolutional NN with skip branches.
Only layers with trainable weights are shown.
A multi-task framework is used for CNN-pixel, consisting of:
1) predicting the density value at the center of the image patch;
2) predicting the number of people in the image patch.
The two tasks share the same CNN feature extraction layers.}
\label{fig:arch}
\end{figure}

We consider two approaches for generating full-resolution density maps, as shown in Figure \ref{fig:arch}.
The first approach uses a classic CNN regressor for pixel-wise density prediction, i.e., given an image patch, predict the density at the center pixel.
The full-resolution density map is obtained using a sliding window to obtain density values for all pixels inside the ROI.
Although pixel-wise prediction does not explicitly model the relationship between neighboring pixels, it still results in smooth density maps -- the pooling operation in the CNN introduces translation invariance, and thus neighboring patches have similar features.
Thus, the pixel-wise predictions using CNNs will tend to be smooth and can better maintain the monotonicity, which will benefit localization tasks, such as detection and tracking.
In addition, due to the capability of CNNs to learn feature representations, density maps predicted by CNNs are less noisy and well localized around the objects, as compared to methods using handcrafted features (e.g., \cite{Arteta2014}).

The second approach uses a fully-convolutional NN to perform image-wise prediction.
Since the convolution/pooling stride operations in the lower-level convolutionl layers result in loss of spatial information (reduced resolution), subsequent upsampling and convolution layers are used to obtain a full-resolution density map.
The FCNN also inherits the smoothness property of pixel-wise prediction, but is more efficient in the prediction stage since it reuses computations from overlapping regions of neighboring patches.

\subsection{Pixel-wise Architecture}\label{text:CNN-pixel-arch}
Our network architecture for pixel-wise prediction (denoted as CNN-pixel) is shown in \reffig{fig:arch} (top).
The input for our network is 33$\times$33 image patch, while the output is the density value at the center pixel of the input patch.
The patch size is selected to be similar to the size of the largest person in the image.
Image feature extraction consists of two convolution-pooling layers, followed by 3 fully-connected layers to predict the density value.
A characteristic of the features extracted by CNNs is its hierarchical property: higher layers learn more abstract image features.
On the contrary, crowd density is a mid-level feature, which reflects the count and spatial distribution of objects, and hence high-level abstract features are not necessary.
Indeed, hand-crafted low-level and local features work very well in previous regression-based counting methods \cite{Chan2008,Chan2012,Ryan2009,NIPS2010_4043}.
Hence, instead of a very deep architecture (e.g., \cite{AlexNet,Szegedy_2015_CVPR,Simonyan14c,He2015}), we instead use an architecture that extracts mid-level features while keeping the model small and efficient.
\par
For training, a ground-truth density map is generated from the ``dot'' annotations of people in the training images following \cite{NIPS2010_4043}.
For training image ${\cal I}_n$, the density map is obtained by convolving the annotation map with a Gaussian (see \reffig{fig:dmap-comp}b),
\begin{align}
  D_n(p)=\sum_{P \in {\cal P}_n} {\cal N}(p; P, \sigma^2 I),
  \label{eq:gen_gt_density_map}
\end{align}
where $p$ denotes a pixel location, and ${\cal P}_n$ is the set of annotated positions for image ${\cal I}_n$.
${\cal N}(p;P,\sigma^2 I)$ is a Gaussian density with mean $P$ and isotropic covariance $\sigma^2I$.
For an image patch, its ground truth density value is the value of the ground-truth density map at the center of the patch.
We use the squared Euclidean distance as the loss function,
\begin{align}
  \ell_{\mathrm{density}}(d, \hat{d}) = (d - \hat{d})^2,
\end{align}
where $\{d, \hat{d}\}$ are the ground truth and predicted density values.

Similar to \cite{Sijin,li20143d,Zhang2015}, we introduce an auxiliary task, which shares the same CNN feature extraction layers of the primary regression task, in order to guide the network to find good image features and to make the training of the regression task less sensitive to weight initialization and the learning rate.
We choose a classification task as our auxiliary task because, empirically, classification is more robust to train than regression since it only needs to decide to which class the input belongs, rather than predicts an exact real number~\cite{Niu2016,Rothe2015}. Also, cross entropy loss for classification is less sensitive to outliers compared to L2 loss for regression~\cite{CS231n}.
The auxiliary task is a multi-class classification task where each class is the count of people within the image patch.
The ground-truth count for this auxiliary task is the sum of the dot annotations within the region of the image patch.
The categorical cross entropy is used as the loss function of the auxiliary task,
\begin{align}
  \ell_{\mathrm{aux}}(p, \hat{p}) = \sum_{i}{-p_i \log \hat{p}_i}
\end{align}
where $p_i$ is the true probability of class $i$ (i.e., 1 if the true class is $i$, and 0 otherwise), and $\hat{p}_i$ is the predicted probability of class $i$.
It should be noted that the classification loss $\ell_{\mathrm{aux}}$ may not be a natural choice for measuring count prediction error, since it ignores the difference between the prediction and the true count.
Nonetheless, this classification task is only used as an {\em auxiliary} task in the training stage.
At test time, the count is estimated as the sum of the predicted density map.
The regression and classification tasks are combined into a single weighted loss function for training,
\begin{align}
\ell = \lambda_1 \ell_{\mathrm{density}}(d,\hat{d}) + \lambda_2 \ell_{\mathrm{aux}}(p, \hat{p}),
\end{align}
where weights $\lambda_1=100$ and $\lambda_2=1$ in our implementation.

\par
Given a novel image, we obtain the density map by predicting the density values for all pixels using a sliding window over the image.

\subsection{Fully Convolutional Architecture}

Fully convolutional neural networks (FCNNs) \cite{Long2015,Chen2014,Xie2015,Torr2015} have gained popularity for semantic image segmentation and other tasks requiring dense prediction because they can efficiently predict whole segmentation maps by reusing computations from overlapping patches.
Efficiency can be further increased through up-sampling with a trainable up-sample filter.
\cite{Long2015,Chen2014} introduced ``skip branches'' to the FCNN, which add features from the lower convolutional layers to the up-sampled layers, in order to compensate for the loss of spatial information due to the stride in the convolution/pooling layers.

\par
Our CNN-pixel is adapted to a fully-convolutional architecture, and includes skip-branches (see \reffig{fig:arch} bottom, denoted as FCNN-skip) as in \cite{Long2015}.
In particular, the two pooling operation reduces the resolution of the density map by 4.
Two up-sampling layers are then used to obtain the density map at the original resolution.
Each up-sampling layer consists of two parts: each pixel is replicated to a 2$\times$2 region, and then a trainable 3$\times$3 convolutional layer is applied.
The skip branches pass the feature maps from a lower convolution stage through a convolution layer, which is then added to the density map produced by the upsampling-convolution layer.
The combination of features from different levels from a network has been shown to be helpful in better recovering lost spatial information \cite{Long2015,Chen2014}.
For semantic segmentation in \cite{Long2015}, the skip connections help to preserve the fine details of the segments.
For our counting task, by merging the low-level (but high resolution) features with the high-level (but low-resolution) features, the predicted high-resolution density map has more accurate per-pixel predictions than both pre-defined and learned interpolations. The skip-connections serve as a complementary part, which resembles residual learning, to refine the less accurate upsampled density map.

For FCNN training, we use both pixel-wise loss and patch-wise count loss,
\begin{align}
\ell_{\mathrm{pixel}} &= \sum_{i,j} (d_{i,j} - \hat{d}_{i,j})^2, \\
\ell_{\mathrm{count}} &= \left(\sum_{i,j}{d_{i,j}} - \sum_{i,j}{\hat{d}_{i,j}}\right)^2,
\end{align}
where $(i,j)$ index each pixel in one training patch.
The pixel-wise loss ensures that the predicted density map is a per-pixel reconstruction of the ground-truth map, while the count loss tunes the network for the final target of counting.
During training, we initialize the FCNN using a trained CNN-pixel model, and hence an auxiliary task is not needed.
In the prediction stage, we give the whole image as input and predict the density map with the same resolution. Similar to CNN-pixel, there are no block artifacts in the predicted density map (e.g., see \reffig{fig:dmap-comp}g and \reffig{fig:demo_TRANCOS}).

\subsection{Detection from Density Maps} \label{text:det}

Detecting objects directly from density maps was first proposed in \cite{Ma_2015_CVPR}.
In \cite{Ma_2015_CVPR}, a sliding window is passed over the density map to calculate the object count within each window.   \cite{Ma_2015_CVPR} used the density maps produced by MESA \cite{NIPS2010_4043}.
Recovering the object locations is then a deconvolution problem, which is implemented as a 2D integer programming problem since there can only be a nonnegative integer number of objects at each location.
\par
Besides integer programming, \cite{Ma_2015_CVPR} also proposed several simple baseline methods:
1) using non-maximum suppression to find local peaks in the density map;
2) using k-means clustering on the pixel locations with density value above a threshold (a simple form of crowd segmentation);
3) using Gaussian mixture model (GMM) clustering on the thresholded density map.
The latter two, k-means and GMM, only consider the shape of the crowd blobs and ignore the density values.
\par
Because the original-resolution density maps from CNN-pixel have well-defined peaks similar to the ground-truth density maps (see \reffig{fig:dmap-comp}h), we also propose a modification of GMM, where each pixel in the crowd segment is weighted according to its density value.
This weighting process makes the GMM cluster centers move towards the high density value region, which is more likely to be the real location of the object.
In other words, this weighting operation better tries to find local peaks.
In practice, we implement the weighting by discretizing the density map, and then repeating pixel locations according to their discretized density values, where locations with high density values appear more than once. GMM clustering is then applied to the samples.

\subsection{Improving Tracking with Density Maps} \label{text:tracking}

\begin{figure}[tb]
\centering
\includegraphics[width=0.44\textwidth]{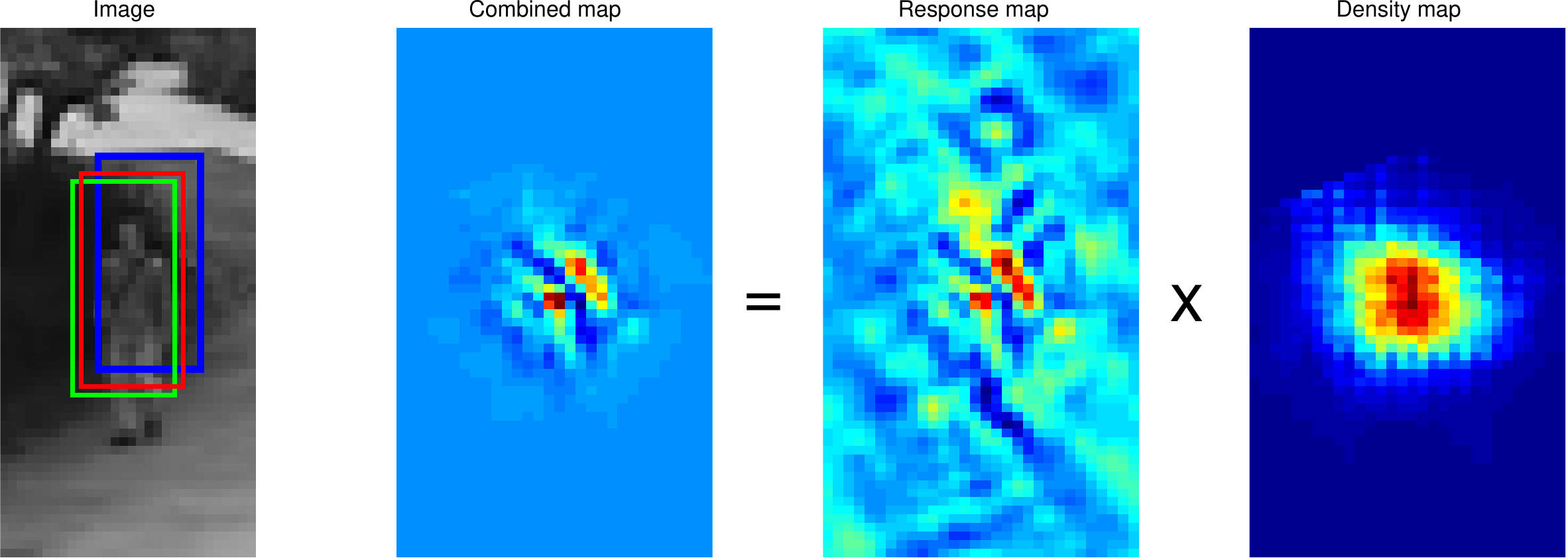}
\caption{An example of tracking by fusing the kernel correlation filter (KCF) response map and the crowd density map.
There are two peaks in the KCF response map, which causes the KCF tracker to drift (blue box).
Fusing the density map and the response map downweights the erroneous peak, and thus prevents tracker drifiting (green box).
The red box is ground truth bounding box.
}
\label{fig:tracking}
\end{figure}

We next describe how density maps can be used to improve state-of-the-art tracking of people in crowds.
Recently, kernelized correlation filters (KCF) has been widely used for single object tracking \cite{henriques2015high,henriques2012exploiting}.
However, tracking a person in a crowd is still challenging for KCF due to occlusion and background clutter, as well as low-resolution targets.
Here we combine the KCF tracker with the crowd density map so as to focus the tracker on image regions with high object density, which are more likely to contain the target.
This effectively prevents the tracker from drifting to the background.
In particular, our fusion strategy is to element-wise multiply the KCF response map with the corresponding crowd density map.
The maximum response of the combined map is then selected as the tracked position.
An example is shown in \reffig{fig:tracking}.

\section{Density Map Quality} \label{measure}

As high-quality per-pixel reproductions of the density map will likely yield better localization (detection and tracking) performance, in this section, we measure the quality of high-resolution density maps using various attributes.  We then relate these measures to localization performance in the next section.

We compare the following density map methods: MESA \cite{NIPS2010_4043}, ridge regression (RR) \cite{Arteta2014}, density-patch CNN (CNN-patch) \cite{Zhang2015}, multi-column CNN (MCNN) \cite{zhang2016single}, as well as full-resolution density maps produced by our pixel-wise CNN (CNN-pixel) and fully-convolutional NN (FCNN-skip).
We also test MCNN followed by upsampling-convolution layers to get full resolution density maps, denoted as MCNN-up.
All the predicted density maps are from UCSD ``max'' dataset (see Section~\ref{dataset_details}) and implementation details are in Section~\ref{text:implementation}.

\subsection{Per-pixel Reproduction}

\begin{figure}[tb]
\centering
\small
\begin{tabular}{@{}c@{\hspace{1mm}}c@{\hspace{1mm}}c@{}}
  (a) MESA \cite{NIPS2010_4043} & (b) Ridge regression \cite{Arteta2014} &
  (c) CNN-patch \cite{Zhang2015} \\
  \includegraphics[width=0.32\linewidth]{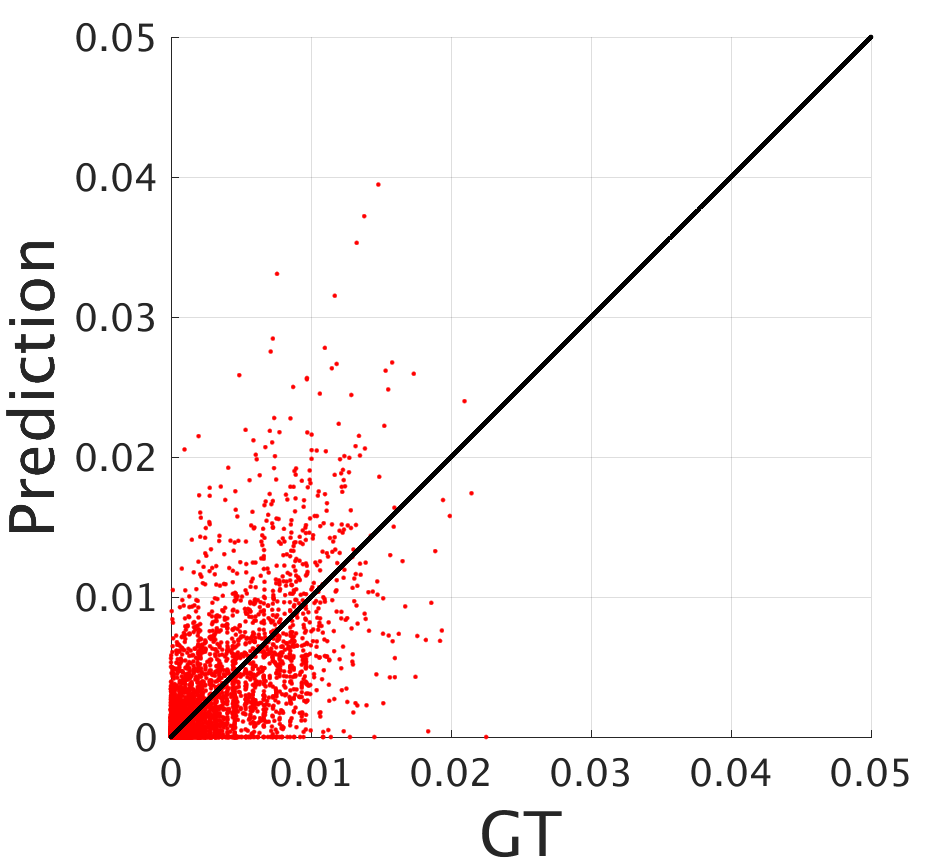} &
  \includegraphics[width=0.32\linewidth]{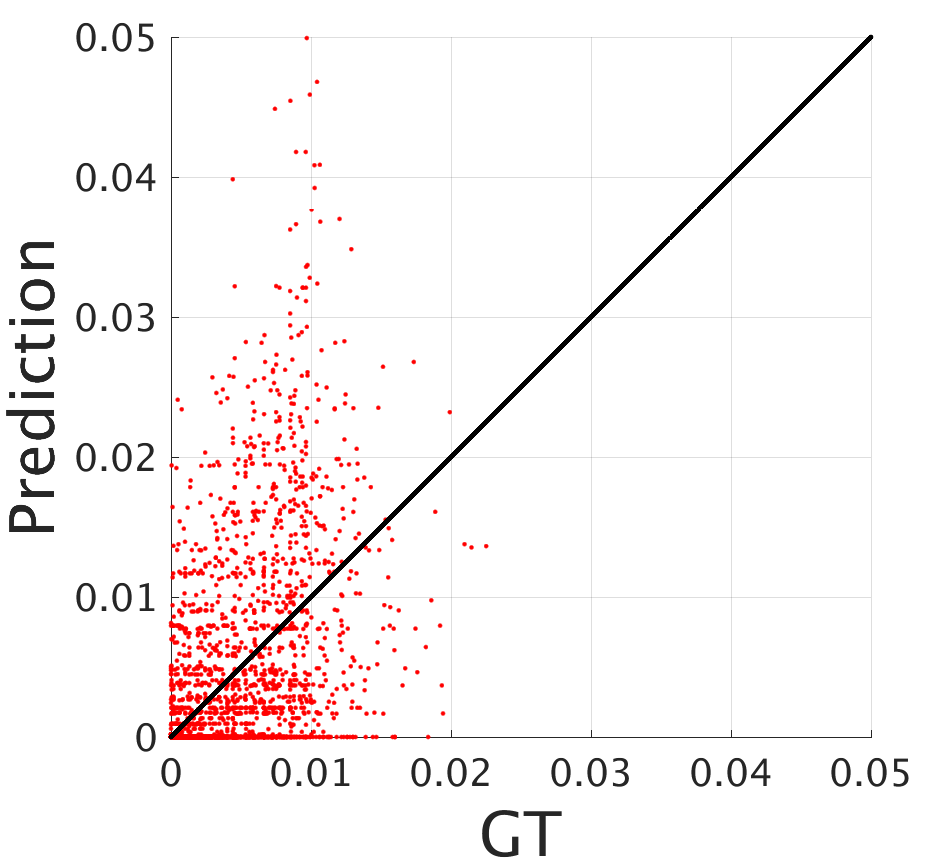} &
  \includegraphics[width=0.32\linewidth]{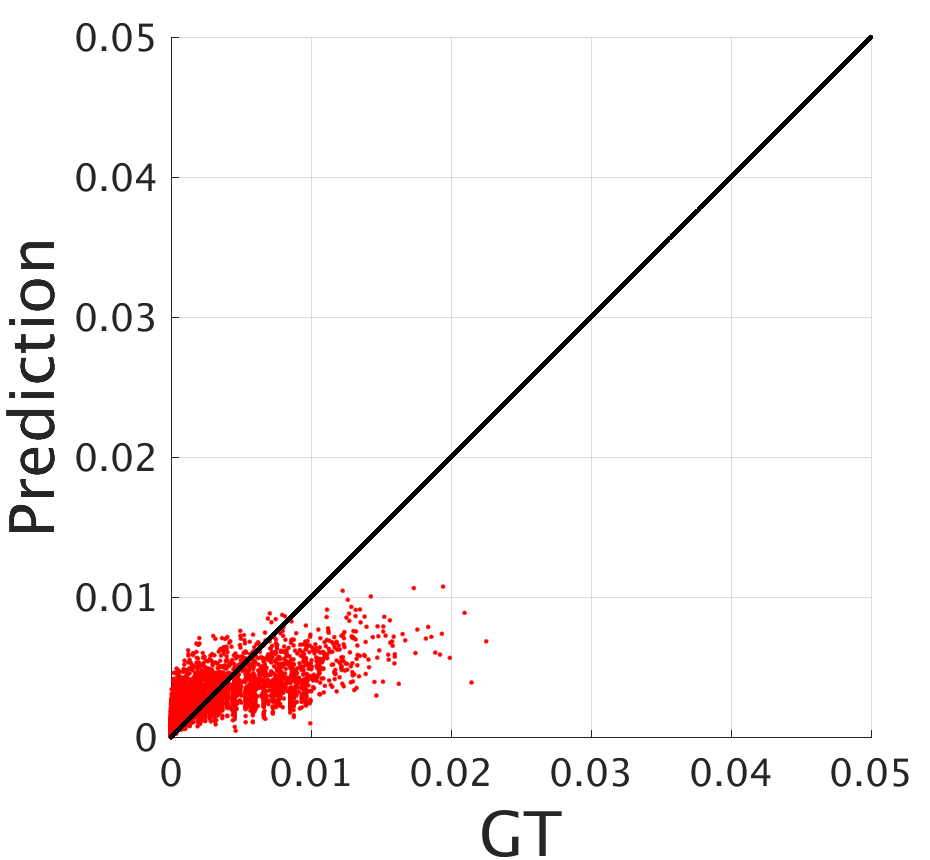} \\
  (d) MCNN \cite{zhang2016single} & (e) FCNN-skip (ours) & (f) CNN-pixel (ours) \\
  \includegraphics[width=0.32\linewidth]{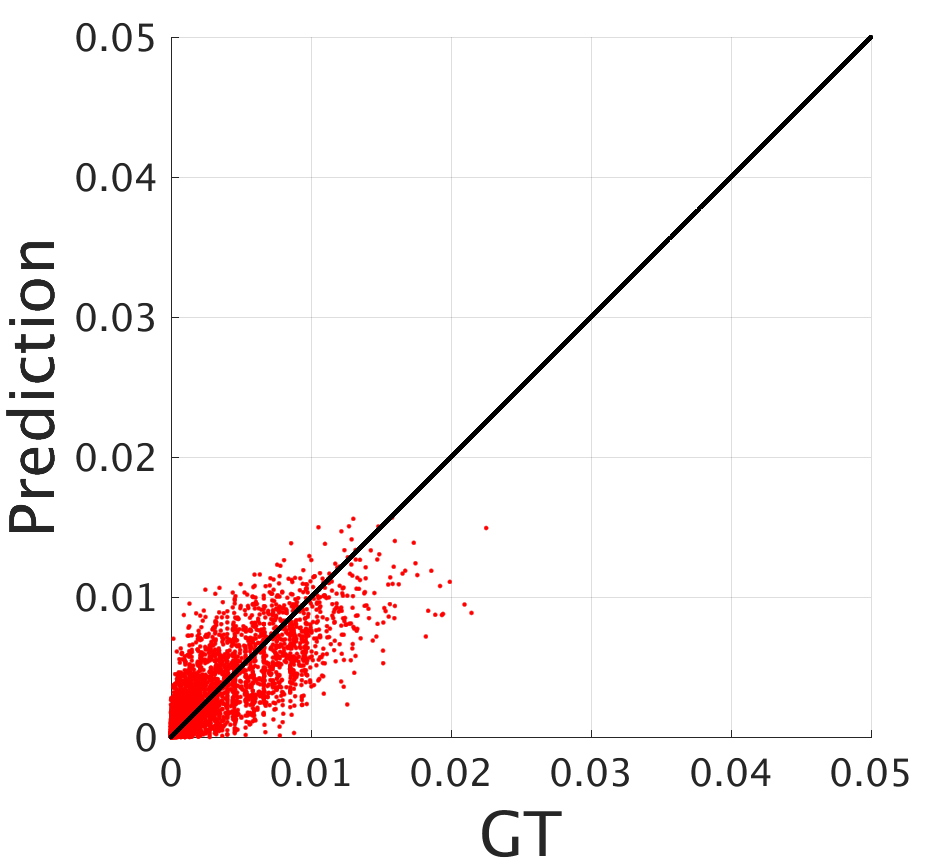} &
  \includegraphics[width=0.32\linewidth]{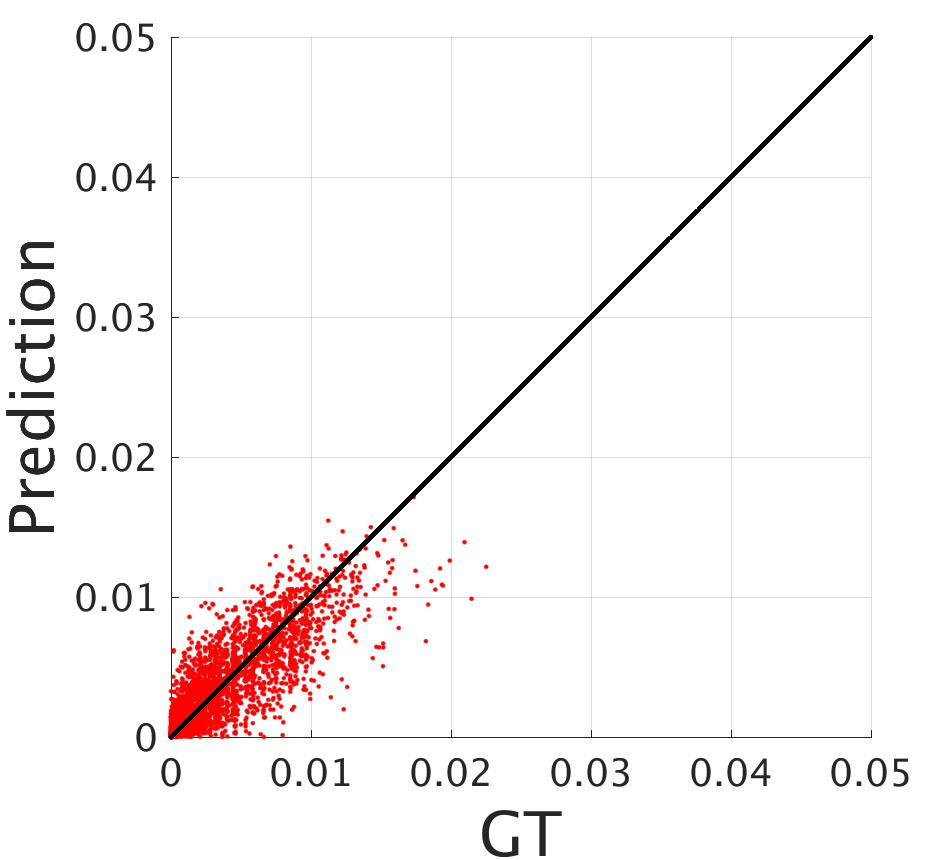} &
  \includegraphics[width=0.32\linewidth]{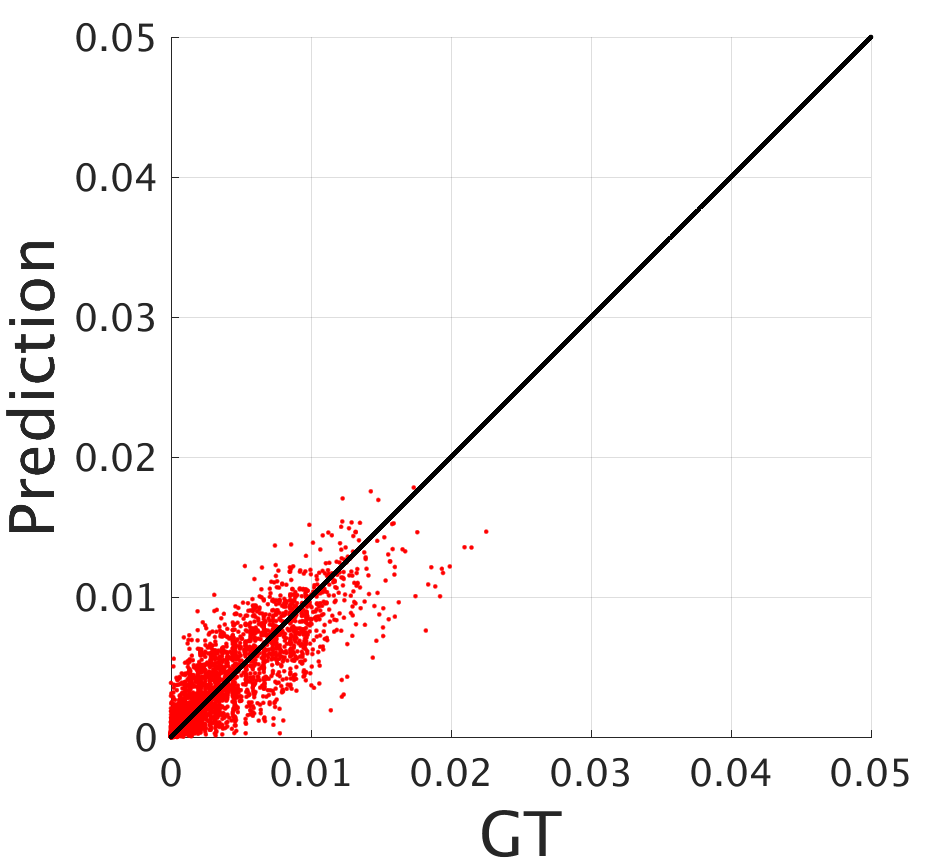} \\
\end{tabular}
\caption{Scatter plots of GT density and predicted density values.}
\label{fig:scatter_plot}
\end{figure}

We first consider how well each method can reproduce the per-pixel values in the ground-truth density map.
\reffig{fig:scatter_plot} shows a scatter plot between the ground-truth density pixel values and the predicted density pixel values for the various methods.
MCNN, CNN-pixel and FCNN-skip show the best correlation between ground-truth and prediction, and are more concentrated around the diagonal.
Ridge regression (RR) tends to over-predict the density values, because the filtering of negative feature weights compacts the density map causing higher density values.
On the other hand, CNN-patch \cite{Zhang2015} under-predicts the values, because the low-resolution predictions and patch-wise averaging spreads out the density map causing lower overall density values.
MESA \cite{NIPS2010_4043} predictions are also concentrated around the diagonal, but less so than MCNN, FCNN-skip, and CNN-pixel.
This is because MESA optimizes the count prediction error within rectangular sub-regions of the density map, rather than per-pixel error.

\subsection{Compactness and Localization} \label{text:compactlocal}

Compactness and localization are two important properties for using density maps for detection and tracking.
Compactness means the density values are concentrated in tight regions around a particular position, while localization means that these positions are located near a ground-truth dot annotation.

\par
To measure the compactness and localization, we place a bounding box (scaled for perspective) over each dot annotation (ground-truth position).
If a density map is not compact, then some density will leak outside the bounding boxes.
The amount of leakage can be measured by calculating the ratio of density inside the bounding boxes to the total density in the image, which is denoted as {\em bounding box density ratio} (BBDR).
If the prediction is localized well, then the total predicted density inside the bounding boxes should match the corresponding total ground-truth density. Hence, the measure of localization is the MAE inside all boxes, denoted as {\em bounding box MAE} (BBMAE).

\par
\begin{figure}[tbp]
\centering
\footnotesize
\begin{tabular}{@{}c@{}c@{}}
  (a)  &  (b)  \\
  \includegraphics[width=0.25\textwidth]{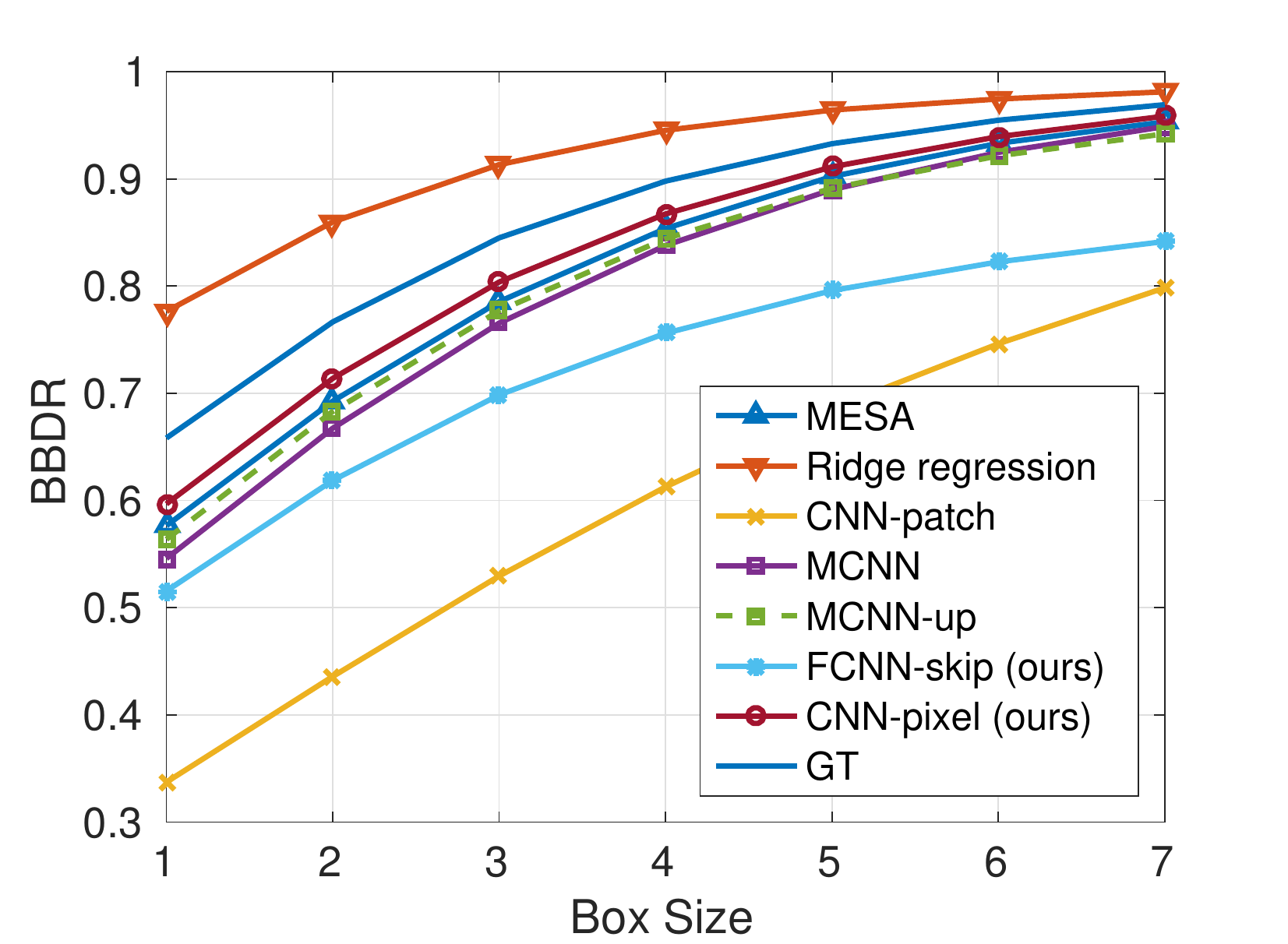} &
  \includegraphics[width=0.25\textwidth]{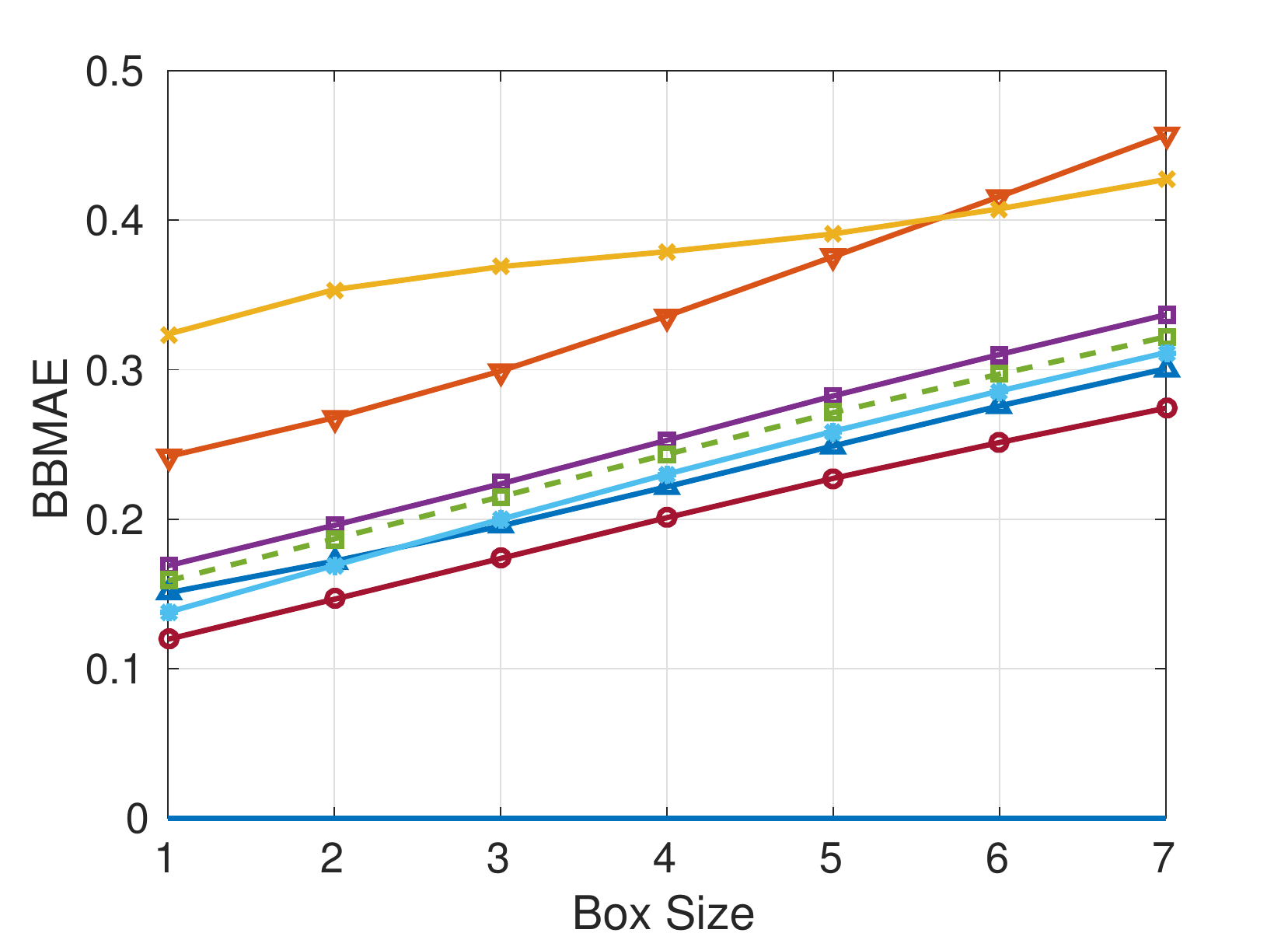}
\end{tabular}
\caption{Measures of compactness and localization:
(a) bounding box density ratio (BBDR). Higher values means more compact, e.g. ridge regression prediction is the most compact.
(b) bounding box MAE (BBMAE). Lower values mean better localization.}
\label{fig:bbox}
\end{figure}

Figs.~\ref{fig:bbox}a and \ref{fig:bbox}b plot the curves for BBDR and BBMAE by varying the size of the bounding box.
CNN-pixel has the best localization measure (lowest BBMAE), and also has similar compactness (BBDR)  to the ground-truth.
FCNN-skip also has good localization (low BBMAE), but, in contrast, its density maps are more spread out (low BBDR, less compact) due to its upsampling operation.
MCNN-up has slightly higher BBDR and lower BBMAE compared with MCNN, which suggests that the learned upsampling layer can improve compactness and localization, over bicubic upsampling.
RR has the most compact density maps (BBDR even higher than the ground-truth density used for training),
but has poor localization (high BBMAE) because the centroid of the local modes are shifted.

\begin{table*}[tbhp]
\centering
\small
  \begin{tabular}{|c|c||ccc||ccc|}
    \cline{3-8}
    \multicolumn{2}{c||}{}
    & \multicolumn{3}{|c||}{IntProg}
    & \multicolumn{3}{|c|}{GMM-weighted} \\
    \hline
    Density maps
      & MAD
    & ED   & EDD  & Miss rate
    & ED  & EDD    & Miss rate \\
    \hline
    MESA \cite{NIPS2010_4043}
    & 16.34
      & 4.96$\pm$0.71 & 4.89$\pm$0.92 & 13.96\%
    & 2.90$\pm$0.74 & 2.63$\pm$0.81  & 8.69\% \\
    Ridge regression \cite{Arteta2014}
    & 18.34
      & 4.48$\pm$0.69 &  4.45$\pm$0.95& 15.87\%
    & 2.91$\pm$0.81 & 2.33$\pm$0.76 & 13.14\% \\
    \hline
    CNN-patch \cite{Zhang2015}
      & {\bf 1.89}
      & 5.40$\pm$0.74 & 4.23$\pm$1.17 & 41.03\%
    & 4.84$\pm$0.91  &4.50$\pm$1.45& 16.66\% \\
    MCNN \cite{zhang2016single}
      & 3.96
      & 4.22$\pm$0.76 & 3.32$\pm$0.90  &  19.77\%
    & 3.56$\pm$0.76 & 2.24$\pm$0.79 & 10.53\% \\
    MCNN-up
      & 3.96
      & 4.13$\pm$0.89 & 3.12$\pm$1.03  &  16.13\%
    & 3.52$\pm$0.84 & 2.03$\pm$0.70 & 10.54\% \\
    CNN-pixel (ours)
      & 3.96
      & {\bf 3.61}$\pm$0.72 & {\bf 2.90}$\pm$0.83  &  9.82\%
    & 2.78$\pm$0.85 & {\bf 1.88}$\pm$0.75 & \bf{8.43}\% \\
    FCNN-skip (ours)
      & 4.78
      & {\bf 3.61}$\pm$0.74 & 3.38$\pm$1.01  &  13.34\%
    & {\bf 2.73}$\pm$0.76 & 2.26$\pm$0.84 & 10.21\% \\
    \hline
    Ground truth
      & 3.31
      & 3.06$\pm$0.68 & 2.30$\pm$1.22 & 2.59\%
    & 1.29$\pm$1.01 &  1.18$\pm$0.93 & 5.24\% \\
    \hline
  \end{tabular}
\caption{Comparisons of temporal smoothness of density maps and detections.}
\label{tab:det_combo}
\end{table*}

\subsection{Temporal Smoothness}

We measure temporal smoothness of the density maps and the detection results.
Density maps and detections that are smooth in time suggest that finding each person's trajectory should be easier.
Hence, these temporally smooth density maps could aid object tracking.
Table~\ref{tab:det_combo} shows the mean absolute difference (MAD) between density maps of consecutive frames.
The temporal smoothness of deep learning methods are better than traditional methods even though they do not directly use frame difference information.
This shows that the CNN feature extractor produces smoothly varying feature representations, compared to the random forest representation used with MESA/RR.
The density maps of CNN-patch are the most temporally smooth (lowest MAD), but this is because those maps are more spread out (less compact), which cannot help to localize and track the object (refer to Section \ref{tracking_exps}).

\par
To measure the temporal smoothness of the detections, we calculate the error vector between a detection location and its corresponding ground-truth.
The error distance (ED) measures the distance to the ground-truth, and is equal to the length of the error vector.
We also measure the length of the difference between two consecutive error vectors corresponding to the same ground-truth trajectory, and denote it as EDD (error difference distance).
EDD measures the stability of the detected point over time, when compared to the GT trajectory.
Finally, we measure the miss rate, which is the percentage of missed detections.
Table~\ref{tab:det_combo} shows results for IntProg and GMM-weighted.
With IntProg, CNN-pixel and FCNN-skip have similar ED, and CNN-pixel has the lowest EDD, which indicates that the detected points are more stable over time for CNN-pixel.
Using GMM-weighted, CNN-pixel and FCNN-skip obtain the most accurate detected positions (lowest ED), but CNN-pixel is more temporally stable.
These results suggest that temporally smooth density maps, produced using per-pixel prediction, could be helpful for object tracking, e.g., by incorporating them into a tracking-by-detection framework.

\subsection{Summary}

In summary, CNN-pixel performs best on various metrics, including compactness (BBDR), localization (BBMAE), and temporal smoothness (EDD).
In contrast, reduced-resolution maps that require either fixed upsampling (CNN-patch and MCNN) or learned upsampling-convolution (FCNN-skip and MCNN-up) had worse quality metrics.
The downsampling operations in the CNN obfuscate the true position of the people in the reduced-resolution maps.  This loss of spatial information can only be partially compensated using learned upsampling and skip connections.

\section{Experiments} \label{experiments}

In this section we present experiments on several crowd analysis tasks to compare density maps estimated by recent methods, including MESA \cite{NIPS2010_4043}, ridge regression (RR) \cite{Arteta2014}, Regression forest \cite{Fiaschi2012}, COUNT forest \cite{Pham2015}, density-patch CNN (denoted as CNN-patch) \cite{Zhang2015}, multi-column CNN (MCNN) \cite{zhang2016single}, Hydra CNN \cite{onoro2016towards} and CNN-boost \cite{Walach2016},
as well as full-resolution density maps produced by our CNN-pixel and FCNN-skip.
We first present experiments on crowd counting using benchmark datasets for people and cars. We then present experiments on people detection using density maps, to compare the localization ability of the density maps.  Next we conduct experiments on people tracking by fusing the density maps and visual tracker response, in order to compare the localization ability and the temporal smoothness of the density maps. Finally, we perform an ablation study on CNN-pixel.

\subsection{Implementation Details}
\label{text:implementation}
The CNN-pixel architecture appears in \reffig{fig:arch}.
The input image patch is a 33$\times$33 grayscale image.
Every convolution layer is followed by an across channel local response normalization layer and a max pooling layer.
The first convolutional layer contains 64 5$\times$5$\times$1 filters (i.e., 5$\times$5 filter on 1 grayscale channel), and the max-pooling is 2$\times$2 with a stride size of 2$\times$2.
The second convolutional layer contains 64 5$\times$5$\times$64 filters, and the max-pooling is  3$\times$3 pooling with a stride size of 2$\times$2.
When the images are high resolution and contain large objects, as in the TRANCOS dataset \cite{guerrero2015extremely} (640$\times$480) and WorldExpo'10 dataset \cite{Zhang2015} (720$\times$576), we use a 65$\times$65 image patch and add one additional convolution-pooling layer to capture enough context information, and to keep the final output feature map a consistent size.
The CNN-pixel is trained using the regression and classification tasks, as described in Section \ref{text:CNN-pixel-arch}.
The training set is constructed by densely extracting all gray-scale patches, whose centers are inside the ROI, from the training images.
The network is trained with standard back-propagation \cite{rumelhart1988learning}.
Momentum and weight decay are used, and one dropout layer is used after every non-output fully connected layer (refer to \reffig{fig:arch}) to reduce over-fitting.

\par
Our FCNN-skip is adapted from the CNN-pixel model.
The structure of the lower convolutional layers and pooling layers remains the same as CNN-pixel.
As in \cite{Long2015}, the first fully connected layer is adapted to a convolutional layer
whose filter size is the same as the output feature map size produced by our CNN-pixel model (9$\times$9). The remaining fully connected layers are implemented as convolutional layers using a filter size of 1$\times$1.
For the skip-branches, a convolutional layer with 1$\times$1$\times$64 filters is applied to the lower convolutional layer, and then merged with the density map using an element-wise sum.

\par
The training of the FCNN-skip network is initialized with the weights from CNN-pixel network.
The weights of the convolutional layer after up-sampling are initialized as an average filter.
The weights of the skip branches are initialized as zeros (as in \cite{Long2015}),
so that the training first focuses on tuning the initial weights (from CNN-pixel) to the FCNN architecture.

\par
In the inference stage, CNN-pixel densely extracts test patches using the training patch size, and predicts the density values in a pixel-wise manner.
FCNN-skip takes any size image as input and predicts the whole density map.
Both CNN-pixel and FCNN-skip density maps do not have block artifacts caused by combining overlapping patch predictions.
The count in the region-of-interest of an image is the sum of the predicted density map inside the region.
We do not apply any post-processing to the estimated count, cf., the ridge regression used in \cite{Zhang2015} to remove systematic count errors.
Finally, note that we do not change the input patch size according to its location in the scene,
which is a form of perspective normalization.

\subsection{Counting Experiments} \label{dataset_details}

We first compare the density estmation methods on three crowd datasets, UCSD pedestrian, WorldExpo'10, and UCF\_CC\_50, and a car dataset, TRANCOS.
The statistics of the datasets are summarized in Table \ref{tab:datasets}, and example images can be found in the supplementary.

\begin{table}[tbhp]
\centering
\small
\begin{tabular}{|c|c|c|c|c|}
  \hline
  Dataset                          & $N_f$    & Res.            & Range   & $T_p$
  \\
  \hline
  UCSD \cite{Chan2008}             & 2000     & 238$\times$158  & 11-45   & 49885
  \\
  WorldExpo'10 \cite{Zhang2015}    & 3980     & 720$\times$576  & 1-253   & 199923
  \\
  UCF\_CC\_50 \cite{Idrees2013}    & 50       & varies          & 96-4633 & 63974
  \\
  TRANCOS \cite{guerrero2015extremely}    & 1244     & 640$\times$480  & 9-107   & 46796
  \\
  \hline
\end{tabular}
\caption{Statistics of the four tested datasets.
$N_f$ is the number of annotated images or frames;
Res.~is the image/frame resolution;
Range is the range of number of objects inside the ROI of a frame;
$T_p$ is the total number of labeled objects.}
\label{tab:datasets}
\end{table}

\subsubsection{Datasets}
\par
The UCSD dataset is a low resolution (238$\times$158) surveillance video with perspective change and heavy occlusion between objects.
It contains 2000 video frames and 49,885 annotated pedestrians.
On UCSD, we test the performance on the traditional setting \cite{Chan2008} and four down-sampled splits \cite{Ryan2009}.
In the first protocol \cite{Chan2008}, all frames from 601-1400 are used for model training and the remaining 1200 frames are for testing.
The second protocol is based on four downsampled training splits \cite{Ryan2009}: ``maximal'' (160 frames, 601:5:1400), ``downscale'' (80 frames, 1201:5:1600), ``upscale'' (60 frames, 801:5:1100), and ``minimal'' (10 frames, 641:80:1361).
These splits test robustness to the amount of training data (``maximal'' vs.~``minimal''), and generalization across crowd levels (e.g., ``upscale'' trains on small crowds, and tests on large crowds).
The corresponding test sets cover all the frames outside the training range (e.g. 1:600 and 1401:2000 for ``maximal'').
The ground-truth density maps use $\sigma=4$.

\par
The WorldExpo'10 dataset \cite{Zhang2015} is a large scale people counting dataset captured from the Shanghai 2010 WorldExpo.
It contains 3,980 annotated images and 199,923 annotated pedestrians from 108 different scenes.
Following \cite{Zhang2015}, we use a human-like density template, which changes with perspective, to generate the ground-truth density map, and the fractional density value in the ROI is used as the ground-truth since there are many people near to the ROI boundary.
Models are trained on the 103 training scenes and tested on 5 novel test scenes.

\par
The UCF dataset \cite{Idrees2013} contains 50 very different images with the number of people varying from 96 to 4,633.
In contrast to the previous two datasets, UCF only contains one image for each scene, and each image has a different resolution and camera angle.
This dataset measures the accuracy, scalability and practicality of counting methods, as well as tests how well methods handle extreme situations.
The ground-truth density map is generated using $\sigma=6$.
Following \cite{Idrees2013,Zhang2015}, the evaluation is based on 5-fold cross-validation.

\par
The TRANCOS dataset \cite{guerrero2015extremely} is a benchmark for vehicle counting in traffic congestion situations, where partial-occlusion frequently occurs.
It has training, validation and test sets, containing 1,244 annotated images from different scenes and 46,796 annotated vehicles.
Following the first training strategy in \cite{guerrero2015extremely}, we use the training and validation data as the training and validation sets for model training, and evaluate the model on the test set.
Different from the WorldExpo and UCF datasets, TRANCOS dataset contains multiple scenes but the same scenes appear in  the training, validation, and test sets.
The ground-truth density map is generated using $\sigma=10$.

\par
Counting predictions are evaluated using the mean absolute error (MAE) or mean square error (MSE) with respect to the ground truth number of people. The fractional number is used for WorldExpo and integer number is used for the other datasets, following the convention.

\subsubsection{UCSD pedestrian dataset}

We present results on the two experimental protocols for UCSD.
The results for the first protocol (full training set) are shown in Table~\ref{tab:ucsd_800},
where the last five CNN methods are based on density maps, and the remaining are traditional regression-based methods.
The CNN-based density estimation methods have lower error than the traditional methods, especially the last four methods which reduce error by a large margin.
MCNN has slightly lower MAE than CNN-boost and CNN-pixel (1.07 vs 1.10 and 1.12).
FCNN-skip method achieves worse performance than CNN-pixel (1.22 vs 1.12) but consumes much less time in the prediction stage (16 ms per frame vs.~4.6 sec per frame on UCSD).
In comparison, MCNN takes 31 ms and CNN-patch takes 37 ms per frame respectively.

\begin{table}[tbp]
\centering
\small
\begin{tabular}{|c|c|c|}
  \hline
  Method                                          & MAE   & MSE  \\
  \hline
  Kernel Ridge Regression \cite{an2007face}       & 2.16  & 7.45 \\
  Ridge Regression \cite{Chen2012}                & 2.25  & 7.82 \\
  Gaussian Process Regression \cite{Chan2008}     & 2.24  & 7.97  \\
  Cumulative Attribute Regression \cite{Chen2013} & 2.07  & 6.86 \\
  COUNT forest \cite{Pham2015}                    & 1.61  & 4.4 \\
  \hline
  CNN-patch+RR \cite{Zhang2015}                   & 1.60  & 3.31  \\
  MCNN \cite{zhang2016single}                     & \bf{1.07}  & \bf{1.35}  \\
  CNN-boost fine-tuned using 1 boost \cite{onoro2016towards}   & 1.10  & - \\
  CNN-pixel (ours)                                & 1.12  & 2.06 \\
  FCNN-skip (ours)                                & 1.22  & 2.25 \\
  \hline
\end{tabular}
\caption{Test errors on the UCSD dataset when using the whole training set.}
\label{tab:ucsd_800}
\end{table}

\begin{table}[tbhp]
\centering
\resizebox{.48\textwidth}{!}{
  \begin{tabular}{|c|c|c|c|c||c|}
    \hline
    Method                              & Max  & Down  & Up   & Min & Avg \\
    \hline
    MESA \cite{NIPS2010_4043}           & 1.70  & 1.28  & 1.59 & 2.02 & 1.65 \\
    Regression forest \cite{Fiaschi2012}& 1.70  & 2.16  & 1.61 & 2.20 & 1.92 \\
    Ridge regression \cite{Arteta2014}  & \bf{1.24}   & 1.31  & 1.69  & \bf{1.49} & 1.43 \\
    COUNT forest \cite{Pham2015}        & 1.43  & 1.30  & 1.59  & 1.62   & 1.49  \\
    \hline
    CNN-patch+RR \cite{Zhang2015}       & 1.70  & \bf{1.26}  & 1.59 & 1.52 & 1.52 \\
    MCNN \cite{zhang2016single}         & 1.32  & 1.71  & 2.05  & 1.56  & 1.66 \\
    CCNN \cite{onoro2016towards}        & 1.65  & 1.79  & \bf{1.11}  & 1.50  & 1.51 \\
    Hydra 2s \cite{onoro2016towards}    & 2.22  & 1.93  & 1.37  & 2.38  & 1.98 \\
    CNN-pixel (ours)                    & 1.26  & 1.35  & 1.59 & \bf{1.49} & {\bf 1.42} \\
    FCNN-skip (ours)                    & \bf{1.24}  & 1.38  & 2.13  & 1.83   & 1.65 \\
    \hline
    CNN-pixel-VS                 & 1.48  & ---   & ---  & ---       & --- \\
    \hline
  \end{tabular}
}
\caption{Comparison of MAE between density-based counting methods on the UCSD dataset using 4 training splits.}
\label{tab:ucsd_count}
\end{table}

Table~\ref{tab:ucsd_count} shows the MAE for the four training splits of the second protocol.
Here we only list results of density-based approaches, which represent the current state-of-the-art counting performance.
Over the four splits, CNN-pixel and RR have the lowest average MAE.
MCNN and FCNN-skip, which are both FCNNs, performs a little worse, mainly on the ``up'' split, than CNN-patch and CNN-pixel but are more computationally efficient.
Examining the error heat maps in the supplementary shows that the FCNN-skip errors are overall larger on the ``up'' split, and without any systematic bias.

\begin{figure}[tbp]
\centering
\small
\begin{tabular}{@{}c@{\hspace{1mm}}c@{\hspace{1mm}}c@{}}
  (a) Image    & (b) Ground truth & (c) Prediction \\
  \includegraphics[width=0.16\textwidth]{fig_2_a-demo_image} &
  \includegraphics[width=0.16\textwidth]{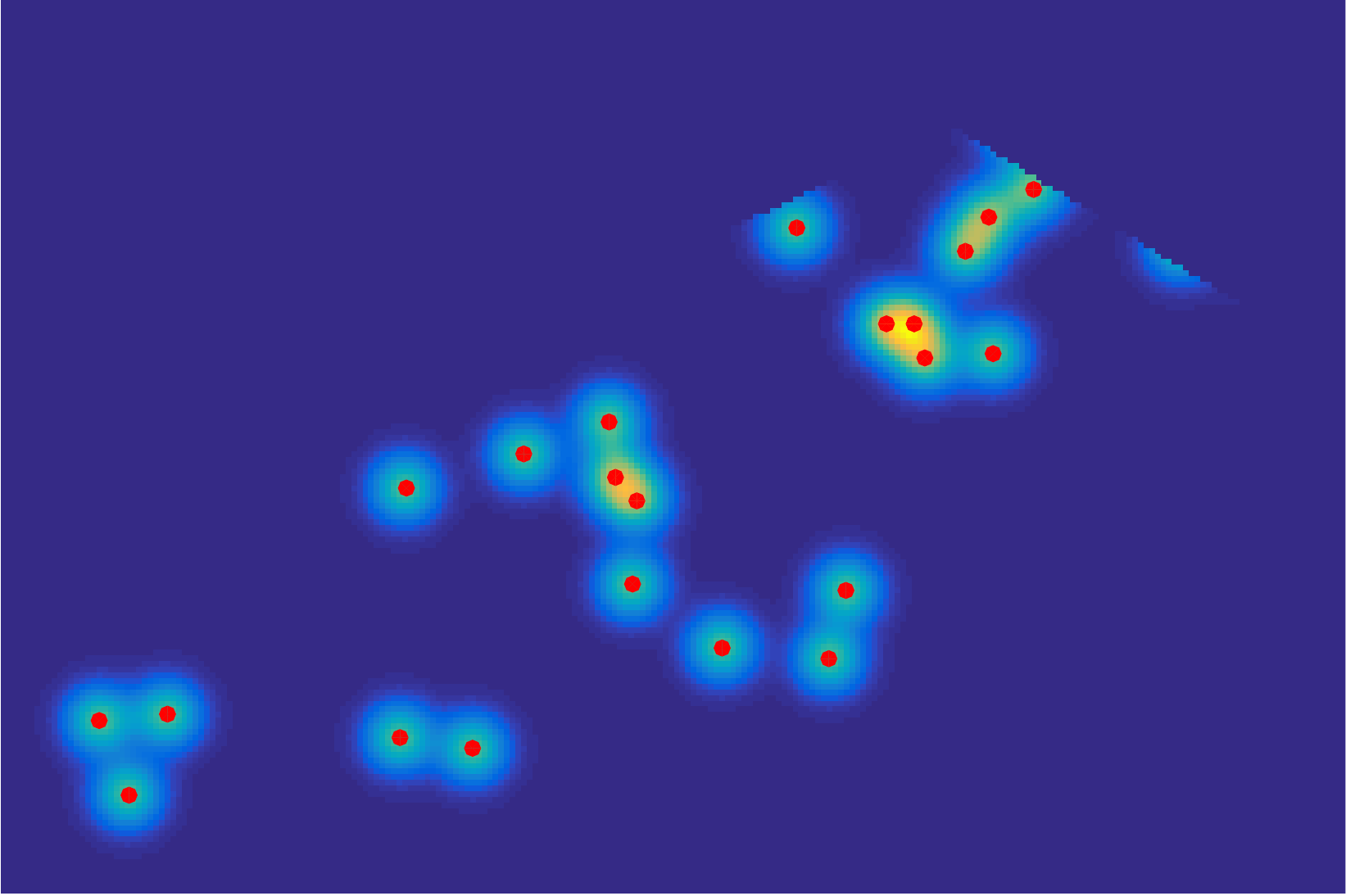} &
  \includegraphics[width=0.16\textwidth]{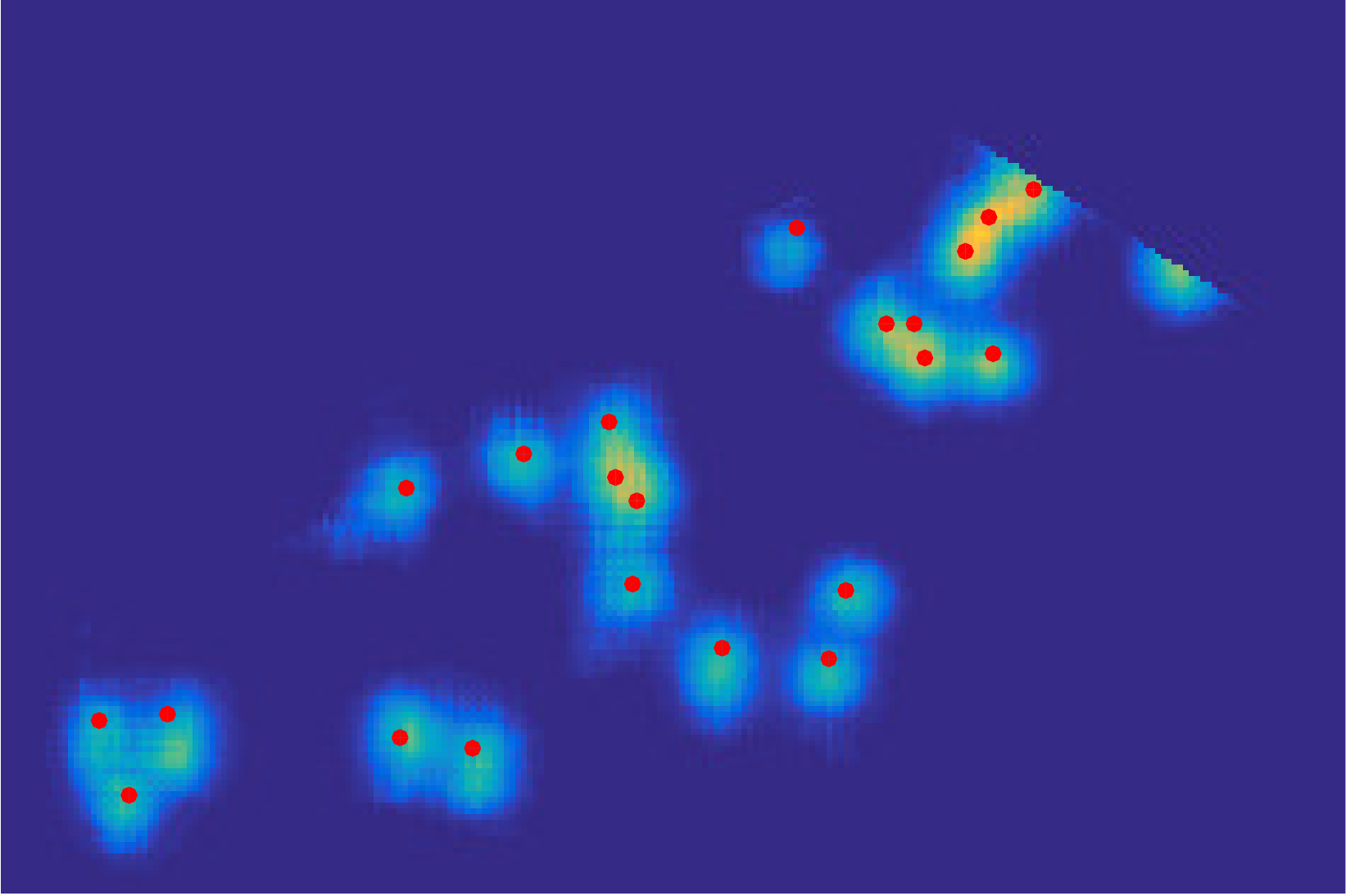} \\
  (d) Large values & (e) Small values & (f) Negative values \\
  \includegraphics[width=0.16\textwidth]{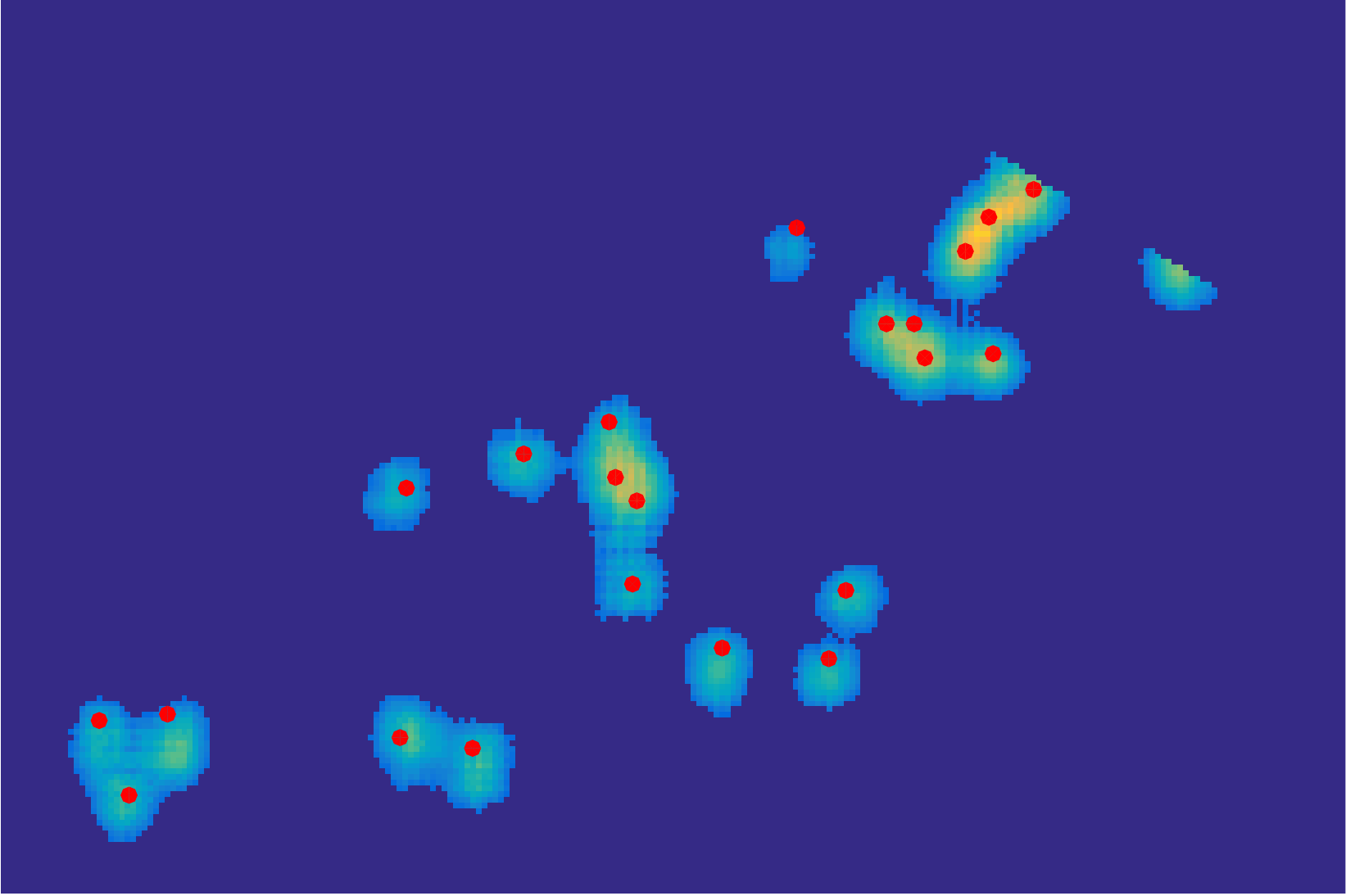} &
  \includegraphics[width=0.16\textwidth]{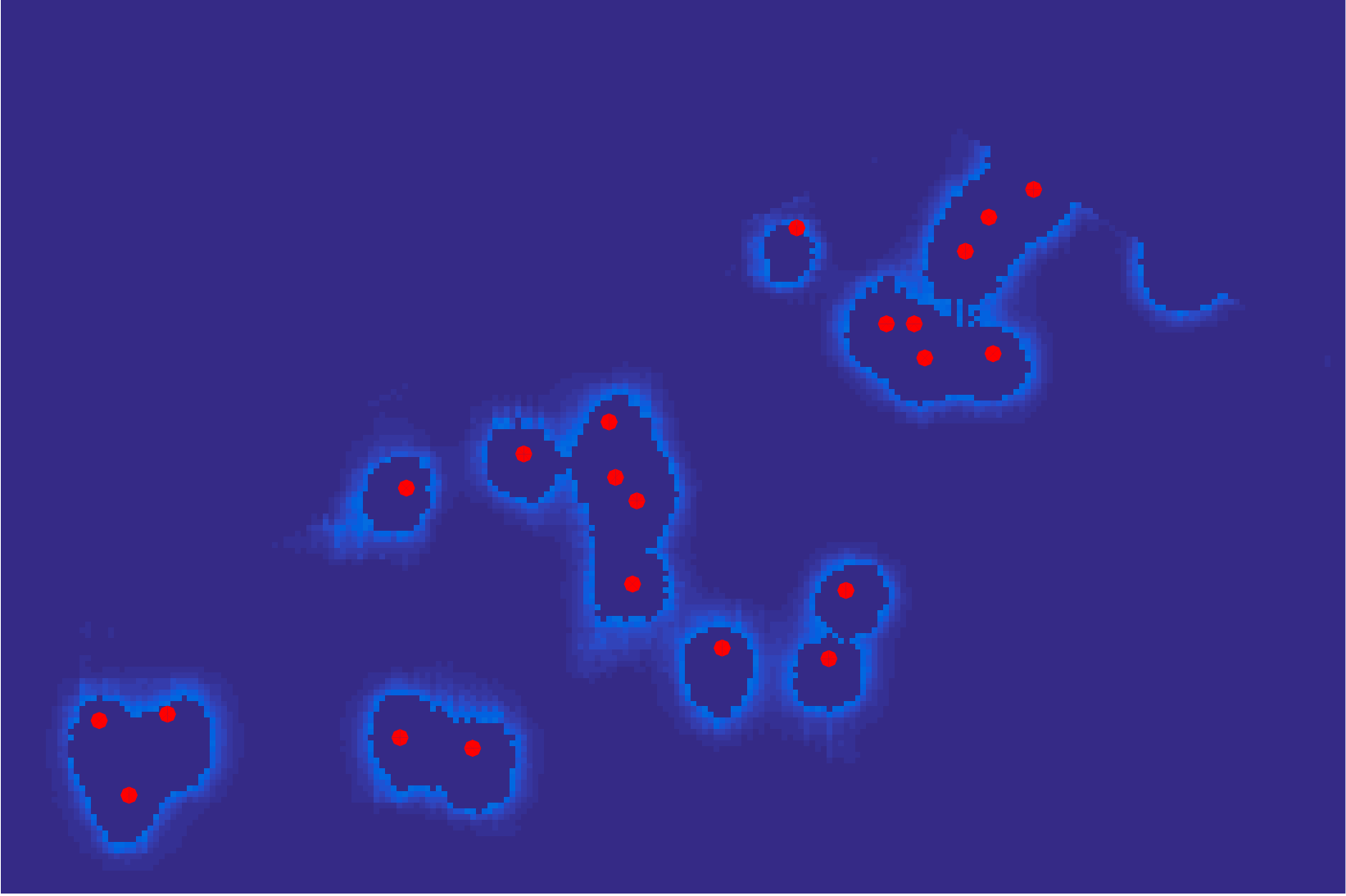} &
  \includegraphics[width=0.16\textwidth]{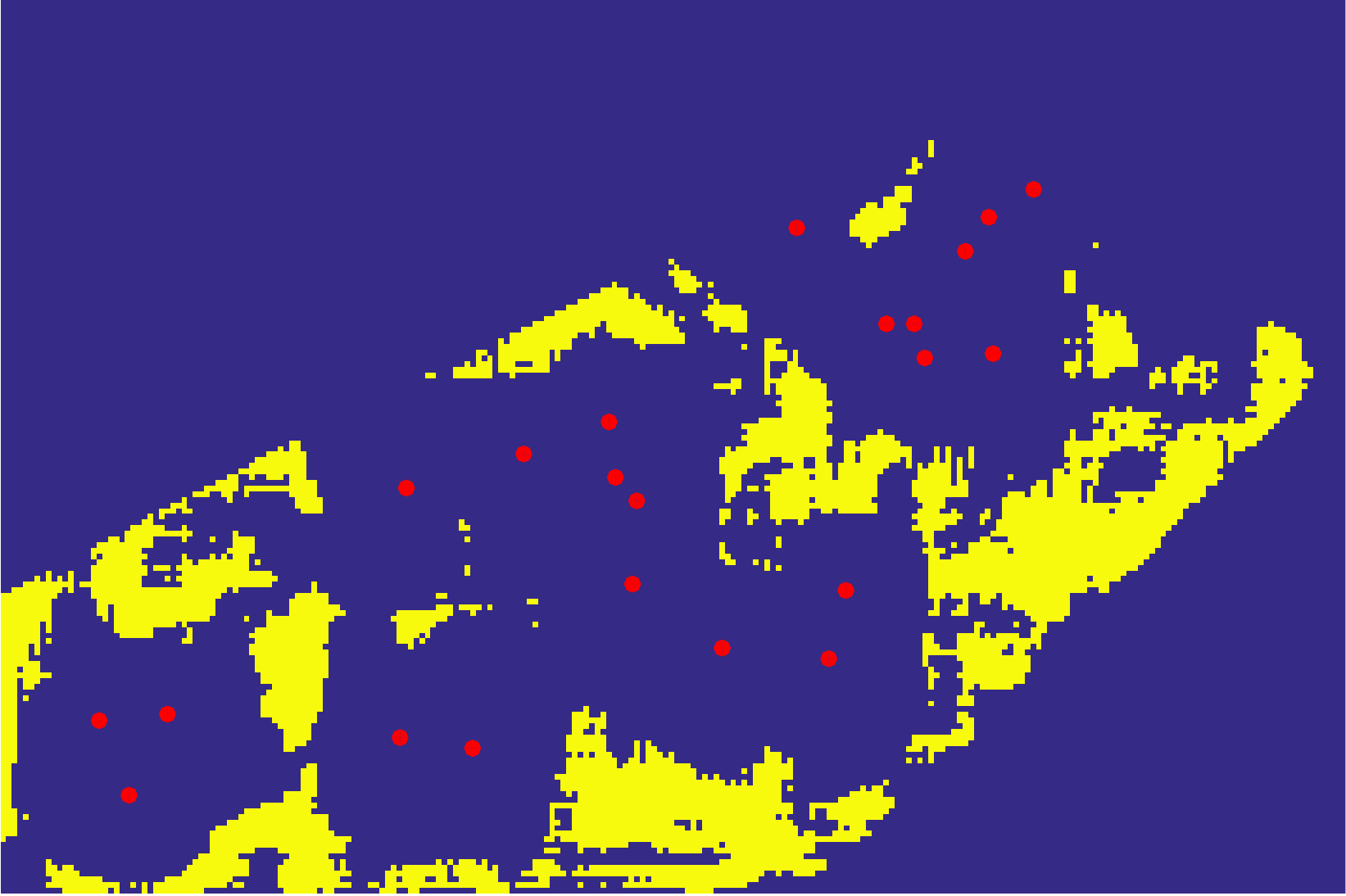}
\end{tabular}
\caption{An example density map produced by our CNN-pixel model. Our density maps are concentrated around the ground-truth annotations with little ``leakage'' of density into the background. The sum of the negative values is very small (-0.10), and does not affect the counting, detection or tracking performance.
}
\label{fig:our-dmap-demo}
\end{figure}

\par
One advantage of CNN-based method is that they can improve their performance with more training data.
When using the full training set (see Table~\ref{tab:ucsd_800} for methods trained with the full training set), both CNN-pixel and MCNN can lower their MAE compared with the ``max'' split in Table~\ref{tab:ucsd_count}, which uses only 1/5 of the training set.
An example density map estimated by our CNN-pixel model is shown in \reffig{fig:our-dmap-demo}, and \reffig{fig:dmap-comp} compares the density maps from different methods.

\subsubsection{WorldExpo'10 dataset}

\begin{figure}[tb]
\centering
\footnotesize
\begin{tabular}{@{}c@{\hspace{0mm}}c@{\hspace{0mm}}c@{}}
   Image & Ground-truth Density & Predicted Density \\
  \includegraphics[width=0.16\textwidth]{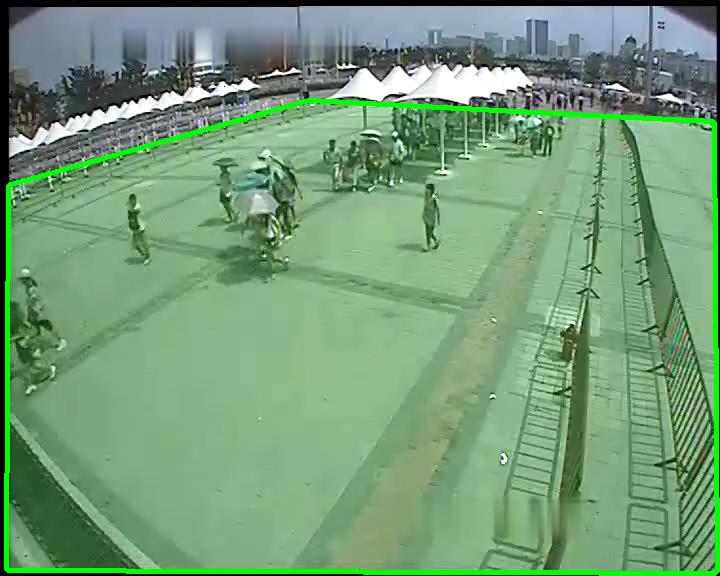} &
  \includegraphics[width=0.17\textwidth]{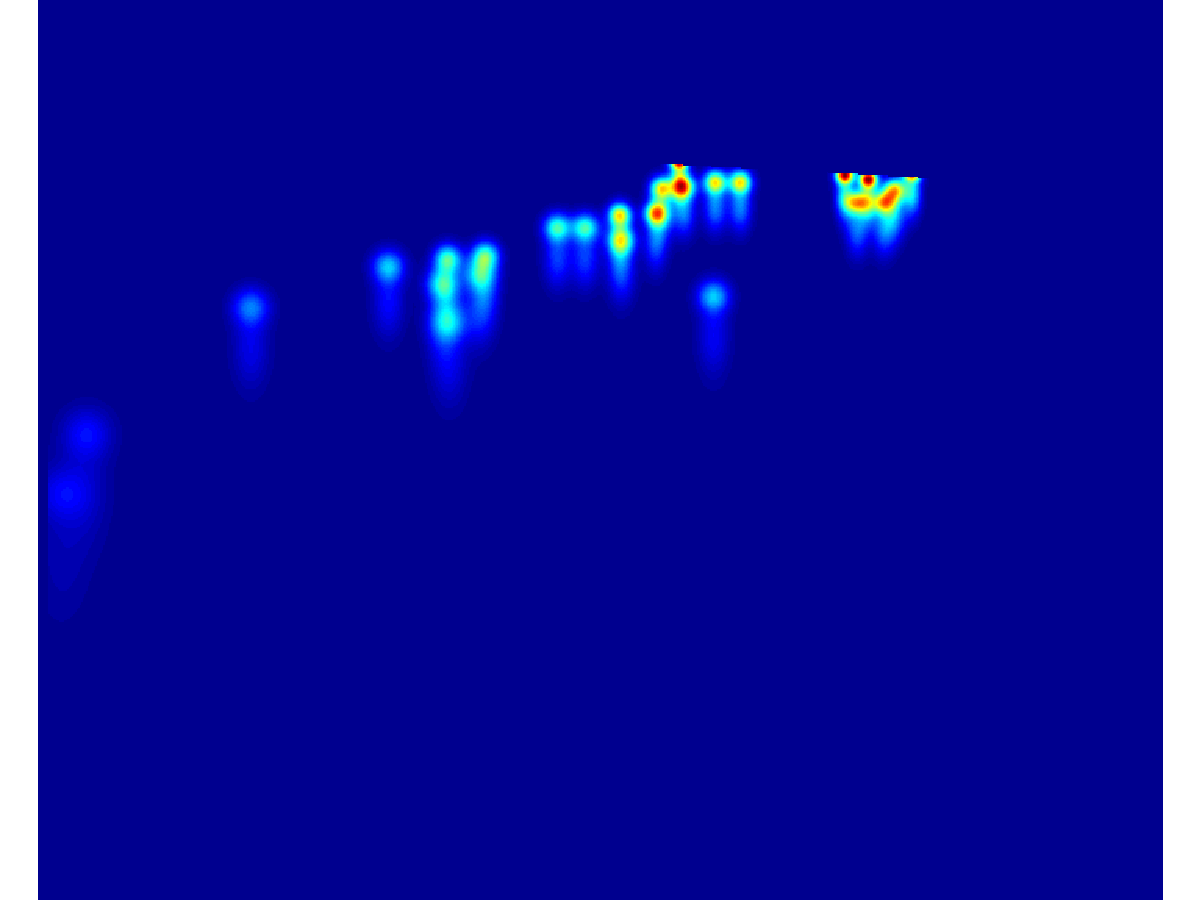} &
  \includegraphics[width=0.17\textwidth]{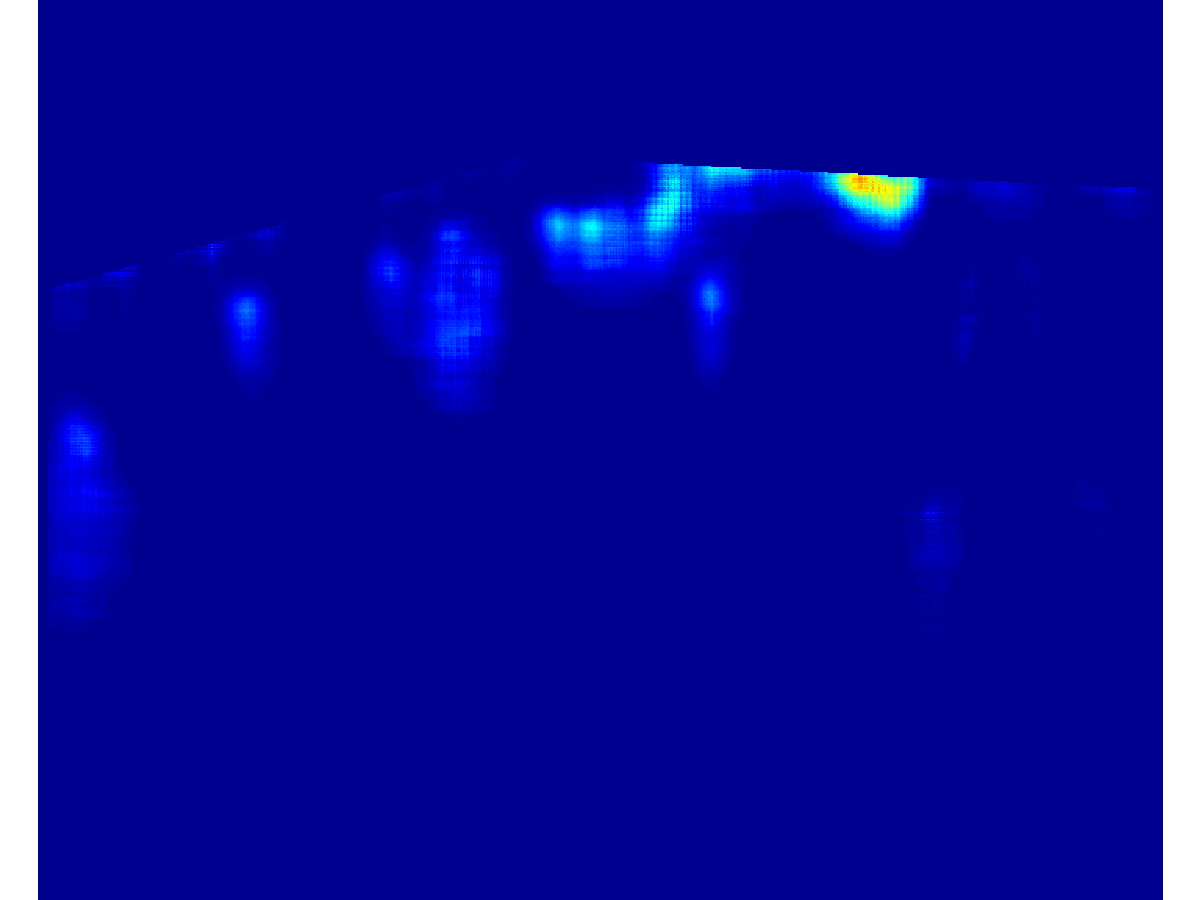} \\
  \includegraphics[width=0.16\textwidth]{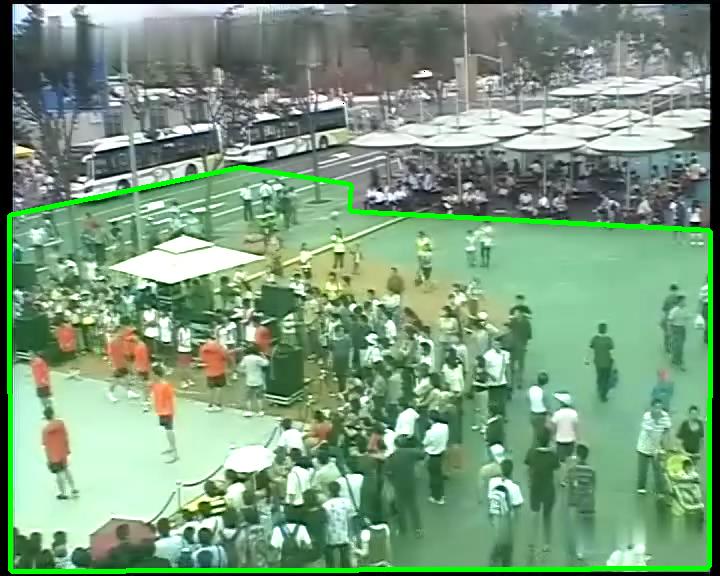} &
  \includegraphics[width=0.17\textwidth]{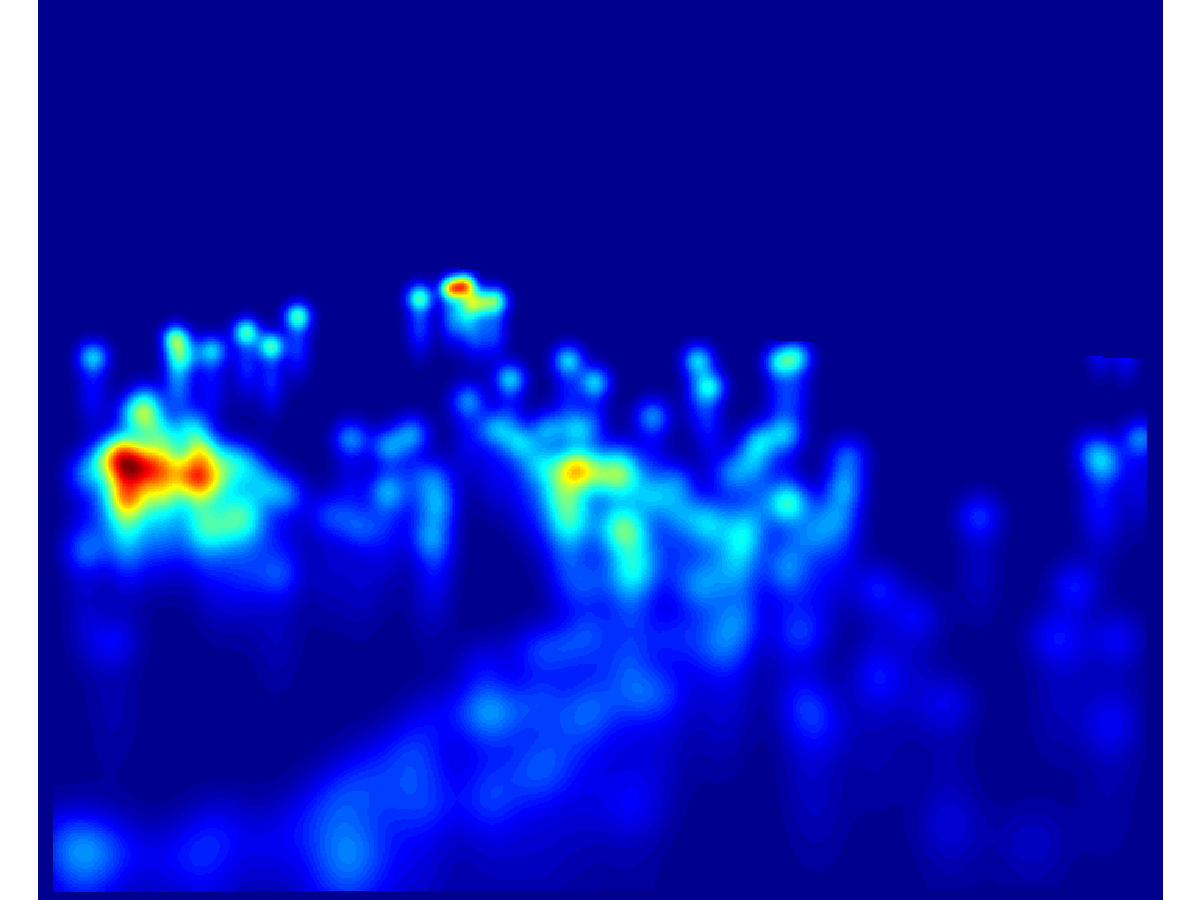} &
  \includegraphics[width=0.17\textwidth]{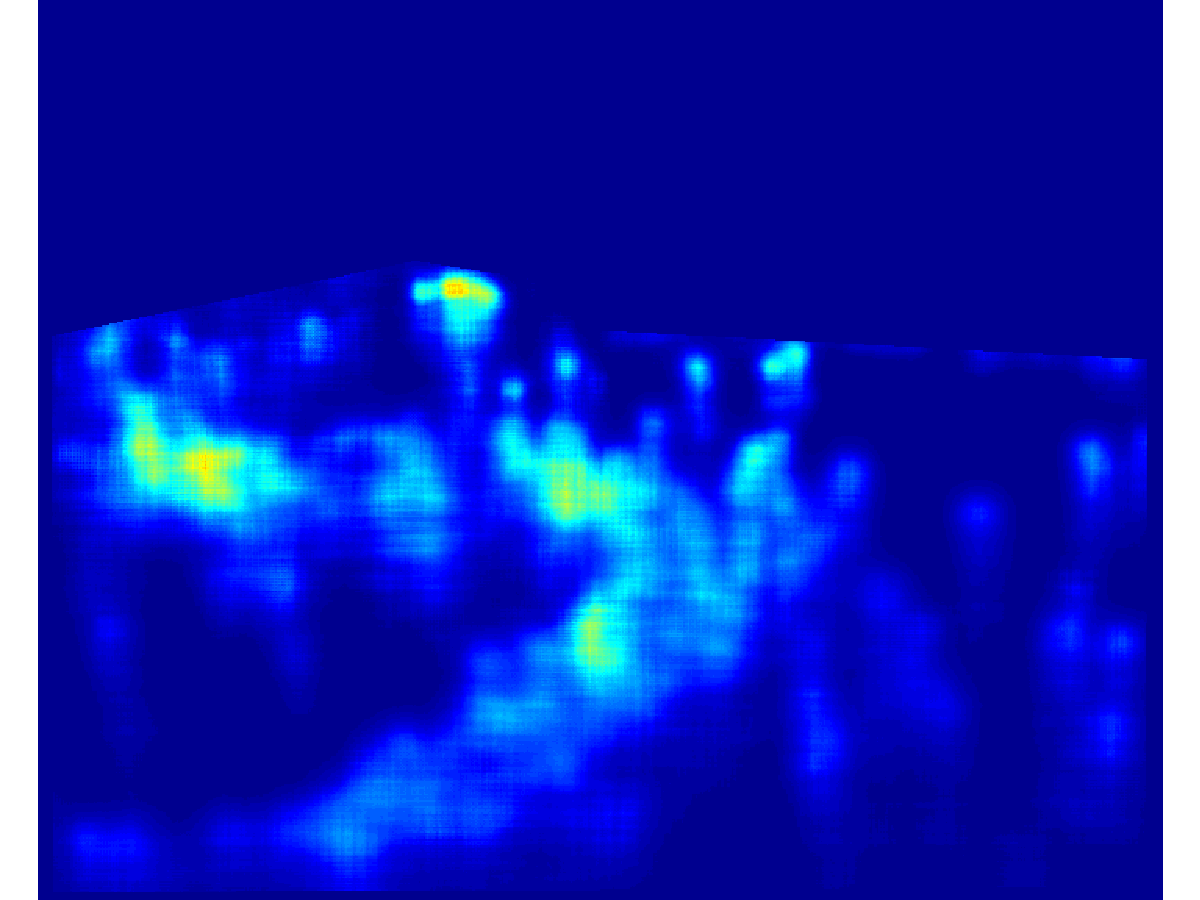} \\
  \includegraphics[width=0.16\textwidth]{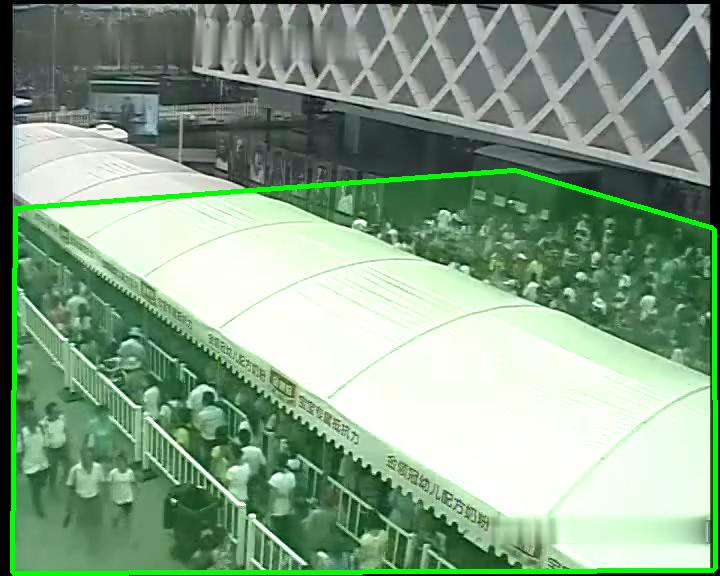} &
  \includegraphics[width=0.17\textwidth]{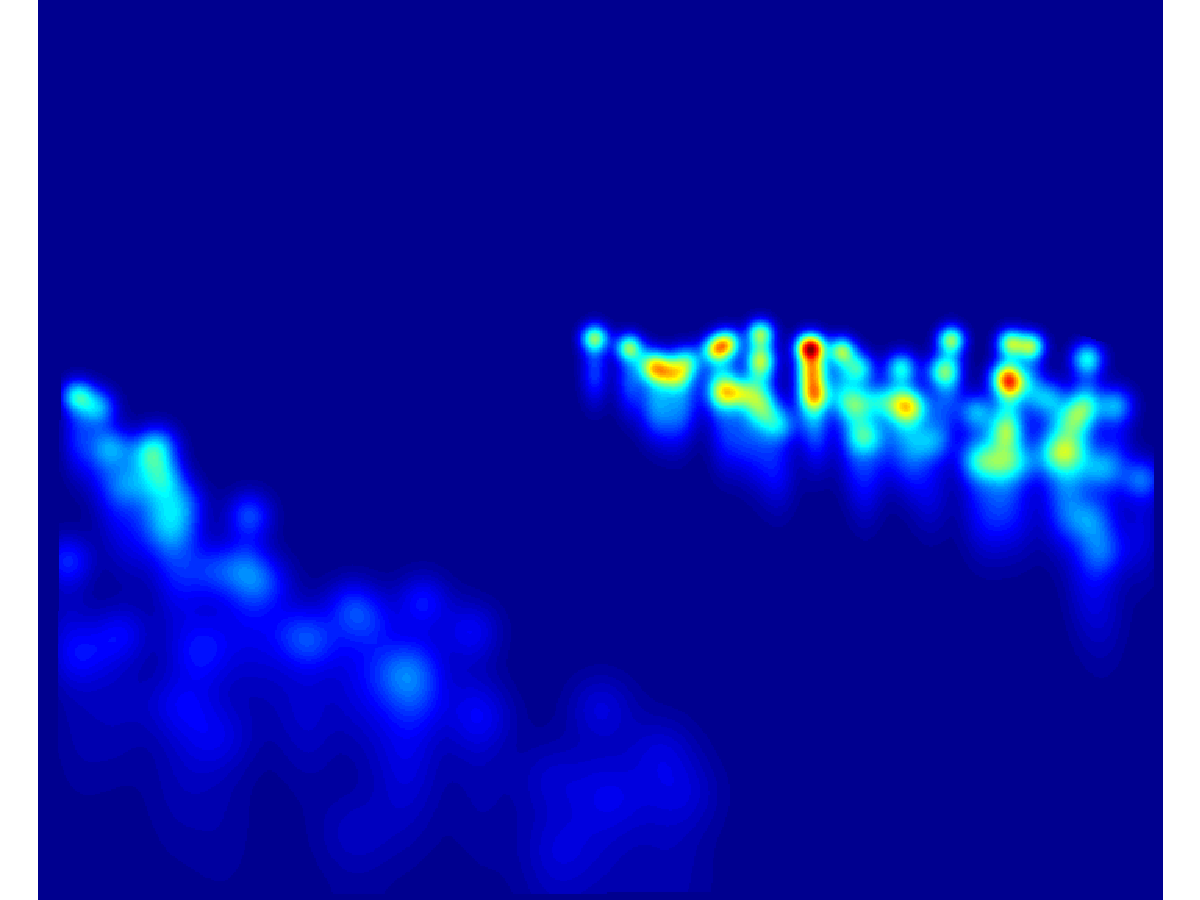} &
  \includegraphics[width=0.17\textwidth]{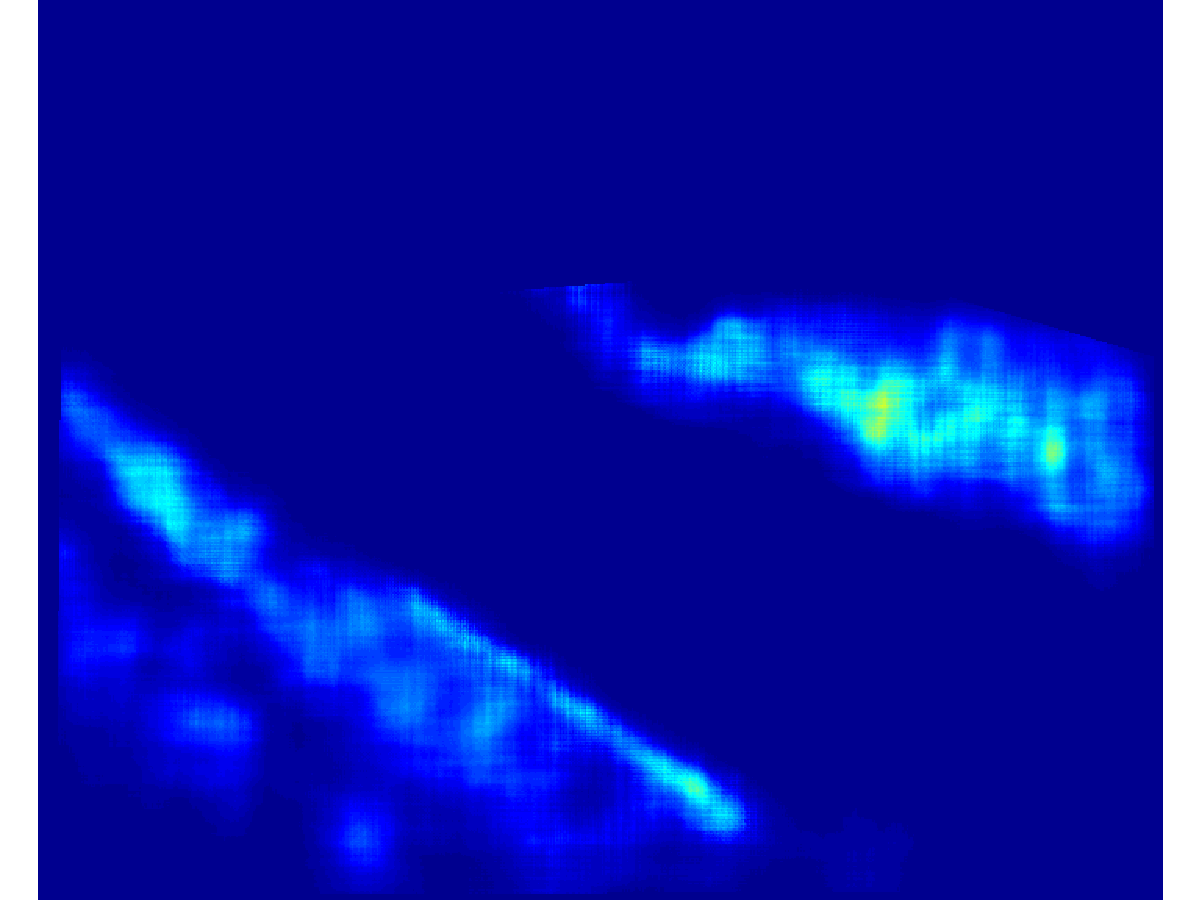} \\
  \includegraphics[width=0.16\textwidth]{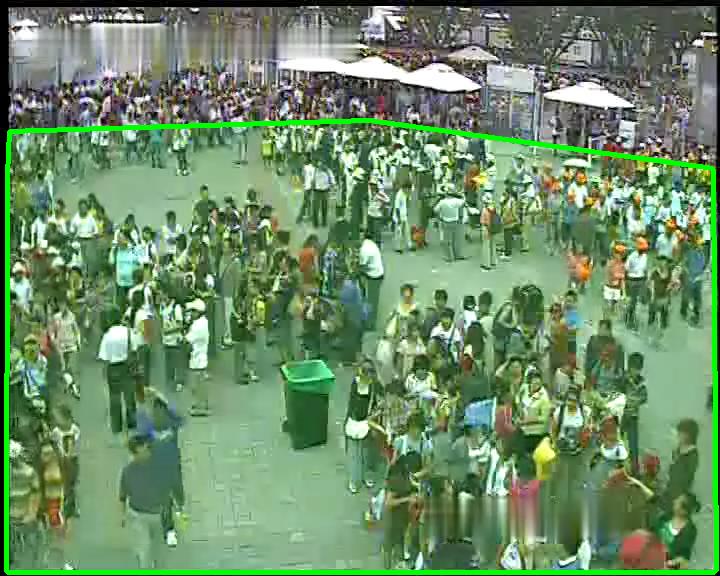} &
  \includegraphics[width=0.17\textwidth]{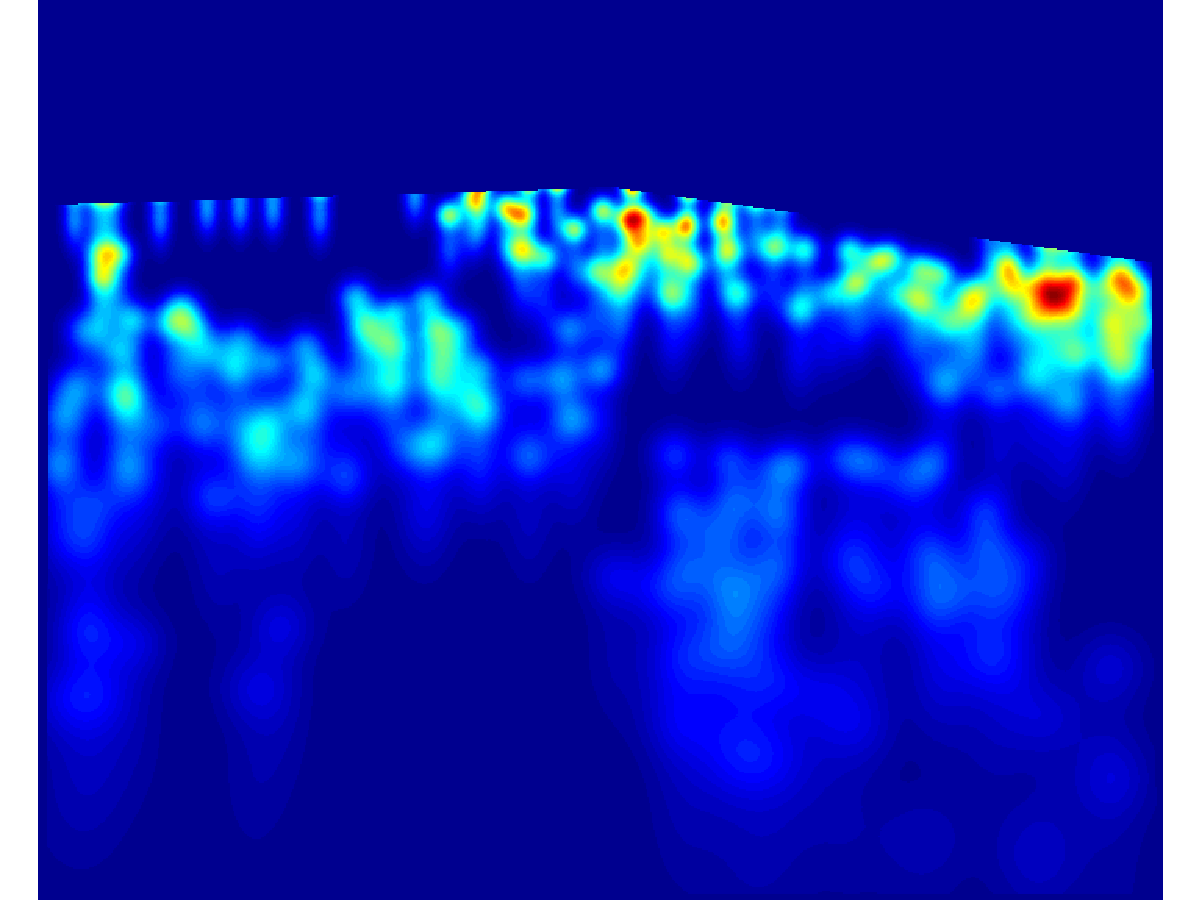} &
  \includegraphics[width=0.17\textwidth]{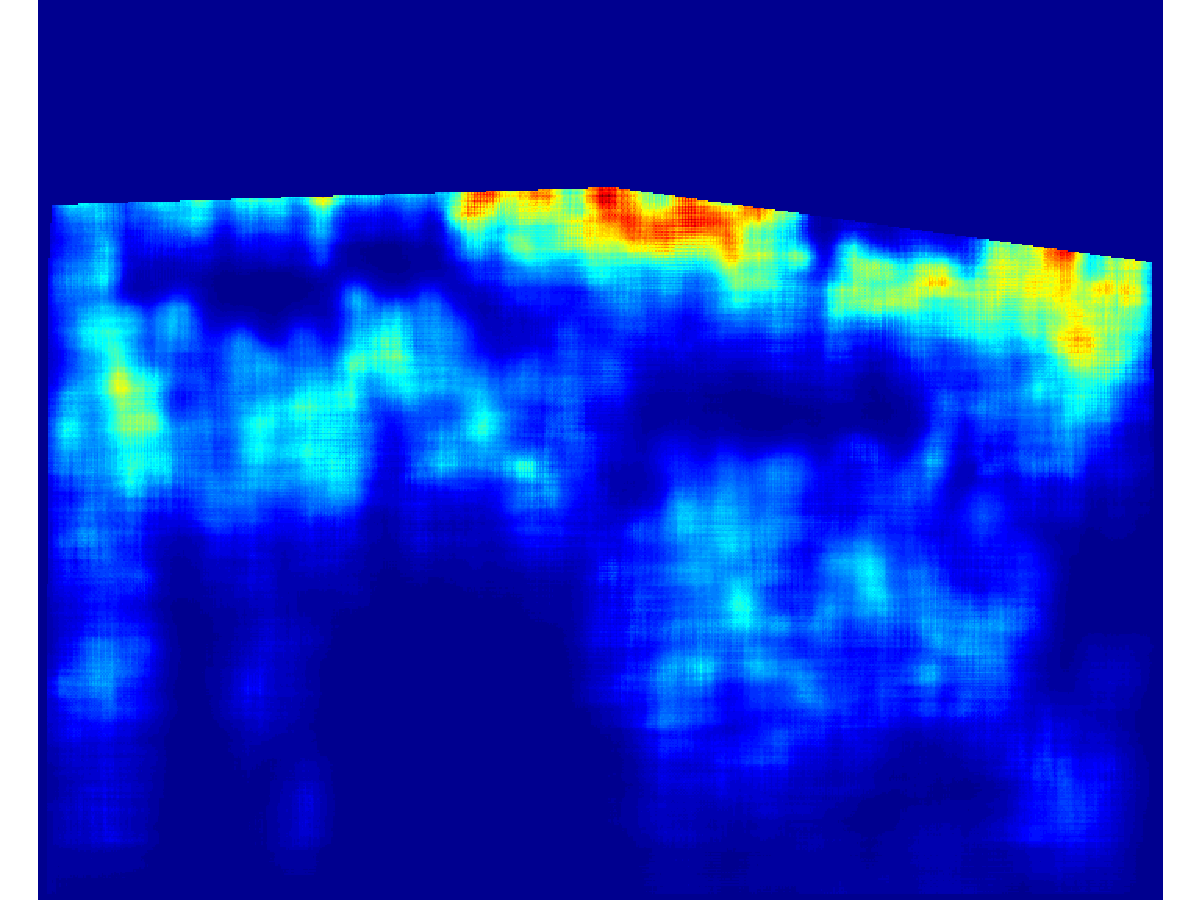} \\
  \includegraphics[width=0.16\textwidth]{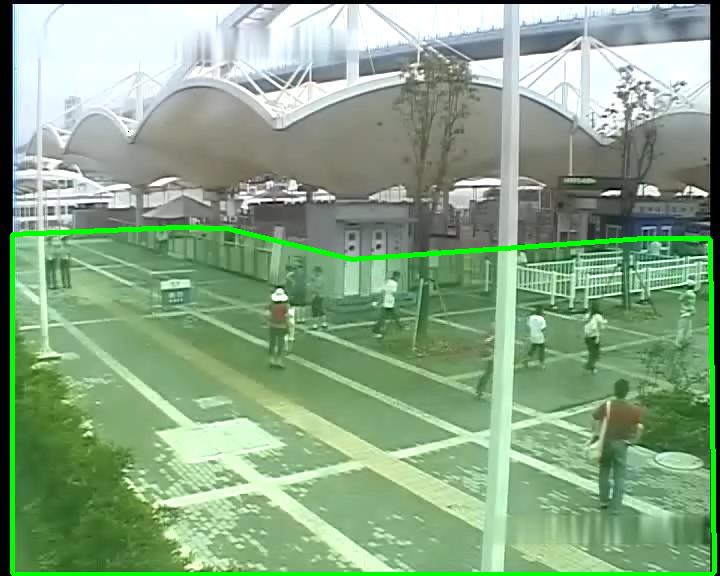} &
  \includegraphics[width=0.17\textwidth]{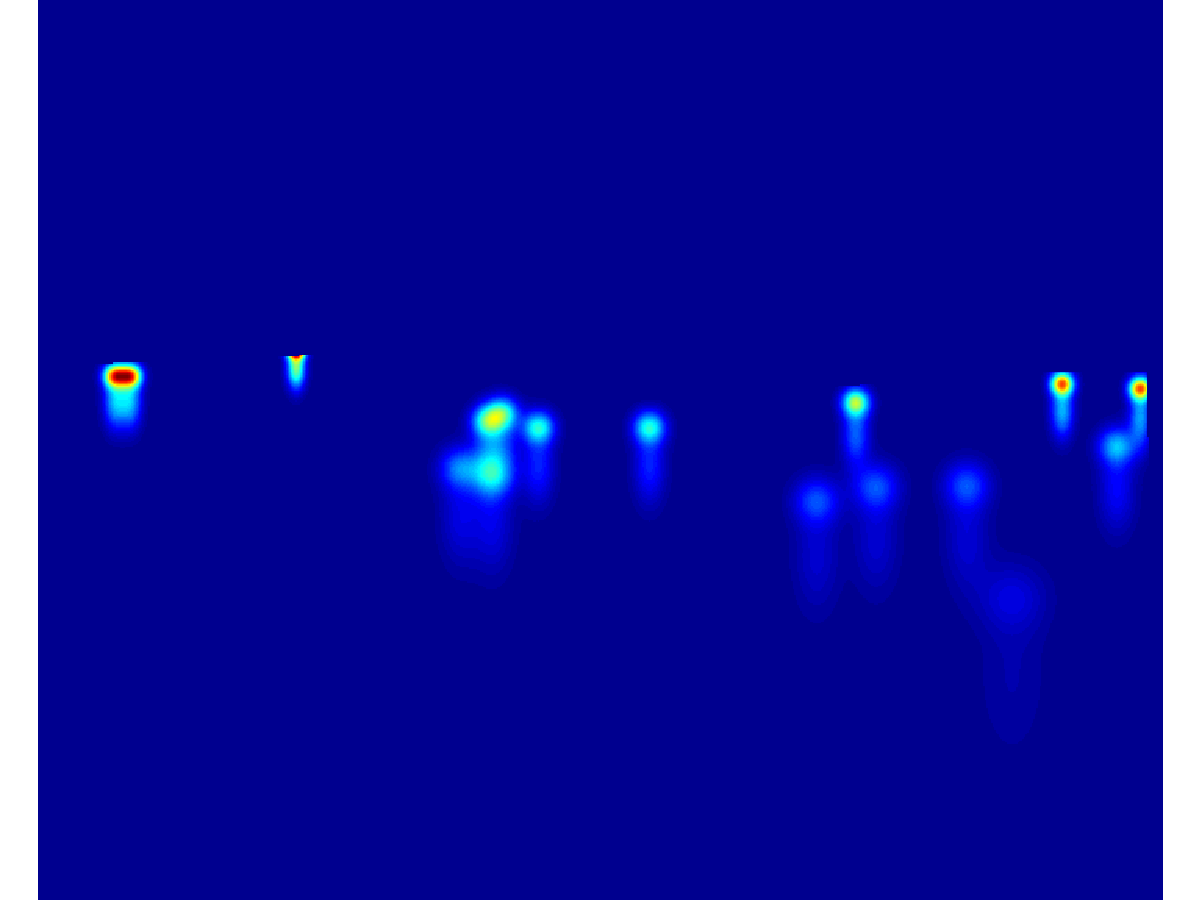} &
  \includegraphics[width=0.17\textwidth]{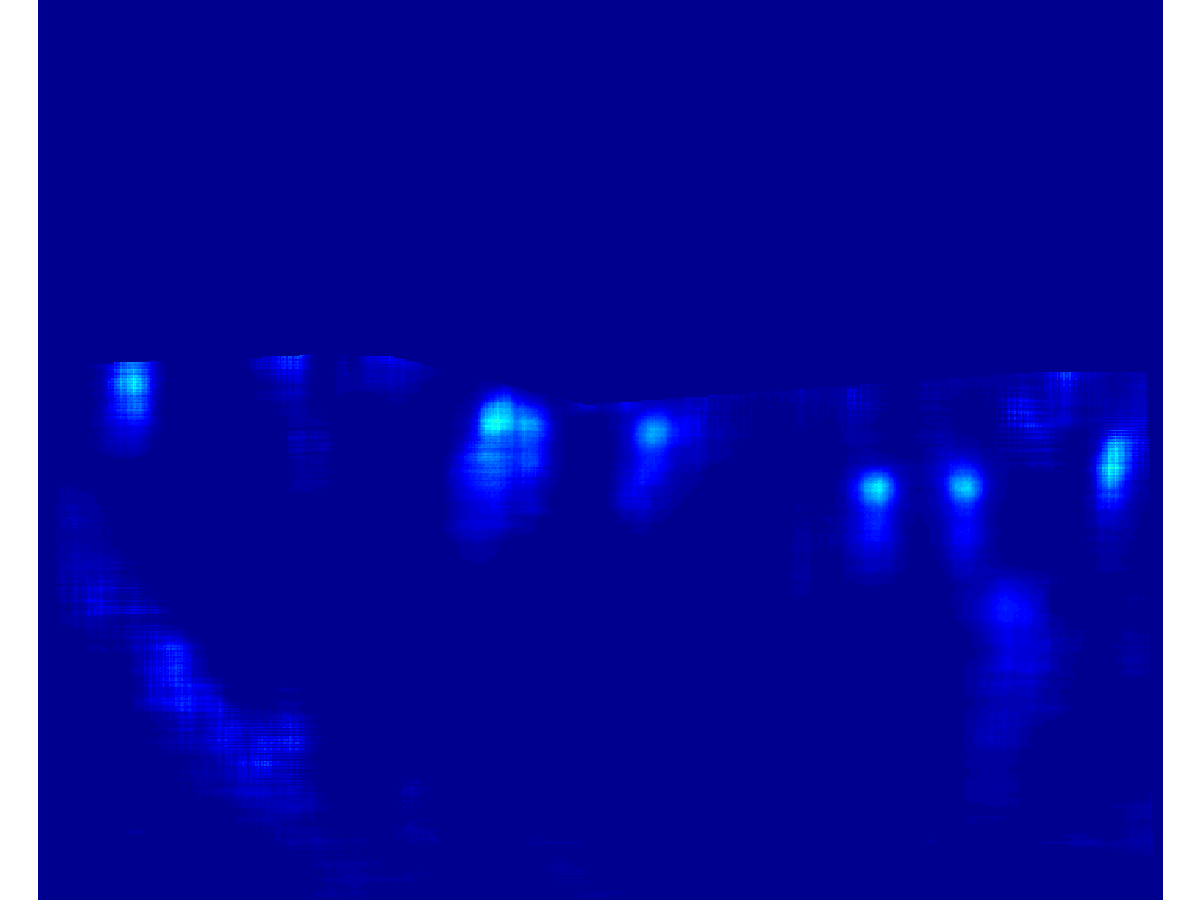}
\end{tabular}
\centering
\caption{Results on WorldExpo dataset using our CNN-pixel model.}
\label{fig:demo_WorldExpo}
\end{figure}

\begin{table*}[tbhp]
\small
\centering
\begin{tabular}{|c|c|c|c|c|c|c|}
  \hline
  Method                              & Scene 1  & Scene 2  & Scene 3  & Scene 4 & Scene 5 & Average \\
  \hline
  LBP+RR                              & 13.6     & 58.9     & 37.1     & 21.8    & 23.4    & 31.0    \\
  CNN-patch \cite{Zhang2015}          & 10.0     & 15.4     & 15.3     & 25.6    & 4.1     & 14.1    \\
  CNN-patch + cross-scene fine-tuning \cite{Zhang2015} & 9.8   & \bf{14.1}     & 14.3     & 22.2    & \bf{3.7}     & 12.9 \\
  MCNN \cite{zhang2016single}         & 3.4      & 20.6     & \bf{12.9}     & \bf{13.0}    & 8.1    & \bf{11.6} \\
  \hline
  CNN-pixel (ours)                    & {\bf 2.9}      & 18.6     & 14.1     & 24.6    & 6.9     & 13.4    \\
  FCNN-skip (ours)                    & 3.9      &  16.9     & 19.3     & 27.7    & 5.6     & 14.7    \\
  \hline
\end{tabular}
\caption{Mean absolute error (MAE) on WorldExpo'10 dataset.}
\label{tab:WorldExpo_count}
\end{table*}

The results on the WorldExpo'10 dataset \cite{Zhang2015}
are shown in Table \ref{tab:WorldExpo_count} and \reffig{fig:demo_WorldExpo}.
MCNN has the lowest average error, while CNN-pixel, FCNN-skip and CNN-patch without fine-tuning perform similarly.
Looking at the individual scenes, different methods perform better on each scene.
All the methods have larger errors on Scenes 2, 3 and 4, compared with Scenes 1 and 5. These three scenes contain more people on average, while only 12\% of all the training frames contain large crowds ($>$80 people).
On Scene 3, CNN-pixel under-predicts the density because there are many people in the upper-right that are partially-occluded by the roof, leaving just the heads visible.
Because the roof is in the ROI, the ground-truth density has non-zero values on the roof (see Fig.~\ref{fig:demo_WorldExpo}).
However, CNN-pixel predicts zero density on the roof region (because it sees no human parts there), which causes the count to be under-predicted.
For Scene 4, the ROI boundary cuts through a large crowd in the background. A large portion of the density is outside the ROI, due to the human-shaped ground-truth, where half of the density is contributed from the small head region. As a result CNN-pixel predictions tend to be larger due to this confounding effect (refer to the count plot and the error heat maps in the supplementary.).

\subsubsection{UCF\_CC\_50 dataset}
\begin{figure}[tb]
\centering
\footnotesize
\begin{tabular}{@{}c@{\hspace{1mm}}c@{\hspace{1mm}}c@{}}
  (a) Image & (b) Ground truth (1566) & (c) Prediction (1330.9)\\
  \includegraphics[width=0.16\textwidth]{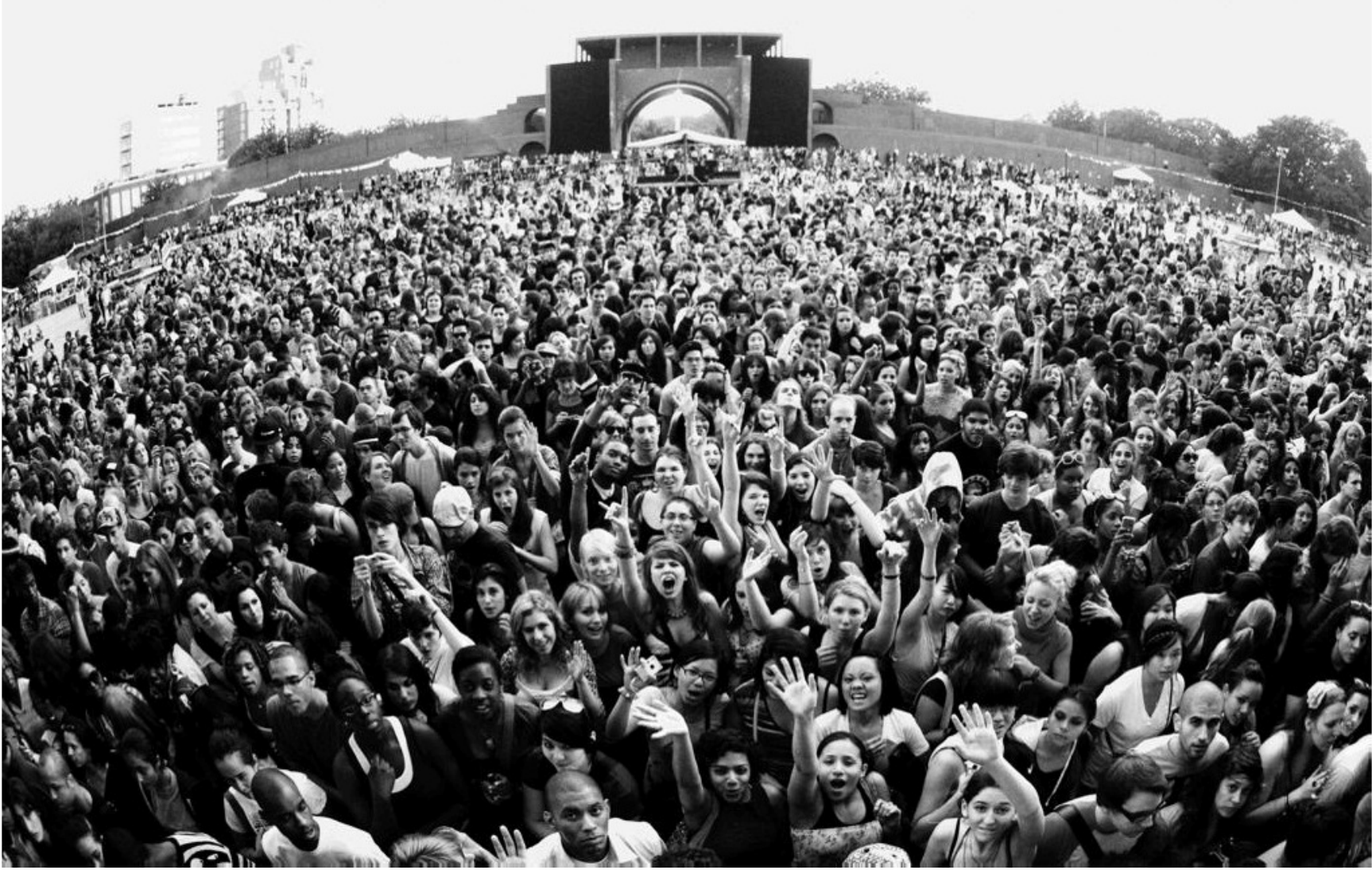} &
  \includegraphics[width=0.16\textwidth]{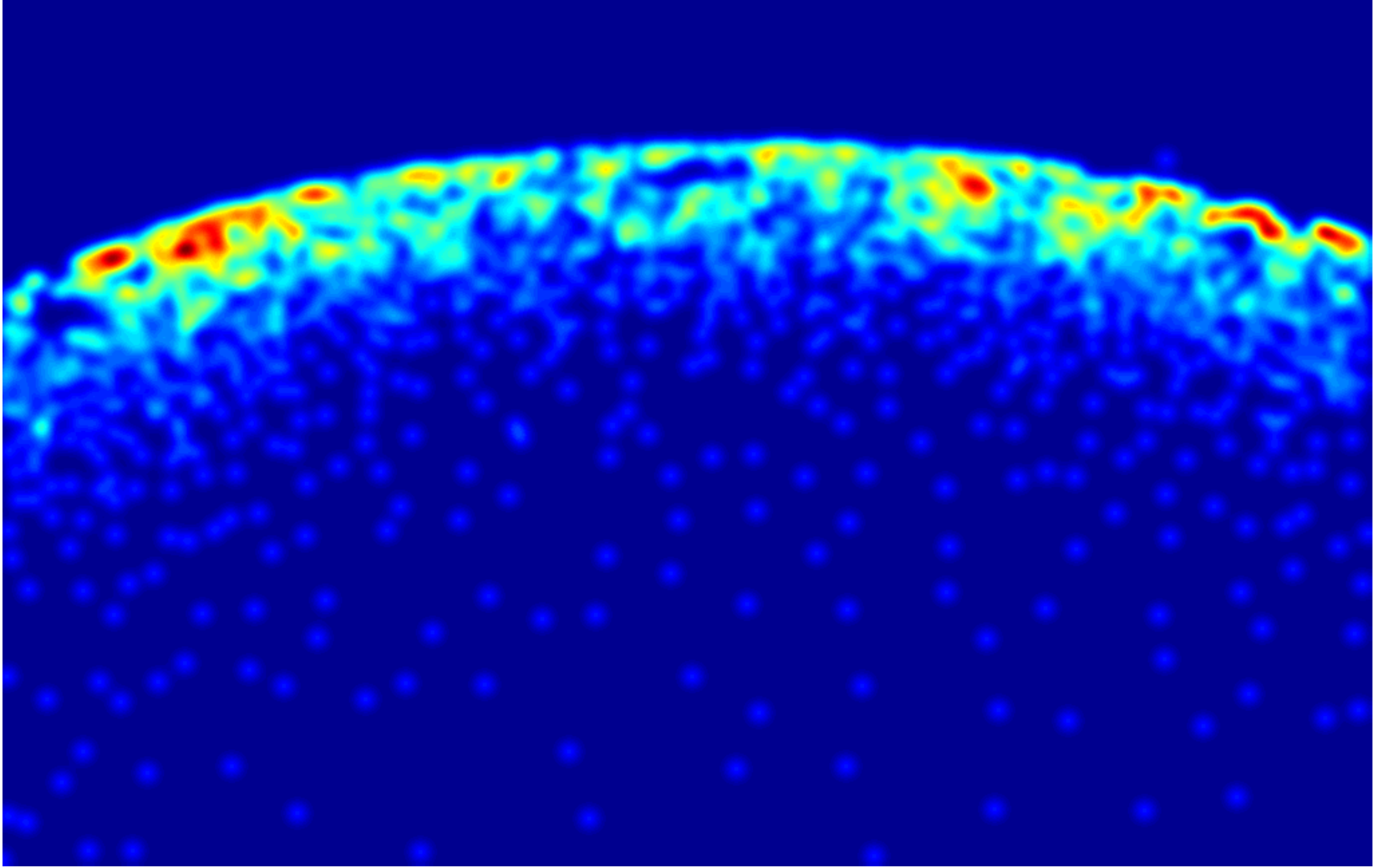} &
  \includegraphics[width=0.16\textwidth]{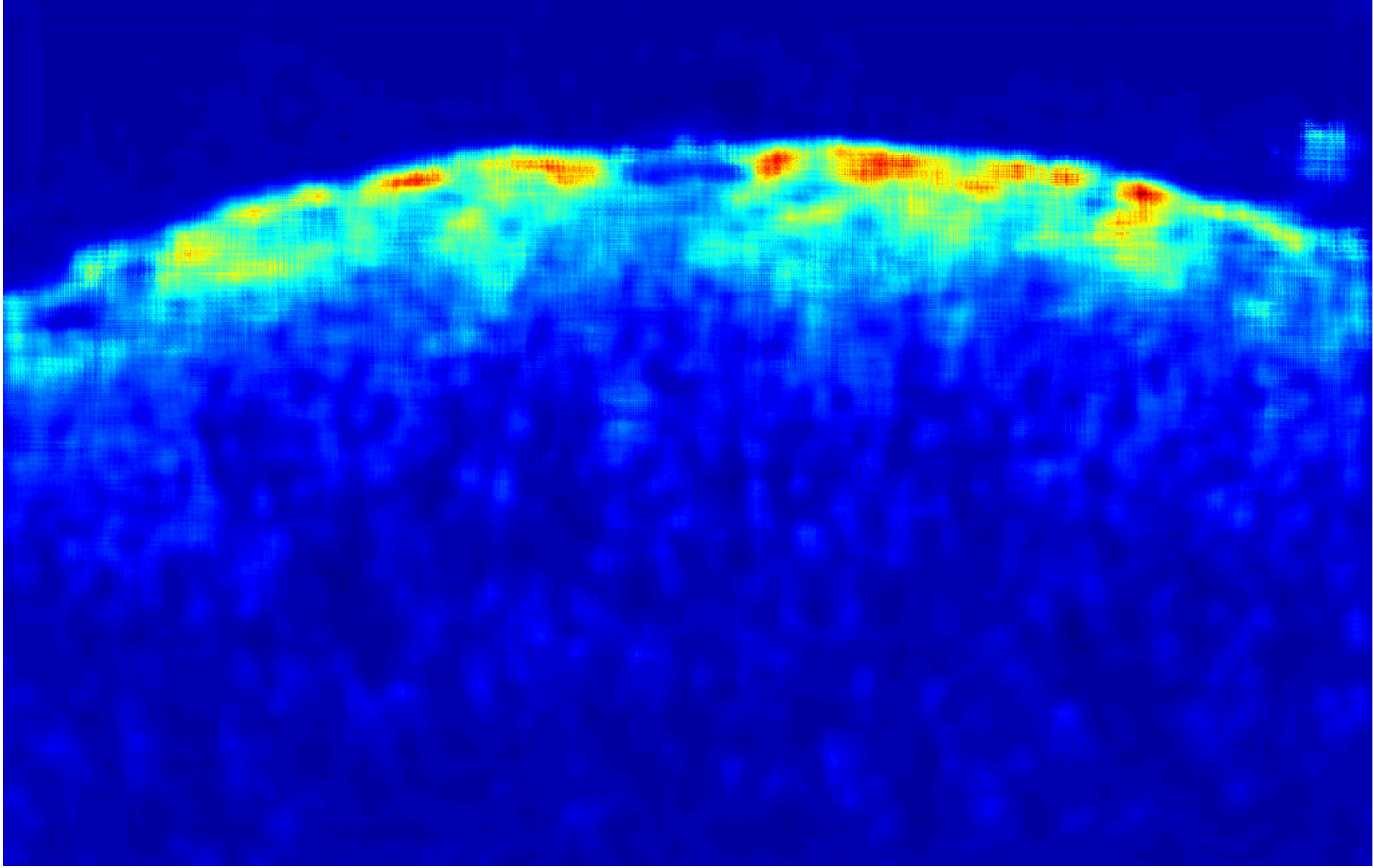} \\
  (d) Image & (e) Ground truth (440) & (f) Prediction (358.1) \\
  \includegraphics[width=0.16\textwidth]{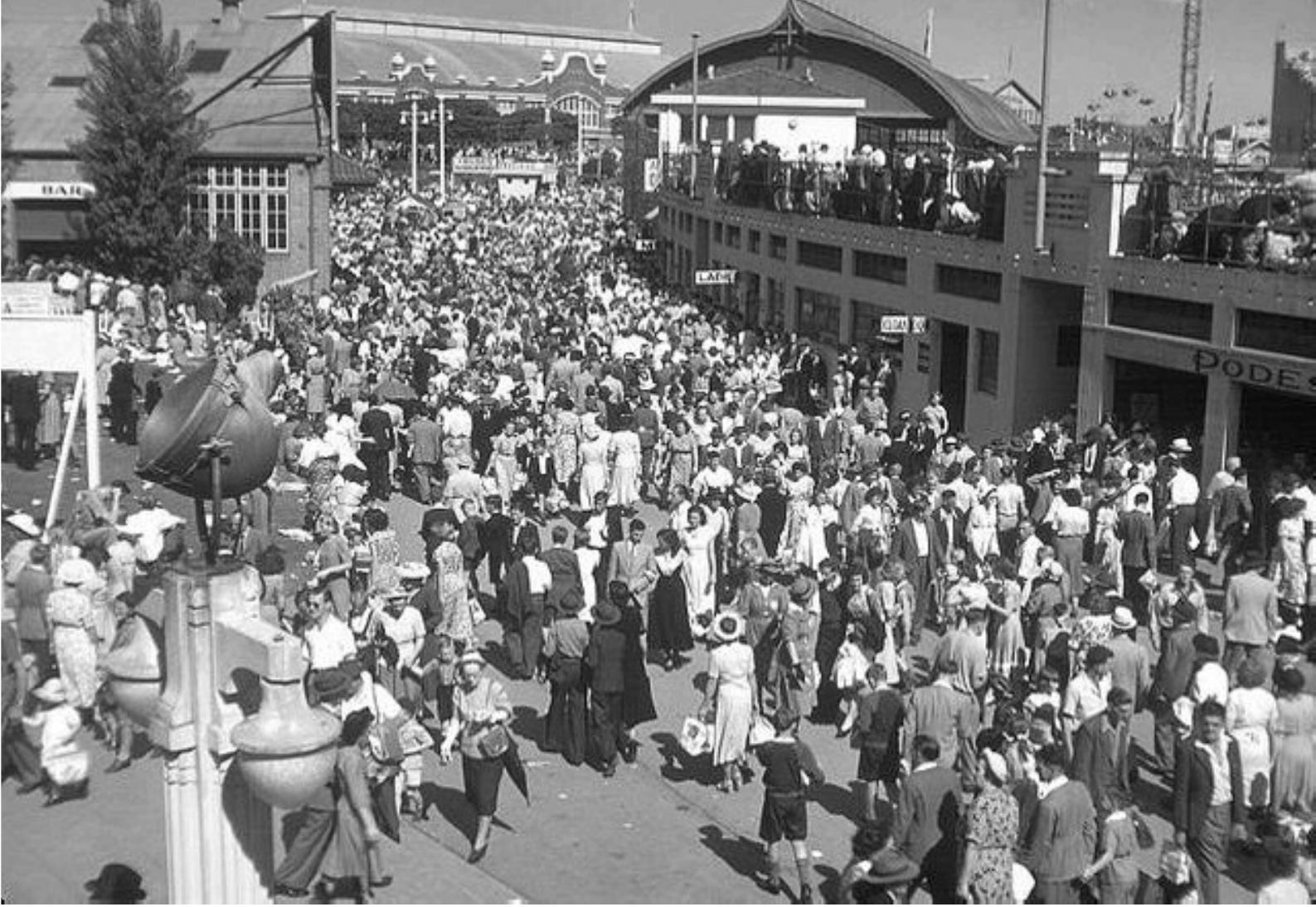} &
  \includegraphics[width=0.16\textwidth]{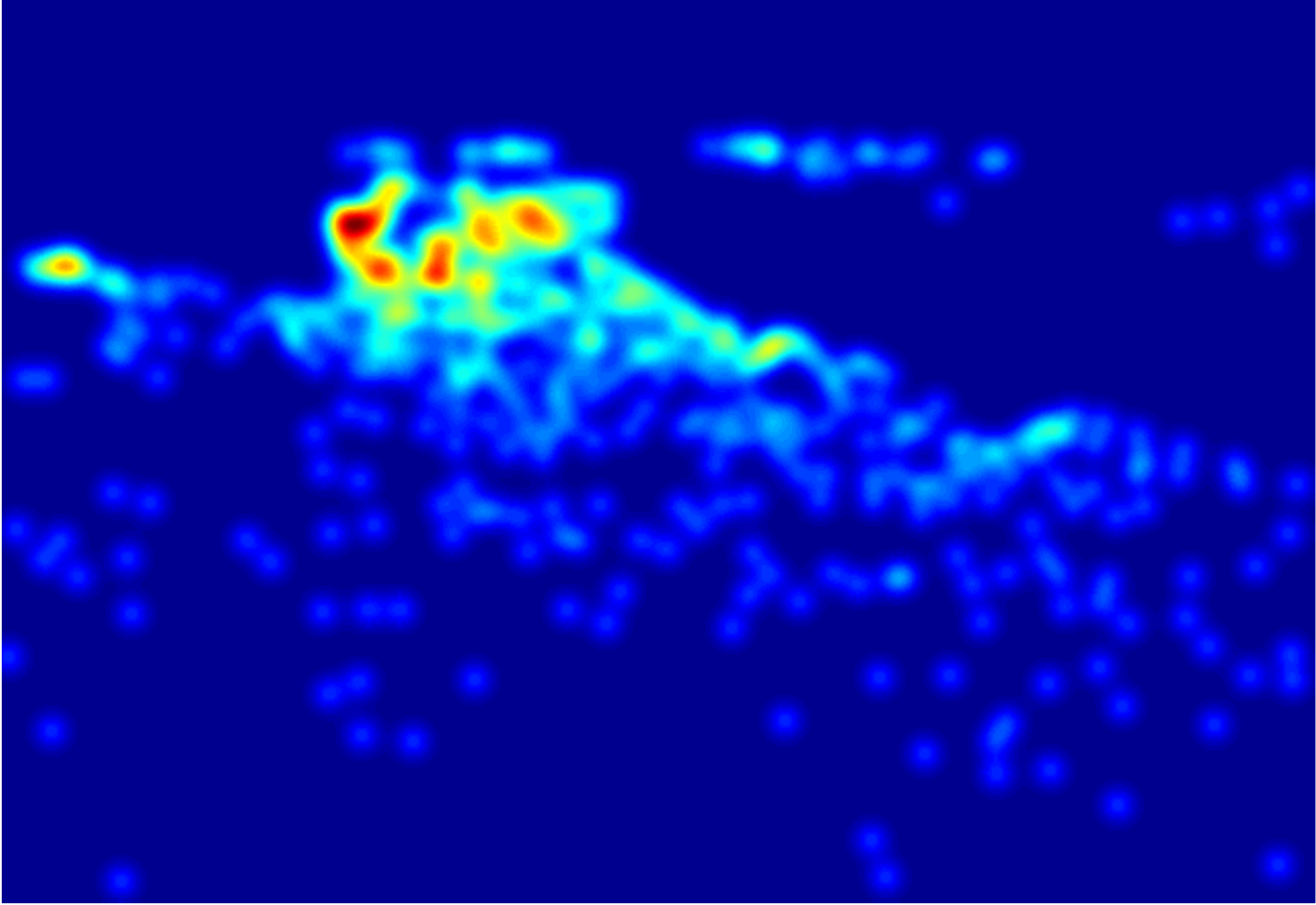} &
  \includegraphics[width=0.16\textwidth]{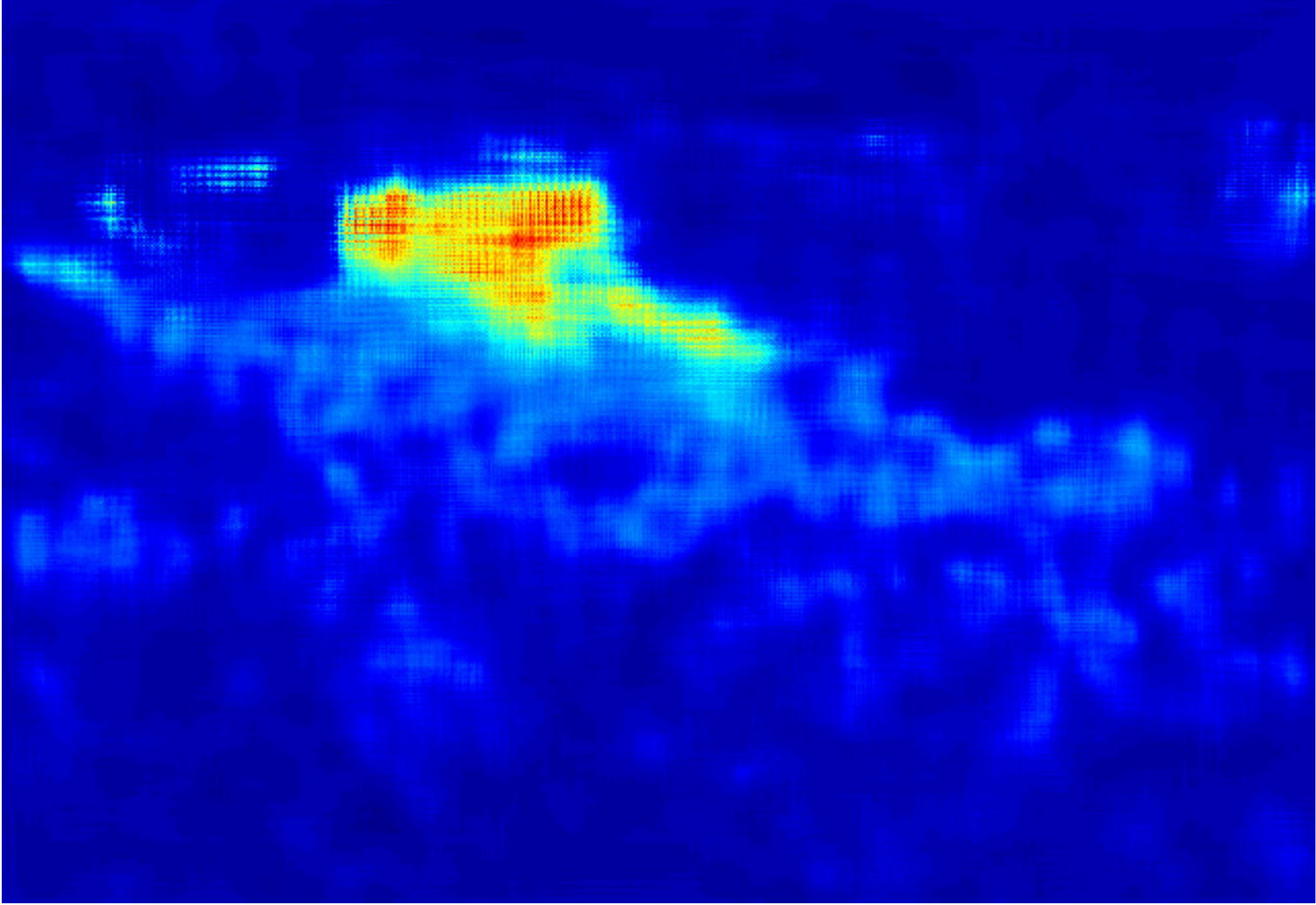} \\
\end{tabular}
\caption{Example results on UCF\_CC\_50 dataset using our CNN-pixel model. The number in parenthesis is the count.}
\label{fig:demo_ucf}
\end{figure}

The results on the UCF dataset are presented in Table~\ref{tab:ucf_count}, and examples of our predicted density maps appear in \reffig{fig:demo_ucf}.
The CNN-based methods show their capability of handling these extremely crowded images.
Note that although there are only 40 training images, there are still many patches of people that can be extracted to train the CNN from scratch.

\begin{table}[tbp]
\small
\centering
\begin{tabular}{|c|c|c|}
  \hline
  Method                              & MAE  & Std  \\
  \hline
  MESA \cite{NIPS2010_4043} & 493.4   & 487.1   \\
  Density-aware \cite{Rodriguez2011}  & 655.7    & 697.8   \\
  FHSc \cite{Idrees2013}              & 468.0    & 590.3   \\
  FHSc + MRF \cite{Idrees2013}        & 419.5    & 541.6   \\
  \hline
  CNN-patch \cite{Zhang2015}          & 467.0    & 498.5   \\
  MCNN \cite{zhang2016single}         & 377.6    & 509.1   \\
  Hydra 2s \cite{onoro2016towards}    & \bf{333.7} & 425.3 \\
  CNN-boost fine-tuned using 1 boost \cite{onoro2016towards} & 364.4  & \bf{341.4} \\
  CNN-pixel (ours)                    & 406.2 & 404.0 \\
  FCNN-skip (ours)                    & 431.6 & 379.6 \\
  \hline
\end{tabular}
\caption{Mean absolute error (MAE) on the UCF dataset. Std is the standard deviation of the MAE.}
\label{tab:ucf_count}
\end{table}

\subsubsection{TRANCOS dataset}
\begin{figure}[tbp]
\centering
\footnotesize
\begin{tabular}{@{}c@{\hspace{1mm}}c@{\hspace{1mm}}c@{}}
  (a) Image & (b) Ground truth (58) & (c) Prediction (53.85)\\
  \includegraphics[width=0.16\textwidth]{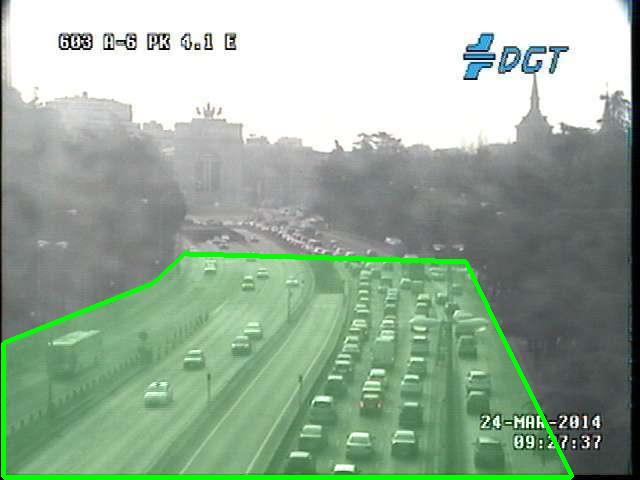} &
  \includegraphics[width=0.16\textwidth]{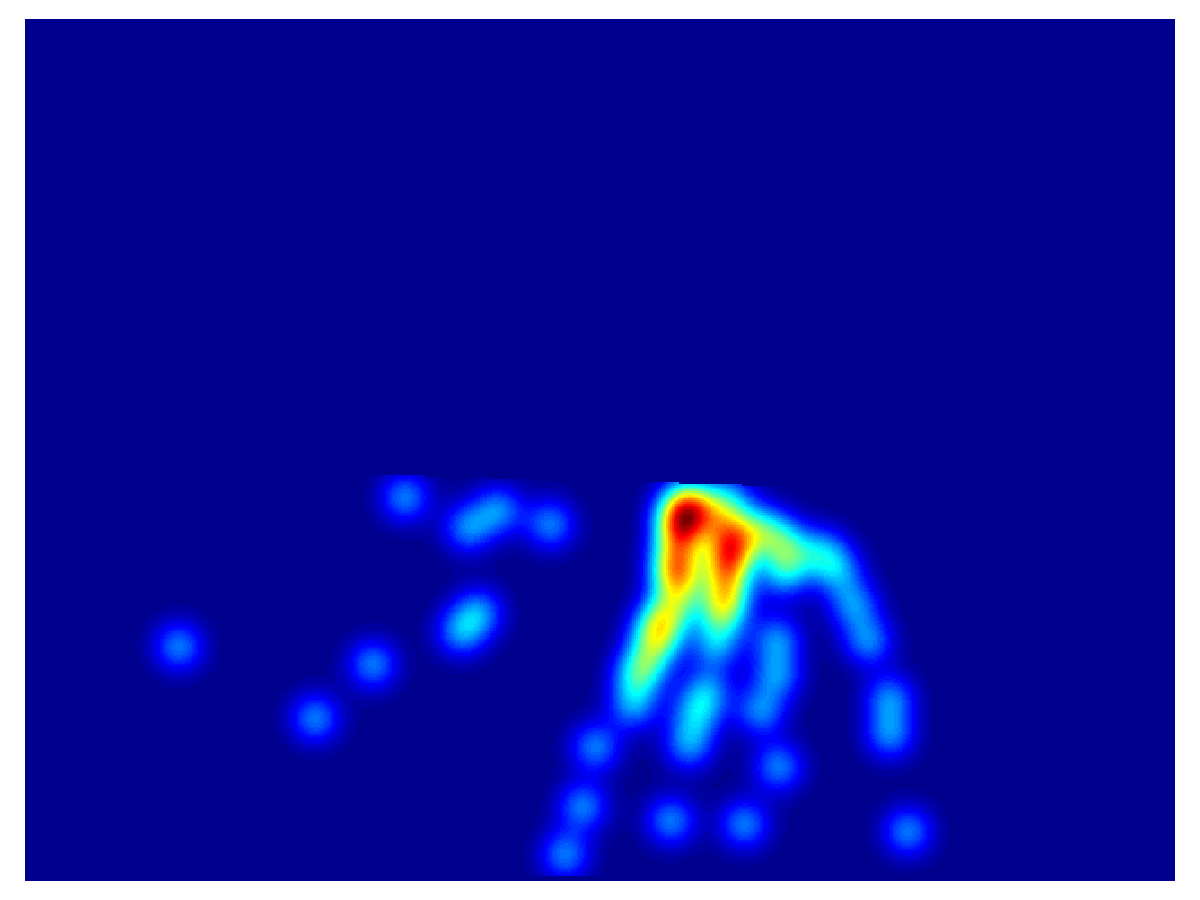} &
  \includegraphics[width=0.16\textwidth]{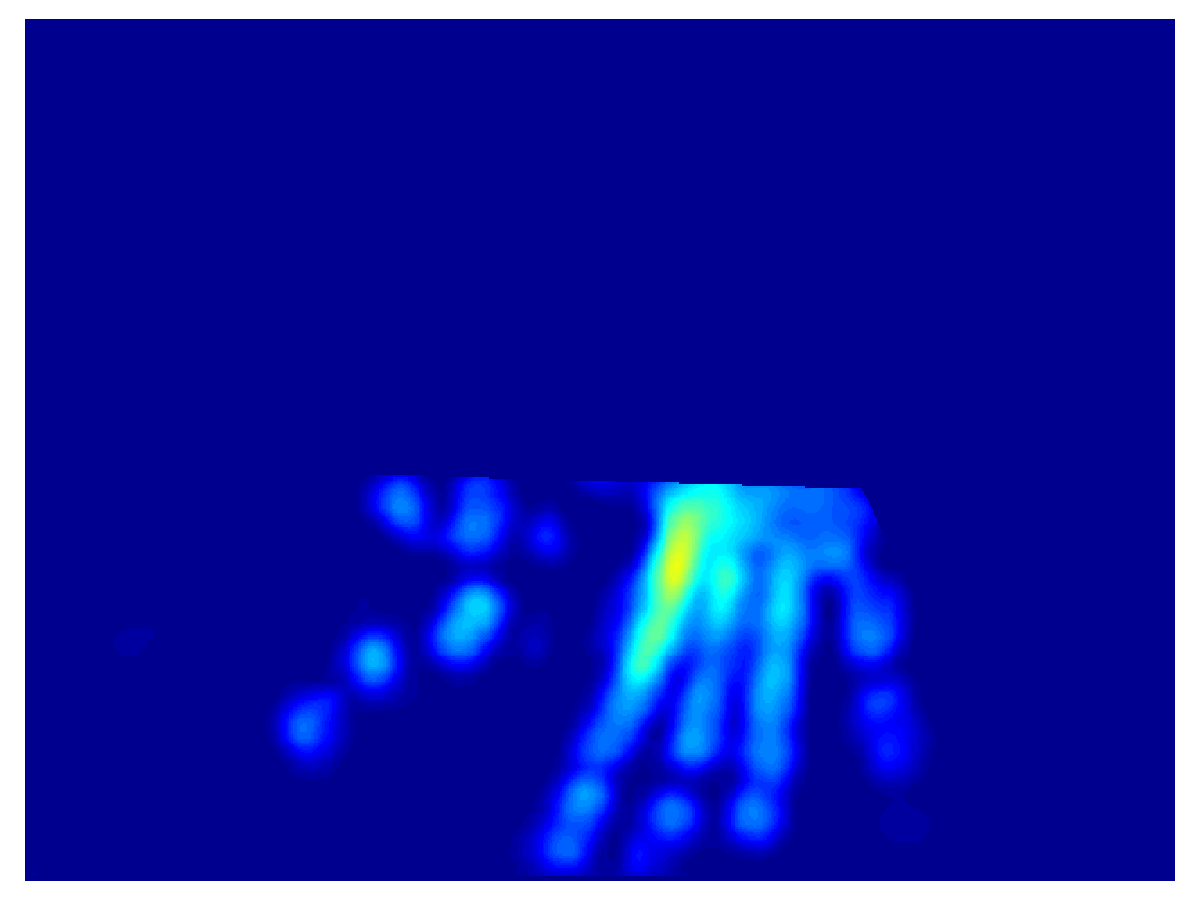} \\
  (d) Image & (e) Ground truth (55) & (f) Prediction (56.83) \\
  \includegraphics[width=0.16\textwidth]{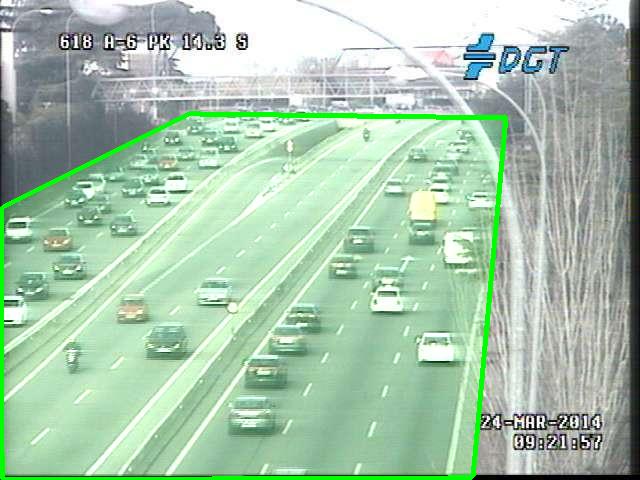} &
  \includegraphics[width=0.16\textwidth]{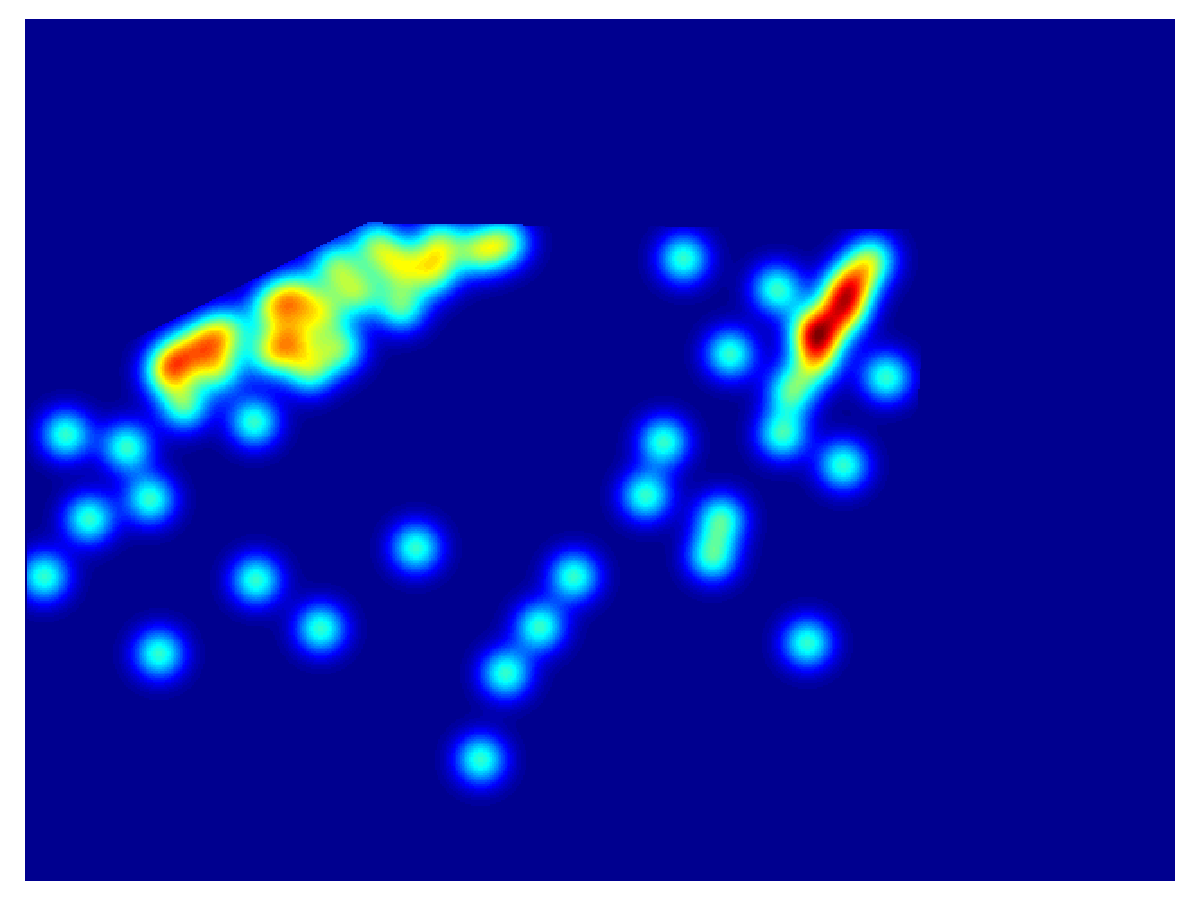} &
  \includegraphics[width=0.16\textwidth]{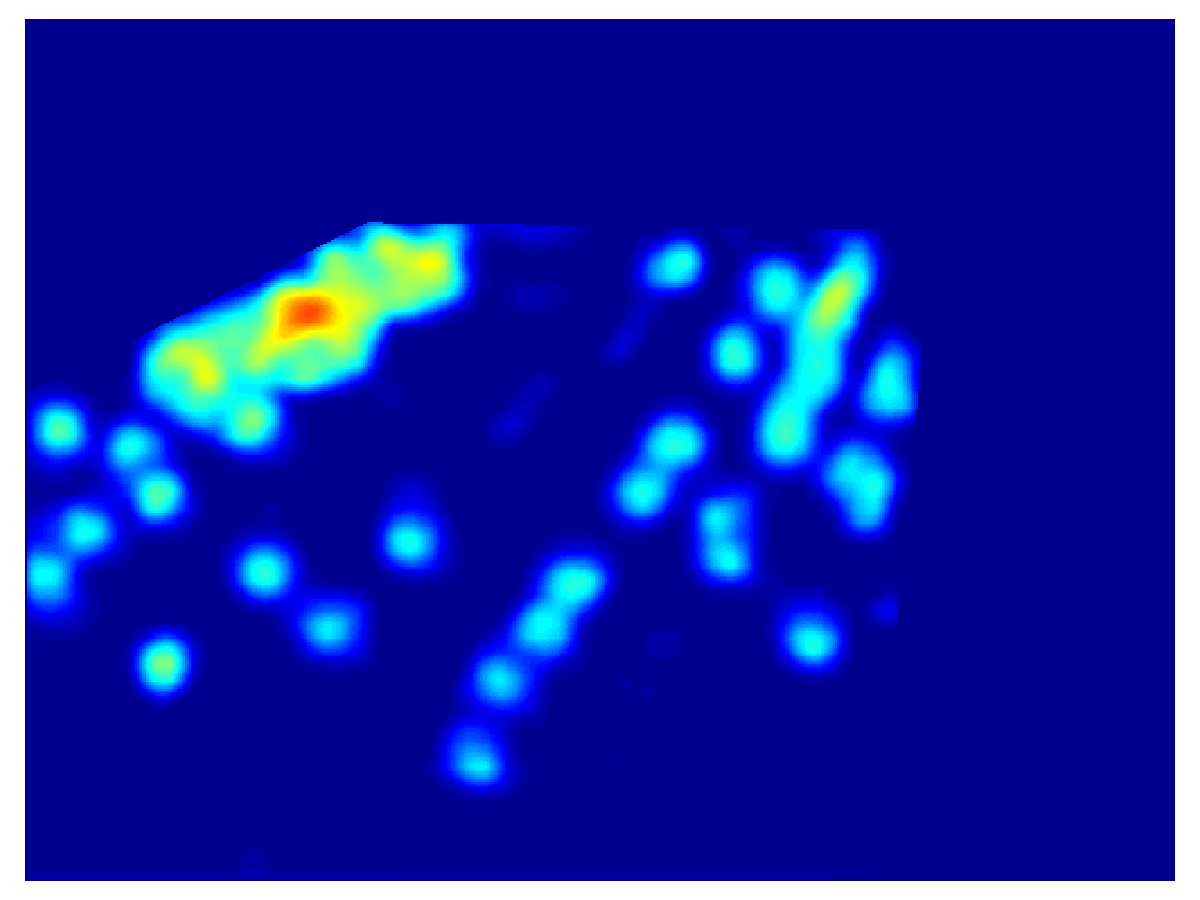} \\
  (g) Image & (h) Ground truth (38) & (i) Prediction (47.52) \\
  \includegraphics[width=0.16\textwidth]{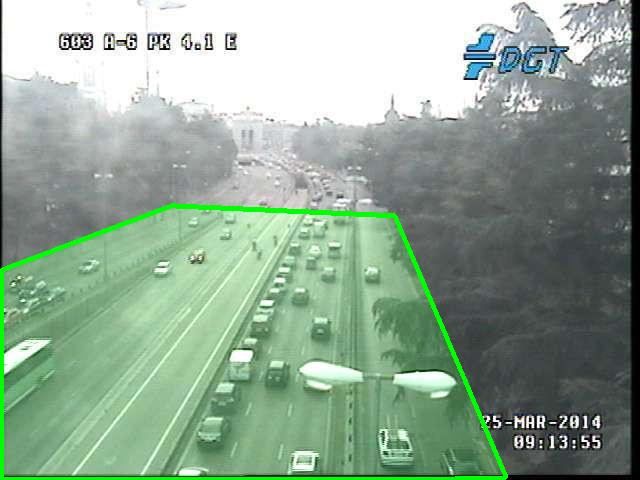} &
  \includegraphics[width=0.16\textwidth]{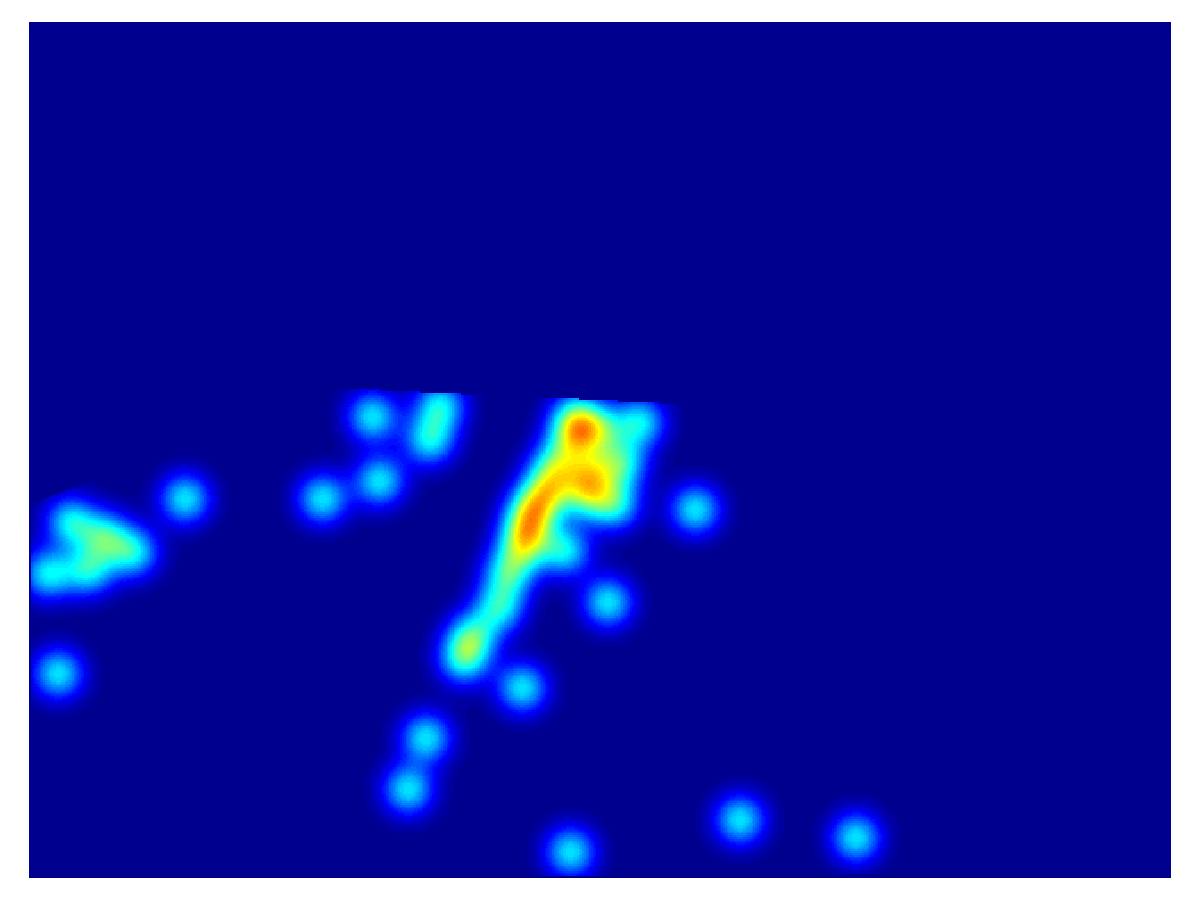} &
  \includegraphics[width=0.16\textwidth]{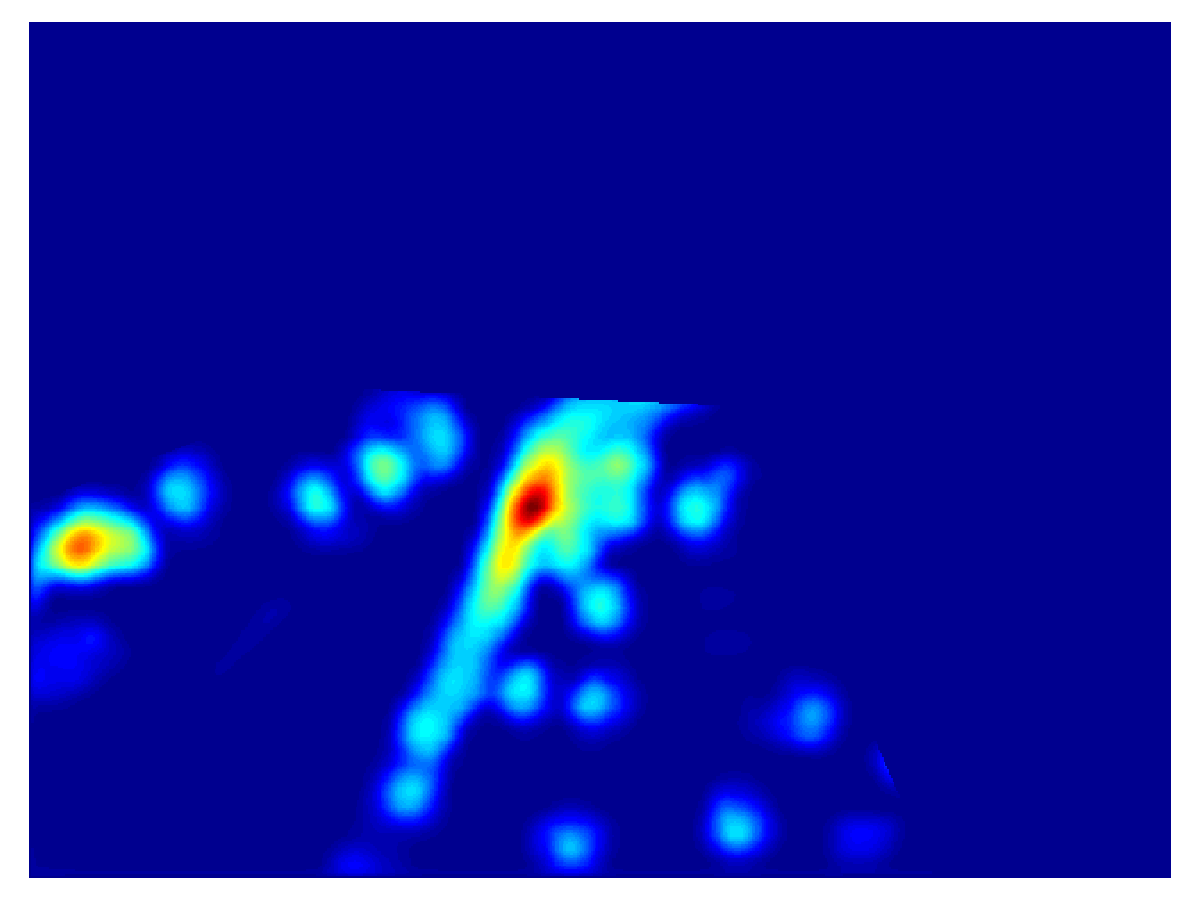}
\end{tabular}
\caption{Example results on TRANCOS using our FCNN-skip model. The number in parenthesis is the count.}
\label{fig:demo_TRANCOS}
\end{figure}

For TRANSCOS, the evaluation metric is the Grid Average Mean absolute Error (GAME) \cite{guerrero2015extremely}, which measures the error within grid sub-regions,
\begin{equation}
  \mathrm{GAME}(L)=\frac{1}{N}\sum_{n=1}^N \sum_{l=1}^{4^L}|\hat{c}_n^l-c_n^l|
\end{equation}
where, $\hat{c}_n^l$ and $c_n^l$ are the estimated count and ground truth count in region $l$ of image ${\cal I}_n$. The number of sub-regions is determined by the level $L$. For $L=0$, GAME(0) is equivalent to MAE.
The results are shown in Table \ref{tab:TRANCOS_count} and example predictions are presented in \reffig{fig:demo_TRANCOS}.
The CNN methods surpass the methods using traditional features by a large margin, especially the last three in Table \ref{tab:TRANCOS_count}.
FCNN-skip has the lowest MAE, GAME(1), and GAME(2) among the methods.

\begin{table*}[tbp]
\small
\centering
\begin{tabular}{|c|c|c|c|c|}
  \hline
  Method                              & GAME(0), MAE  & GAME(1)   & GAME(2)    & GAME(3)  \\
  \hline
  Regression forest \cite{Fiaschi2012} + RGB Norm + Filters & 17.68 & 19.97   & 23.54 & 25.84   \\
  MESA \cite{NIPS2010_4043} + SIFT    & 13.76        & 16.72     & 20.72      & 24.36  \\
  HOG-2 \cite{Sudowe2011}             & 13.29        & 18.05     & 23.65      & 28.41  \\
  \hline
  CNN-patch \cite{Zhang2015}          & 11.24      & 12.36   & 14.51    & 18.67 \\
  Hydra 3s \cite{onoro2016towards}    & 10.99        & 13.75     & 16.69      & 19.32  \\
  MCNN \cite{zhang2016single}         & 7.51         & 9.12      & 11.50      & \bf{15.85} \\
  CNN-pixel (ours)                    & 5.87         & 8.63      & 11.43      & 16.31  \\
  FCNN-skip (ours)                    & \bf{4.61}    & \bf{8.39} & \bf{11.08} & 16.10  \\
  \hline
\end{tabular}
\caption{Evaluation on the TRANCOS car dataset.}
\label{tab:TRANCOS_count}
\end{table*}

\subsection{Detection Experiments} \label{detection_exps}

In this subsection, we test the performance of density maps for people detection.

\subsubsection{Setup}

We use the integer programming method proposed in \cite{Ma_2015_CVPR} (IntProg), as well as 3 baselines described in Section \ref{text:det}:  finding local peaks (Local-max), K-means clustering, and GMM clustering.
We also test our proposed weighted GMM clustering (denoted as GMM-weighted).

\par
For comparison, we run the detection algorithms on a variety of predicted density maps, as well as the ground-truth density map, which provides a reference for detection using these density maps.
We  perform detection on density maps of the same resolution of its input image.
We use the original resolution density maps from CNN-pixel, FCNN-skip, MESA \cite{NIPS2010_4043}, and RR \cite{Arteta2014}.
Since the predicted density maps from MESA and RR are noisy, a Gaussian kernel is used to smooth the density maps (std.~3 for LocalMax and std.~2 for all the other detection methods).
For CNN-patch \cite{Zhang2015} and MCNN \cite{zhang2016single}, we upsample the reduced-resolution density maps to the original resolution using bicubic interpolation.
As MCNN shows strong performance in the counting task, but only produces reduced resolution density maps, we also test a variant of MCNN that predicts a full-resolution density maps (denoted as MCNN-up). Similar to FCNN-skip, two learned upsampling-convolution layers, each with a $3\times3\times1$ filter, are used to upsample the MCNN predicted density map.
We follow the same evaluation procedure as \cite{Ma_2015_CVPR,Fiaschi2012}.
Detections are {\em uniquely} paired with ground-truth locations within a matching distance, and precision, recall, and F1 score are calculated.

\subsubsection{Results}

\begin{table*}[tbhp]
\small
\setlength{\tabcolsep}{2.5pt}
\centering
\small
  \begin{tabular}{|c|ccc|ccc|ccc|ccc|ccc|}
    \hline
    & \multicolumn{3}{|c|}{IntProg} & \multicolumn{3}{|c|}{GMM-weighted} & \multicolumn{3}{|c|}{GMM} &\multicolumn{3}{|c|}{K-means} & \multicolumn{3}{|c|}{Local-max} \\
    \hline
    Density map & R\%  & P\%  & F1\%  & R\%  & P\%  & F1\%  & R\%  & P\%  & F1\%  & R\%  & P\%  & F1\%  & R\%  & P\%  & F1\%  \\
    \hline
    MESA \cite{NIPS2010_4043} & 92.39 & 91.18 & 91.78 & 91.31 & 86.69 & 88.94 & 81.55 & 78.78 & 80.14 & 85.34 & 81.75 & 83.50 & 81.01 & 86.58 & \bf{83.70} \\
    Ridge regression \cite{Arteta2014} & 85.04 & 89.26 & 87.10 & 86.86 & 84.27 & 85.55 & 86.80 & 80.59 & \bf{83.58} & 87.74 & 81.46 & \bf{84.49} & 72.90 & 88.00 & 79.74 \\
    \hline
    CNN-patch \cite{Zhang2015} & 59.28 & 85.82 & 70.12 & 83.34 & 78.45 & 80.82 & 51.57 & 49.28 & 50.40 & 55.99 & 53.49 & 54.71 & 39.55 & 83.64 & 53.70 \\
    MCNN \cite{zhang2016single} & 80.27 & 93.90 & 86.55 & 89.47 & 89.78 & 89.63 & 79.36 & 78.82 & 79.09 & 81.82 & 81.26 & 81.54 & 70.44 & 85.46 & 77.22 \\
    MCNN-up          & 83.87 & 94.87 & 89.03 & 89.46 & 90.84 & 90.14 & 81.20 & 81.68 & 81.44 & 83.70 & 84.19 & 83.95 & 72.49 & 83.34 & 77.54 \\
    FCNN-skip (ours) & 88.29 & 88.19 & 88.24 & 91.77 & 86.20 & 88.90 & 82.44 & 76.92 & 79.59 & 85.40 & 79.69 & 82.45 & 74.77 & 84.54 & 79.35 \\
    CNN-pixel (ours) & 90.19 & 95.75 & \bf{92.89} & 91.57 & 89.82 & \bf{90.69} & 80.05 & 78.38 & 79.20 & 82.68 & 80.95 & 81.81 & 71.19 & 87.12 & 78.36 \\
    CNN-pixel-VS     & 79.26 & 94.64 & 86.27 & 87.40 & 92.03 & 89.66 & 75.04 & 79.61 & 77.26 & 77.54 & 82.26 & 79.83 & 62.29 & 91.45 & 74.11  \\
    \hline
    CNN-pixel (ours) [full]  & 88.02 & 97.45 & 92.49 & 89.90 & 92.46 & 91.16 & 78.55 & 80.84 & 79.68 & 80.80 & 83.15 & 81.96 & 73.11 & 86.78 & 79.36 \\
    \hline
    GT density map  & 97.41 & 98.51 & 97.96 & 94.76 & 94.56 & 94.66 & 85.25 & 86.64 & 85.94 & 87.78 & 89.05 & 88.41 & 80.51 & 84.88 & 82.64 \\
    \hline
  \end{tabular}
\caption{Detection performance on UCSD dataset (``max'' split) using density maps with different detection methods. For comparison, SIFT-SVM \cite{NIPS2010_4043} obtains an F1 score of 68.46 and region-SVM \cite{Arteta2013a} obtains an F1 score of 89.53. [full] denotes training with the full 800 frames of the training set. Bold numbers indicate best performance for each detection method among the density maps.}
\label{tab:ucsd_det}
\end{table*}

\par
Table \ref{tab:ucsd_det} presents the detection results on UCSD using the various density maps and detection methods.
Using IntProg with density maps from CNN-pixel or MESA yields the best F1 score.
Although the density maps from MESA look ``noisy'' (e.g., see \reffig{fig:dmap-comp}c), actually the location information is preserved well by the MESA criteria, as its BBMAE is 2nd lowest among the methods (see \reffig{fig:bbox}b).
However, MESA density maps are less compact than most other CNN-based methods (lower BBDR in \reffig{fig:bbox}a), and cannot maintain the monotonicity (per-pixel reproduction \reffig{fig:scatter_plot}), and thus detection with GMM-weighted is worse than MCNN and MCNN-up.
Detection with CNN-pixel has higher precision than MESA, due to the more compact density maps of CNN-pixel, but also lower recall.

The detection result for FCNN-skip is worse than CNN-pixel because its density maps are less compact (lower BBDR in \reffig{fig:bbox}a) and less isolated due to the upsampling process.
While the upsampling and skip connections compensate for the loss of spatial information, there is still a degradation in detection accuracy.

Using the upsampling-convolution layers with MCNN (MCNN-up) improves the F1 score on all methods compared to MCNN, although the counting performance decreases slightly (1.37 vs. 1.32).
The improved detection performance of MCNN-up is due to better compactness (higher BBDR) and localization (lower BBMAE) of its density maps, compared to MCNN (see \reffig{fig:bbox}).
Because the objects in UCSD can be smaller than 20 pixels, a downsample factor of 4 has a large influence on localization -- a full resolution density map is crucial for good localization performance, as compared to bicubic upsampling of the reduced-resolution density map.

\par
Detection accuracy on RR maps is not as high as those by MESA or CNN-pixel with IntProg and GMM-weighted.
To maintain positive density values, RR uses a heuristic to remove feature dimensions with negative weight during the training process.
Most of these situations affect the boundary regions of the density blobs (see \reffig{fig:dmap-comp}d), resulting in very compact density maps (highest BBDR in \reffig{fig:bbox}a).
Yet, the disadvantage of this feature removal is that it shifts the center of mass of the density blobs, resulting in  worse localization metrics (lower BBMAE, see \reffig{fig:bbox}b), which affects the detection accuracy with IntProg and GMM-weighted.
On the other hand, the density blobs are very compact, which favors the K-means and GMM detection methods, which only consider the shape of the density blobs.

\par
Although CNN-patch is suitable for crowd counting, the detection results are worse than other density maps. CNN-patch generates a reduced-resolution density maps and also averages overlapping density patch predictions, which smooths out local density peaks (lowest BBDR in \reffig{fig:bbox}a), thus making localization of each person more difficult.

We note that two attributes are important for good detection performance using the tested detection methods:
1) accurate peaks with good monotonicity, otherwise the detections easily drifts;
2) having more compact (higher BBDR) and isolated peaks reduces the search space and makes detection of every individual object easier.
Using higher quality density maps (such as from CNN-pixel), the simple GMM-weighted method, which considers both the shape and density of the blobs, achieves very similar performance to the more complex IntProg.
Finally, there exists a large gap between detection on the ground-truth density map and the predicted density maps (e.g., using IntProg, F1 of 97.96 vs 92.89).
Hence, there still is room to improve density map methods such that the location information is better preserved.

\subsection{Tracking Experiments} \label{tracking_exps}

We next test the ability of density maps to improve pedestrian tracking algorithms.

\subsubsection{Setup}
We test the fusion method described in Section \ref{text:tracking} to combine the response map of the kernel correlation filter (KCF) with the crowd density map.
The position in the resulting map with the maximum response is used as the tracked location,
which is then passed to the tracker for online updating the KCF appearance model.
We test KCF fusion with the same density maps used for the detection experiment (all trained on UCSD ``max'' split), as well as the original KCF without fusion.
The performance of single person tracking is evaluated on each person in the testing set of the UCSD dataset (1200 frames).
In particular, we plot a precision-threshold curve, which shows the percentage of frames where the tracking error is within a distance threshold.
The tracker is initialized with the ground truth location of a person after they have fully entered the video.

\begin{figure}[tbp]
\centering
\includegraphics[width=0.44\textwidth]{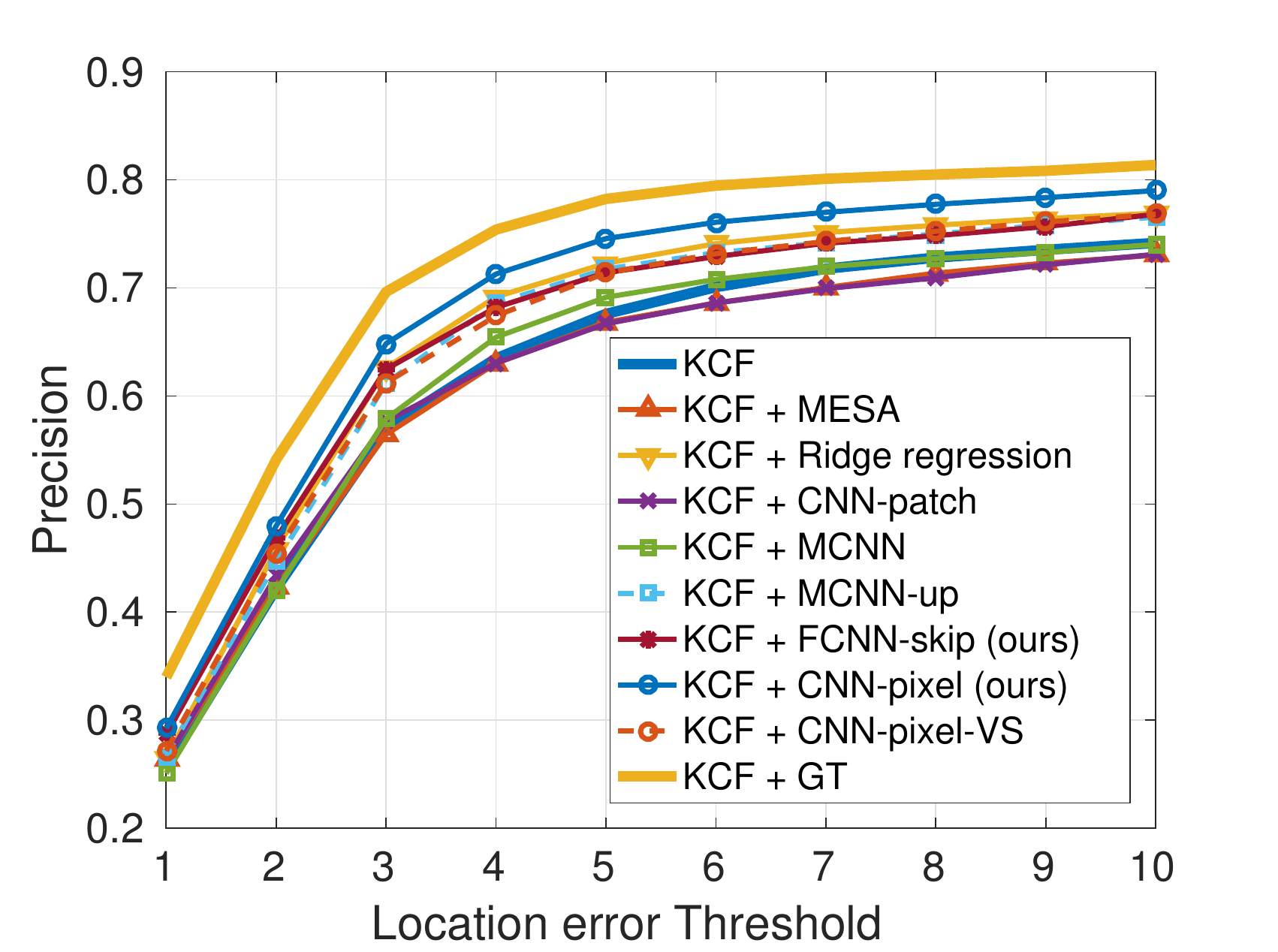}
\caption{Single object tracking performance on UCSD.
}
\label{fig:tracking_res}
\end{figure}

\subsubsection{Results}
\reffig{fig:tracking_res} shows the tracking performance.
Fusing the density map with the KCF response map can improve tracking performance.
Specifically, using CNN-pixel yields the largest increase in precision compared to other methods (e.g., P@4 of 0.713 for CNN-pixel vs. 0.692 for RR), due to its density maps having better localization metric (lower BBMAE, see \reffig{fig:bbox}b) and its detections are temporally most stable (low ED and EDD in Table~\ref{tab:det_combo}).
Fusion with the CNN-patch density maps does not improve the tracker because those density maps are too smooth, and thus the fused map does not change much.
Fusion with MCNN-up improves the tracking performance over MCNN,
because MCNN-up density maps are more compact and better localized (see \reffig{fig:bbox}), resulting in temporally more stable detections (see Table~\ref{tab:det_combo}).
Again, this shows that original resolution density maps are preferred for localization tasks.
Finally, fusing the ground-truth density map with KCF yields the best tracking results, which marks the upper bound of the performance for this type of fusion.

\subsection{Summary}

In summary, in the experiments, we have evaluated density maps for counting, detection, and tracking.
For counting, the resolution of the density map does not affect much the counting accuracy. Methods that produce reduced-resolution maps, e.g., MCNN, can perform well.
In contrast, using original-resolution density maps gives better accuracy for localization tasks, such as detection and tracking.
In particular, per-pixel generation of original-resolution density maps without using upsampling (e.g., CNN-pixel) have higher fidelity
(better compactness and localization metrics, BBDR and BBMAE) than those that are upsampled from reduced resolution maps (e.g., FCNN-skip, MCNN, MCNN-up), leading to better detection and tracking performance.
For FCNN-skip, MCNN, and MCNN-up, downsampling in the CNN obfuscate the true position of the people in the reduced-resolution maps. This loss of spatial information can only be partially compensated using learned upsampling and skip connections.

\subsection{Training and Architecture Variations} \label{Variations}

Next we report results of using various common strategies for training and alternative architectures for CNN-pixel and FCNN-skip.
Experiments were run on the UCSD dataset using the ``max'' split, and the results are summarized in Table \ref{tab:variations}.

\begin{table}[tbhp]
\centering
\scriptsize
\begin{tabular}{|l|c|c|}
  \hline
  Method    & Variation                                 & MAE   \\
  \hline
  CNN-pixel & --                                        & 1.26  \\
  CNN-pixel & varying std. with perspective       & 1.48  \\
  CNN-pixel & fine-tuning with only regression task     & 1.20  \\
  CNN-pixel & feature combo                             & 1.21  \\
  CNN-pixel & 256 neurons for the first FC              & 1.38  \\
  CNN-pixel & 4 or 6 convolution layers                 & 1.26, 1.38 \\
  CNN-pixel & 6 or 12 residual blocks \cite{He2015}     & 1.26, 1.34\\
  CNN-pixel & 2 dense blocks (2, 4 or 8 layers each) \cite{huang2017densely}  & 2.77, 1.82, 2.64 \\
  \hline
  FCNN-skip &  --                                       & 1.24  \\
  FCNN-skip & only pixel-wise loss                      & 1.41  \\
  FCNN      & only pixel-wise loss                      & 1.54  \\
  FCNN      & only count loss                           & 1.82  \\
  FCNN      & hole convolution  \cite{Chen2014,Li2014}  & 1.93  \\
  \hline
\end{tabular}
\caption{Comparison of variations in training and architectures of CNN-pixel and FCNN-skip. All the experiments are on the UCSD dataset using the ``max'' split.
}
\label{tab:variations}
\end{table}

\mysubsubsection{Data augmentation}
Because we extract all the possible image patches to train the network, no random translations are needed when cropping patches.
We also added small random Gaussian noise and horizontally flipped the patches, but no obvious improvement was observed.
Most likely this is because the dense patch sampling already covers many useful permutations of the input.

\mysubsubsection{CNN-pixel fine-tuning with only the regression task}
As the classification task is an auxiliary task for guiding the CNN during training,
it is possible that it could adversely affect the training of the density regressor.
Starting from a well-trained network, we removed the classification task and fine-tuned the network only using the regression task.
However, with fine-tuning we obtained similar regression performance (MAE: 1.20).
Most likely this is because counting (implemented as classification) and density prediction are related tasks, which can share common image features, so it is not necessary to explicitly fine-tune the density regressor alone.

\mysubsubsection{CNN-pixel with combined features from different convolution layers}
A commonly used strategy to improve the performance of CNNs is to combine features from different convolution layers \cite{Sijin,li20143d}.
We use a fully connected layer (in contrast to the convolutional layer used in the FCNN-skip adaptations) to collect features from different convolutional layer and concatenate them together.
Similar performance to the original CNN-pixel was observed (MAE: 1.21).

\mysubsubsection{CNN-pixel with more convolution layers}
Recently, very deep networks have been used for high-level classification and semantic segmentation tasks \cite{AlexNet,Simonyan14c,Szegedy_2015_CVPR,He2015}.
However, we note that density estimation task is a mid-level task, and thus does not necessarily require very deep networks.
We test CNN-pixel variations with 4 or 6 convolution layers, with 2 pooling layers. No gain is observed using 4 convolution layers (MAE: 1.26), and performance decreases slightly (MAE: 1.38) when using 6 convolution layers.
We also tested a ResNet version, which replaces the convolution layers with residual blocks, and did not see better performance (MAE: 1.26).
We also tried several versions of DenseNet \cite{huang2017densely} with 2, 4, or 8 dense layers in each dense block. The more complicated network has poor performance on the counting task, possibly due to the limited training data relative to the network size.
A similar trend is also observed in \cite{Walach2016}.

\mysubsubsection{CNN-pixel with smaller fully-connected layer}
CNN-pixel uses 512 neurons in the first fully-connected (FC) layer which seems large for such a shallow network.
We test a version of CNN-pixel with fewer neurons in the first FC layer. Using 256 neurons gives worse performance (MAE: 1.38),
possibly because more neurons are needed to capture all the appearance variations associated with the same density value.

\mysubsubsection{FCNN-skip with only pixel-wise loss}
Our FCNN-skip is trained with both the patch-wise count loss and a pixel-wise loss.
Removing the count loss from the training leads to higher error (MAE: 1.41).
This shows the benefit of including the count loss, which focuses on reducing systematic counting errors, which cannot be well reflected in the pixel-wise loss.

\mysubsubsection{FCNN without spatial compensation}
We also compared  FCNN-skip with an FCNN without skip branches (denoted as FCNN).
Here we only train with the pixel-wise loss, and the error increases when no skip branches are used.

\begin{figure}[tbp]
\centering
\footnotesize
\begin{tabular}{@{}c@{\hspace{1mm}}c@{\hspace{1mm}}c@{}}
  (a) Count Loss & (b) Pixel Loss & (c) Pixel \& Count Loss\\
  \includegraphics[width=0.15\textwidth]{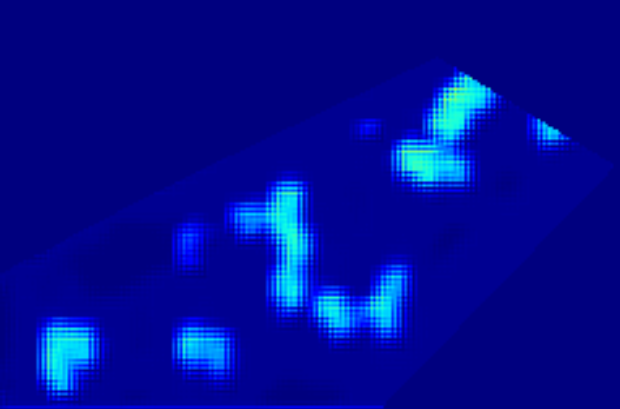} &
  \includegraphics[width=0.15\textwidth]{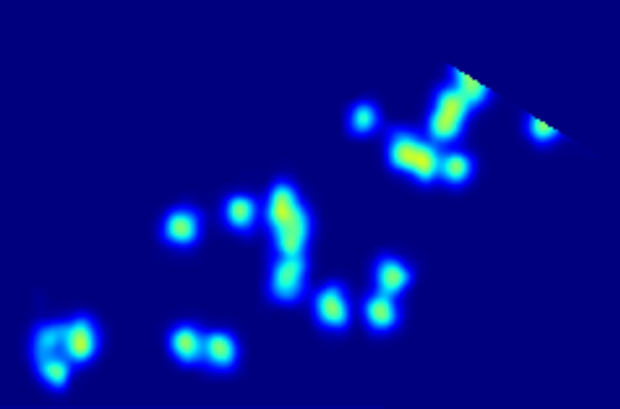} &
  \includegraphics[width=0.15\textwidth]{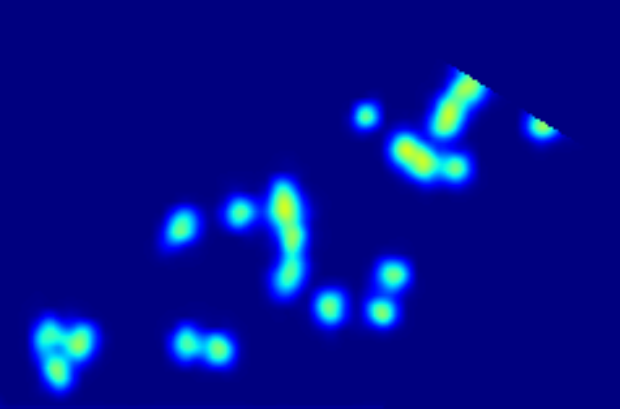}
\end{tabular}
\caption{Example of using different loss functions to train FCNN.
}
\label{fig:demo_loss_comp}
\end{figure}

\mysubsubsection{FCNN with only count loss}
Next, we train FCNN with only the count loss, which results in the predicted density map spreading out, making the high density and low density regions more similar (see \reffig{fig:demo_loss_comp}).
This leads to poor counting performance (MAE: 1.82), and hence the pixel-wise loss is required to maintain the structure of the density map.

\mysubsubsection{Effect of Multi-task learning}

We compared the effect of different $\lambda_2$ on multi-task learning. For $\lambda_2=$\{0, 0.1, 1, 10, 100\}, CNN-pixel gets MAE of \{1.41, 1.27, 1.26, 1.44, 1.47\}. Using $\lambda_2=0.1$ gives similar performance to $\lambda_2=1$, while other settings decrease the performance. Similar observations are also reported in \cite{Sijin}.

\mysubsubsection{Effect of varying Gaussian widths}

We have conducted experiments to understand the effect of including perspective in the ground-truth densities for UCSD. As in \cite{Zhang2015}, the standard deviation of the Gaussian is varied using the perspective map.
Results using CNN-pixel with the new density maps (denoted as ``CNN-pixel-VS'') for counting, detection, and tracking are shown in Tables \ref{tab:ucsd_count} and \ref{tab:ucsd_det}, and \reffig{fig:tracking_res}.
The counting results using CNN-pixel-VS are worse than CNN-pixel (MAE 1.48 vs 1.24).
Detection performance also is worse, since IntProg and GMM-weight use the predicted count as either a constraint or the number of clusters.
For tracking, CNN-pixel-VS also performs worse than CNN-pixel, mainly because the spread out density maps have lower values, which decreases the influence of the density map on the tracking response map.

\mysubsubsection{FCNN using hole convolution}
Another method to get a full resolution density map is to use ``hole convolution''  \cite{Chen2014,Li2014}, which changes all the stride sizes to one and adds zeros into the filters of the convolutional and pooling layers.
We first adapt our CNN-pixel model to FCNN. After the parameters have been trained, we use the hole algorithm to predict a density map with the same resolution as the input.
This approach has poor performance (MAE: 1.93) while taking more time than FCNN with up-sampling because it need to compute responses for every image pixel.
In contrast, using FCNN with up-sampling is more suitable since
it encourages smooth density maps.

\section{Conclusion}

In this work, we compare crowd density maps, produced by different methods, on several crowd analysis tasks, including counting, detection, and tracking.
While reduced-resolution density maps produced by fully-convolutional NN (e.g., MCNN) perform well at counting, their accuracy diminished at localization tasks due to the loss of spatial resolution, which could not be completely recovered using upsampling and skip connections.
In contrast, dense pixel-prediction of a full resolution density map, using CNN-pixel, produced the highest quality density map for localization tasks, with slight degradation for the counting task. However, dense prediction suffers from higher computational complexity, compared to fully-convolutional networks.

\par
We also proposed several metrics for measuring aspects of the density maps to explain why those density maps with similar counting accuracy can perform differently on detection and tracking. These metrics can help to  guide future research in designing density map methods to estimate high quality density maps for both counting and localization tasks.



%







\small
%


\bibliographystyle{IEEEtran}
\bibliography{ref_IEEEtran_formated}

\begin{thebibliography}{10}
\providecommand{\url}[1]{#1}
\csname url@samestyle\endcsname
\providecommand{\newblock}{\relax}
\providecommand{\bibinfo}[2]{#2}
\providecommand{\BIBentrySTDinterwordspacing}{\spaceskip=0pt\relax}
\providecommand{\BIBentryALTinterwordstretchfactor}{4}
\providecommand{\BIBentryALTinterwordspacing}{\spaceskip=\fontdimen2\font plus
\BIBentryALTinterwordstretchfactor\fontdimen3\font minus
  \fontdimen4\font\relax}
\providecommand{\BIBforeignlanguage}[2]{{%
\expandafter\ifx\csname l@#1\endcsname\relax
\typeout{** WARNING: IEEEtran.bst: No hyphenation pattern has been}%
\typeout{** loaded for the language `#1'. Using the pattern for}%
\typeout{** the default language instead.}%
\else
\language=\csname l@#1\endcsname
\fi
#2}}
\providecommand{\BIBdecl}{\relax}
\BIBdecl

\bibitem{Chan2008}
A.~B. Chan, Z.-S.~J. Liang, and N.~Vasconcelos, ``Privacy preserving crowd
  monitoring: Counting people without people models or tracking,'' in
  \emph{Proceedings of the IEEE Conference on Computer Vision and Pattern
  Recognition}, 2008.

\bibitem{Ryan2009}
D.~Ryan, S.~Denman, C.~Fookes, and S.~Sridharan, ``Crowd counting using
  multiple local features,'' in \emph{Digital Image Computing: Techniques and
  Applications}, 2009.

\bibitem{NIPS2010_4043}
V.~Lempitsky and A.~Zisserman, ``Learning to count objects in images,'' in
  \emph{Advances in neural information processing systems}, 2010.

\bibitem{Arteta2014}
C.~Arteta, V.~Lempitsky, J.~A. Noble, and A.~Zisserman, ``Interactive object
  counting,'' in \emph{European Conference on Computer Vision}, 2014.

\bibitem{Idrees2013}
H.~Idrees, I.~Saleemi, C.~Seibert, and M.~Shah, ``Multi-source multi-scale
  counting in extremely dense crowd images,'' in \emph{Proceedings of the IEEE
  Conference on Computer Vision and Pattern Recognition}, 2013.

\bibitem{Wang2015}
C.~Wang, H.~Zhang, L.~Yang, S.~Liu, and X.~Cao, ``{Deep People Counting in
  Extremely Dense Crowds},'' in \emph{Proceedings of the 23rd ACM international
  conference on Multimedia}, 2015.

\bibitem{Fiaschi2012}
L.~Fiaschi, R.~Nair, U.~Koethe, and F.~A. Hamprecht, ``Learning to count with
  regression forest and structured labels,'' in \emph{International Conference
  on Pattern Recognition}, 2012.

\bibitem{Zhang2015}
C.~Zhang, H.~Li, X.~Wang, and X.~Yang, ``Cross-scene crowd counting via deep
  convolutional neural networks,'' in \emph{Proceedings of the IEEE Conference
  on Computer Vision and Pattern Recognition}, 2015.

\bibitem{Pham2015}
V.-Q. Pham, T.~Kozakaya, O.~Yamaguchi, and R.~Okada, ``{COUNT Forest: CO-voting
  Uncertain Number of Targets using Random Forest for Crowd Density
  Estimation},'' in \emph{Proceedings of the IEEE International Conference on
  Computer Vision}, 2015.

\bibitem{zhang2016single}
Y.~Zhang, D.~Zhou, S.~Chen, S.~Gao, and Y.~Ma, ``Single-image crowd counting
  via multi-column convolutional neural network,'' in \emph{Proceedings of the
  IEEE Conference on Computer Vision and Pattern Recognition}, 2016.

\bibitem{Ma_2015_CVPR}
Z.~Ma, L.~Yu, and A.~B. Chan, ``Small instance detection by integer programming
  on object density maps,'' in \emph{Proceedings of the IEEE Conference on
  Computer Vision and Pattern Recognition}, 2015.

\bibitem{Rodriguez2011}
M.~Rodriguez, I.~Laptev, J.~Sivic, and J.-Y. Audibert, ``Density-aware person
  detection and tracking in crowds,'' in \emph{Proceedings of the IEEE
  International Conference on Computer Vision}, 2011.

\bibitem{Xie2015}
W.~Xie, J.~A. Noble, and A.~Zisserman, ``Microscopy cell counting with fully
  convolutional regression networks,'' in \emph{MICCAI 1st Workshop on Deep
  Learning in Medical Image Analysis}, 2015.

\bibitem{onoro2016towards}
D.~Onoro-Rubio and R.~J. L{\'o}pez-Sastre, ``Towards perspective-free object
  counting with deep learning,'' in \emph{European Conference on Computer
  Vision}, 2016.

\bibitem{Arteta2016}
C.~Arteta, V.~Lempitsky, and A.~Zisserman, ``{Counting in The Wild},'' in
  \emph{European Conference on Computer Vision}, 2016.

\bibitem{wu2005detection}
B.~Wu and R.~Nevatia, ``Detection of multiple, partially occluded humans in a
  single image by bayesian combination of edgelet part detectors,'' in
  \emph{Proceedings of the IEEE International Conference on Computer Vision},
  2005.

\bibitem{brostow2006unsupervised}
G.~J. Brostow and R.~Cipolla, ``Unsupervised bayesian detection of independent
  motion in crowds,'' in \emph{Proceedings of the IEEE Conference on Computer
  Vision and Pattern Recognition}, 2006.

\bibitem{rabaud2006counting}
V.~Rabaud and S.~Belongie, ``Counting crowded moving objects,'' in
  \emph{Proceedings of the IEEE Conference on Computer Vision and Pattern
  Recognition}, 2006.

\bibitem{Kong2006}
D.~Kong, D.~Gray, and H.~Tao, ``A viewpoint invariant approach for crowd
  counting,'' in \emph{International Conference on Pattern Recognition}, 2006.

\bibitem{Cho1999}
S.-Y. Cho, T.~W. Chow, and C.-T. Leung, ``A neural-based crowd estimation by
  hybrid global learning algorithm,'' \emph{IEEE Transactions on Systems, Man,
  and Cybernetics, Part B (Cybernetics)}, vol.~29, no.~4, 1999.

\bibitem{Chen2013}
K.~Chen, S.~Gong, T.~Xiang, and C.~Loy, ``Cumulative attribute space for age
  and crowd density estimation,'' in \emph{Proceedings of the IEEE Conference
  on Computer Vision and Pattern Recognition}, 2013.

\bibitem{Chen2012}
K.~Chen, C.~C. Loy, S.~Gong, and T.~Xiang, ``Feature mining for localised crowd
  counting,'' in \emph{Proceedings of BMVC}, 2012.

\bibitem{Chan2012}
A.~B. Chan and N.~Vasconcelos, ``Counting people with low-level features and
  bayesian regression,'' \emph{IEEE Transactions on Image Processing}, vol.~21,
  no.~4, 2012.

\bibitem{Walach2016}
E.~Walach and L.~Wolf, ``Learning to count with cnn boosting,'' in
  \emph{European Conference on Computer Vision}, 2016.

\bibitem{Long2015}
J.~Long, E.~Shelhamer, and T.~Darrell, ``Fully convolutional networks for
  semantic segmentation,'' in \emph{Proceedings of the IEEE Conference on
  Computer Vision and Pattern Recognition}, 2015.

\bibitem{AlexNet}
A.~Krizhevsky, I.~Sutskever, and G.~E. Hinton, ``Imagenet classification with
  deep convolutional neural networks,'' in \emph{Advances in neural information
  processing systems}, 2012.

\bibitem{Szegedy_2015_CVPR}
C.~Szegedy, W.~Liu, Y.~Jia, P.~Sermanet, S.~Reed, D.~Anguelov, D.~Erhan,
  V.~Vanhoucke, and A.~Rabinovich, ``Going deeper with convolutions,'' in
  \emph{Proceedings of the IEEE Conference on Computer Vision and Pattern
  Recognition}, 2015.

\bibitem{Simonyan14c}
K.~Simonyan and A.~Zisserman, ``Very deep convolutional networks for
  large-scale image recognition,'' in \emph{International Conference on
  Learning Representations}, 2015.

\bibitem{He2015}
K.~He, X.~Zhang, S.~Ren, and J.~Sun, ``Deep residual learning for image
  recognition,'' in \emph{Proceedings of the IEEE Conference on Computer Vision
  and Pattern Recognition}, 2016.

\bibitem{Sijin}
S.~Li, Z.-Q. Liu, and A.~B. Chan, ``Heterogeneous multi-task learning for human
  pose estimation with deep convolutional neural network,'' \emph{International
  Journal of Computer Vision}, 2015.

\bibitem{li20143d}
S.~Li and A.~B. Chan, ``3d human pose estimation from monocular images with
  deep convolutional neural network,'' in \emph{Asian Conference on Computer
  Vision}, 2014.

\bibitem{Niu2016}
Z.~Niu, M.~Zhou, L.~Wang, X.~Gao, and G.~Hua, ``Ordinal regression with
  multiple output cnn for age estimation,'' in \emph{Proceedings of the IEEE
  Conference on Computer Vision and Pattern Recognition}, 2016, pp. 4920--4928.

\bibitem{Rothe2015}
R.~Rothe, R.~Timofte, and L.~Van~Gool, ``Dex: Deep expectation of apparent age
  from a single image,'' in \emph{Proceedings of the IEEE International
  Conference on Computer Vision Workshops}, 2015, pp. 10--15.

\bibitem{CS231n}
``Cs231n: Convolutional neural networks for visual recognition,''
  \url{http://cs231n.github.io/neural-networks-2/}, 2018.

\bibitem{Chen2014}
L.-C. Chen, G.~Papandreou, I.~Kokkinos, K.~Murphy, and A.~L. Yuille,
  ``{Semantic Image Segmentation with Deep Convolutional Nets and Fully
  Connected CRFs},'' in \emph{International Conference on Learning
  Representations}, 2015.

\bibitem{Torr2015}
S.~Zheng, S.~Jayasumana, B.~Romera-Paredes, V.~Vineet, Z.~Su, D.~Du, C.~Huang,
  and P.~H. Torr, ``Conditional random fields as recurrent neural networks,''
  in \emph{Proceedings of the IEEE International Conference on Computer
  Vision}, 2015.

\bibitem{henriques2015high}
J.~F. Henriques, R.~Caseiro, P.~Martins, and J.~Batista, ``High-speed tracking
  with kernelized correlation filters,'' \emph{IEEE Transactions on Pattern
  Analysis and Machine Intelligence}, vol.~37, no.~3, 2015.

\bibitem{henriques2012exploiting}
------, ``Exploiting the circulant structure of tracking-by-detection with
  kernels,'' in \emph{European Conference on Computer Vision}, 2012.

\bibitem{guerrero2015extremely}
R.~Guerrero-G{\'o}mez-Olmedo, B.~Torre-Jim{\'e}nez, R.~L{\'o}pez-Sastre,
  S.~Maldonado-Basc{\'o}n, and D.~Onoro-Rubio, ``Extremely overlapping vehicle
  counting,'' in \emph{Iberian Conference on Pattern Recognition and Image
  Analysis}.\hskip 1em plus 0.5em minus 0.4em\relax Springer, 2015, pp.
  423--431.

\bibitem{rumelhart1988learning}
D.~E. Rumelhart, G.~E. Hinton, and R.~J. Williams, ``Learning representations
  by back-propagating errors,'' \emph{Nature}, vol. 323, no. 6088, p. 533,
  1986.

\bibitem{an2007face}
S.~An, W.~Liu, and S.~Venkatesh, ``Face recognition using kernel ridge
  regression,'' in \emph{Proceedings of the IEEE Conference on Computer Vision
  and Pattern Recognition}, 2007, pp. 1--7.

\bibitem{Sudowe2011}
P.~Sudowe and B.~Leibe, ``Efficient use of geometric constraints for
  sliding-window object detection in video,'' in \emph{Computer Vision
  Systems}, 2011.

\bibitem{Arteta2013a}
C.~Arteta, V.~Lempitsky, J.~Noble, and A.~Zisserman, ``Learning to detect
  partially overlapping instances,'' in \emph{Proceedings of the IEEE
  Conference on Computer Vision and Pattern Recognition}, 2013.

\bibitem{huang2017densely}
G.~Huang, Z.~Liu, K.~Q. Weinberger, and L.~van~der Maaten, ``Densely connected
  convolutional networks,'' in \emph{Proceedings of the IEEE Conference on
  Computer Vision and Pattern Recognition}, vol.~1, no.~2, 2017, p.~3.

\bibitem{Li2014}
H.~Li, R.~Zhao, and X.~Wang, ``Highly efficient forward and backward
  propagation of convolutional neural networks for pixelwise classification,''
  \emph{arXiv preprint arXiv:1412.4526}, 2014.

\end{thebibliography}









\begin{IEEEbiography}[{\includegraphics[width=1in,height=1.25in,clip,keepaspectratio]{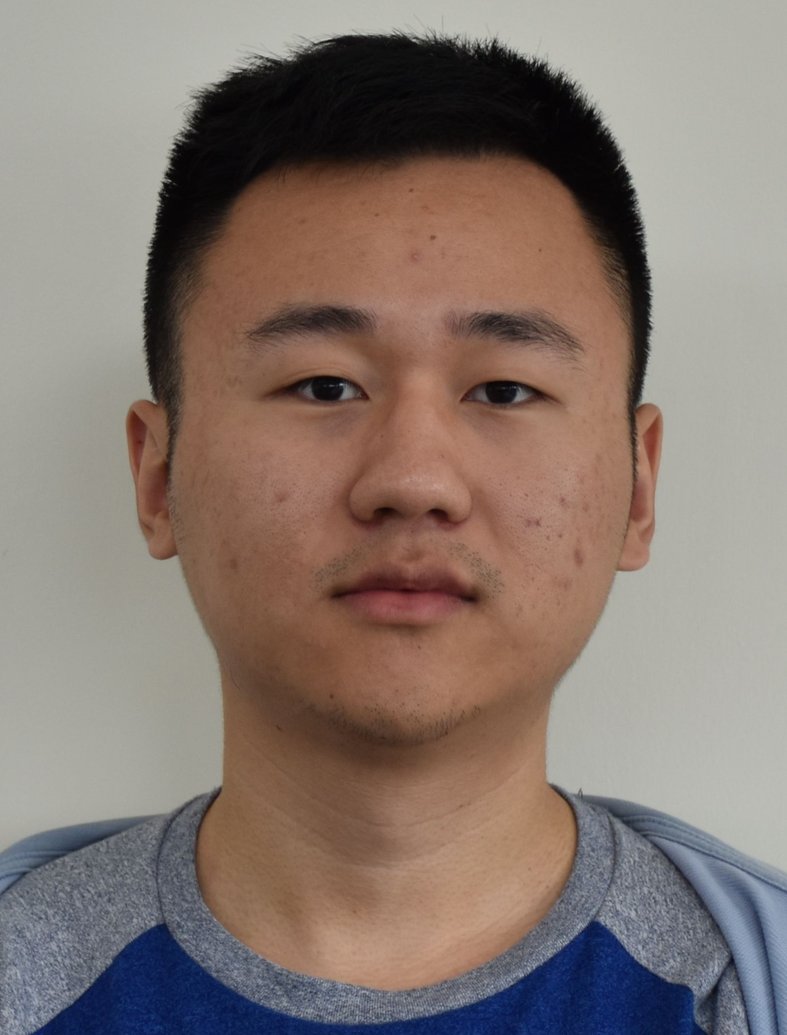}}]
{Di Kang} received the B.Eng. degree in Electronic Engineering and Information Science (EEIS) from the University of Science and Technology of China (USTC) in 2014. He is currently working towards the PhD degree in Computer Science at City University of Hong Kong. His research interests include computer vision, crowd counting and deep learning.
\end{IEEEbiography}

\begin{IEEEbiography}[{\includegraphics[width=1in,height=1.25in,clip,keepaspectratio]{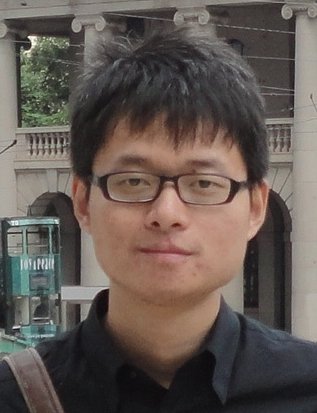}}]
{Zheng Ma} (M'16) received the B.Eng. and M.Sc. from Xi'an Jiaotong University in 2007 and 2011, respectively, and the Ph.D. degree in computer science from City University of Hong Kong. He is currently a SenseTime research scientist. His research interests include computer vision, crowd counting, and object detection.
\end{IEEEbiography}

\begin{IEEEbiography}[{\includegraphics[width=1in,height=1.25in,clip,keepaspectratio]{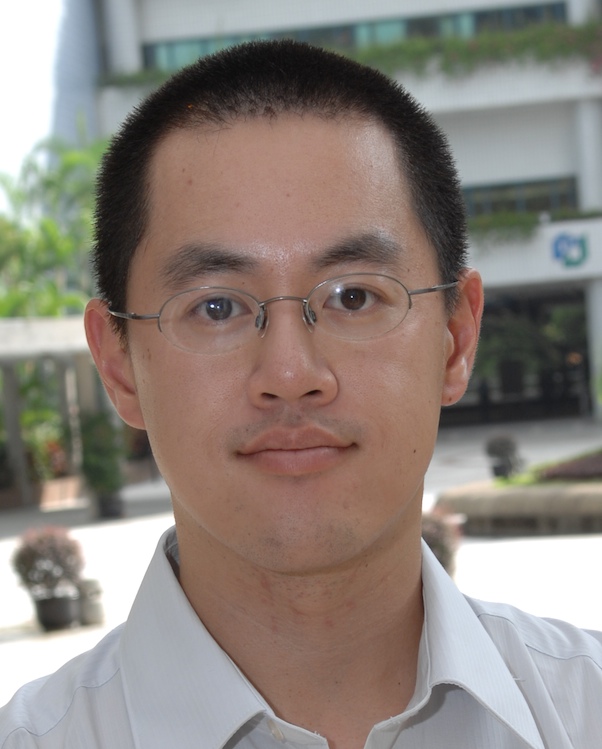}}]
{Antoni B. Chan} (SM'15) received the B.S. and M.Eng. degrees in electrical engineering from Cornell University, Ithaca, NY, in 2000 and 2001, and the Ph.D. degree in electrical and computer engineering from the University of California, San Diego (UCSD), San Diego, in 2008.
In 2009, he was a Postdoctoral Researcher with the Statistical Visual Computing Laboratory, UCSD. In 2009, he joined the Department of Computer Science, City University of Hong Kong, Hong Kong, where he is currently an Associate Professor. His research interests include computer vision, machine learning, pattern recognition, and music analysis.
Dr. Chan was the recipient of an NSF IGERT Fellowship from 2006 to 2008, and an Early Career Award in 2012 from the Research Grants Council of the Hong Kong SAR, China.
\end{IEEEbiography}

\clearpage
\end{document}